\documentclass{article}

\PassOptionsToPackage{numbers,sort&compress}{natbib}
\usepackage[preprint]{neurips_2024}

\usepackage[utf8]{inputenc} %
\usepackage[T1]{fontenc}    %
\usepackage{url}            %
\usepackage{booktabs}       %
\usepackage{amsfonts}       %
\usepackage{nicefrac}       %
\usepackage{microtype}      %
\usepackage{graphicx}
\usepackage[dvipsnames]{xcolor}         %
\usepackage{xspace}
\usepackage[export]{adjustbox}
\usepackage{wrapfig}
\usepackage{caption}
\usepackage{subcaption}
\usepackage{multirow}
\usepackage{amsmath}       %
\usepackage{amssymb}
\usepackage{pifont}
\usepackage{enumitem}
\usepackage{placeins}
\usepackage{numprint}
\npdecimalsign{.}
\newcommand{\np}{\numprint}

\makeatletter
\DeclareRobustCommand\onedot{\futurelet\@let@token\@onedot}
\def\@onedot{\ifx\@let@token.\else.\null\fi\xspace}
\makeatother

\newcommand{\eg}{e.g\onedot}
\newcommand{\ie}{i.e\onedot}
\newcommand{\cf}{cf\onedot}

\newcommand{\versus}{vs\onedot}

\definecolor{ourblue}{rgb}{0.368,0.507,0.71}
\definecolor{ourorange}{rgb}{0.881,0.611,0.142}
\definecolor{ourgreen}{rgb}{0.56,0.692,0.195}
\definecolor{ourred}{rgb}{0.923,0.386,0.209}
\definecolor{ourviolet}{rgb}{0.528,0.471,0.701}
\definecolor{ourbrown}{rgb}{0.772,0.432,0.102}
\definecolor{ourlightblue}{rgb}{0.364,0.619,0.782}
\definecolor{ourdarkgreen}{rgb}{0.572,0.586,0.}

\definecolor{ourcyan2}{rgb}{0.125,0.722,0.804}
\definecolor{ourred2}{rgb}{0.863,0.184,0.047}
\definecolor{ouryellow2}{cmyk}{0,0.16,1.0,0.07}
\definecolor{ourviolet2}{cmyk}{0.55,0.56,0,0.47}
\definecolor{ourorange2}{cmyk}{0,0.46,0.89,0.11}

\usepackage{hyperref}       %
\hypersetup{colorlinks,
            linkcolor=BlueViolet,
            urlcolor=BlueViolet,
            citecolor=BlueViolet}
\usepackage[capitalise,nameinlink,sort&compress]{cleveref}
\crefname{section}{Sec.}{Secs.}
\crefname{table}{Tab.}{Tabs.}
\crefname{appendix}{App.}{Apps.}

\newcommand{\cmark}{\ding{51}}%
\newcommand{\xmark}{\ding{55}}%

\definecolor{samcolor}{rgb}{0.3, 0.3, 0.3}

\newcommand{\method}{FT-\textsc{Dinosaur}\xspace}
\newcommand{\methodshort}{FT-\textsc{Dinosaur}\xspace}

\newcommand{\DINOSAUR}{\textsc{Dinosaur}\xspace}
\newcommand{\SPOT}{\textsc{Spot}\xspace}
\newcommand{\SAM}{\textsc{Sam}\xspace}
\newcommand{\SAMcomp}{\textsc{Sam} \textit{(comp.)}\xspace}
\newcommand{\SAMbest}{\textsc{Sam} \textit{(best.)}\xspace}
\newcommand{\DINO}{\textsc{Dino}\xspace}
\newcommand{\COCO}{\textsc{Coco}\xspace}
\newcommand{\PASCALVOC}{\textsc{Pascal Voc}\xspace}
\newcommand{\ScanNet}{\textsc{ScanNet}\xspace}
\newcommand{\ClevrTex}{\textsc{ClevrTex}\xspace}
\newcommand{\EntitySeg}{\textsc{EntitySeg}\xspace}
\newcommand{\YCB}{\textsc{Ycb}\xspace}
\newcommand{\MOVi}{\textsc{Mov}i\xspace}

\usepackage{amsmath,amsfonts,bm}

\def\1{\bm{1}}

\def\va{{\bm{a}}}

\def\vm{{\bm{m}}}

\def\vx{{\bm{x}}}
\def\vy{{\bm{y}}}

\DeclareMathAlphabet{\mathsfit}{\encodingdefault}{\sfdefault}{m}{sl}
\SetMathAlphabet{\mathsfit}{bold}{\encodingdefault}{\sfdefault}{bx}{n}

\def\gK{{\mathcal{K}}}

\DeclareMathOperator*{\argmax}{arg\,max}

\DeclareMathOperator*{\topk}{topk}
\DeclareMathOperator*{\mysoftmax}{softmax}

\usepackage{todonotes}

\title{Zero-Shot Object-Centric Representation Learning}

\author{%
  Aniket Didolkar$^{1,}$\thanks{equal contribution. Correspondence: \texttt{maximilian.seitzer@tue.mpg.de},\ \ \texttt{adidolkar123@gmail.com}.\\Website: \href{https://rw-ocrl.github.io/ftdinosaur-paper/}{\texttt{rw-ocrl.github.io/ftdinosaur-paper}}}
  \quad\quad
  Andrii Zadaianchuk$^2$
  \quad\quad
  Anirudh Goyal$^1$
  \quad\quad
  Mike C.~Mozer$^3$\\[0.25em]
  \textbf{Yoshua Bengio}$^1$
  \quad\quad\quad
  \textbf{Georg Martius}$^{4}$
  \quad\quad\quad
  \textbf{Maximilian Seitzer}$^{4,\ast}$\\[0.5em]
  $^1$MILA \& University of Montreal\quad\quad$^2$University of Amsterdam\\[0.25em]
  $^3$CU Boulder\quad\quad$^4$MPI for Intelligent Systems \& University of Tübingen
}

\begin{document}

\maketitle

\begin{abstract}
    The goal of object-centric representation learning is to decompose visual scenes into a structured representation that isolates the entities.
    Recent successes have shown that object-centric representation learning can be scaled to real-world scenes by utilizing pre-trained self-supervised features.
    However, so far, object-centric methods have mostly been applied in-distribution, with models trained and evaluated on the same dataset.
    This is in contrast to the wider trend in machine learning towards general-purpose models directly applicable to unseen data and tasks.
    Thus, in this work, we study current object-centric methods through the lens of zero-shot generalization by introducing a benchmark comprising eight different synthetic and real-world datasets. 
    We analyze the factors influencing zero-shot performance and find that training on diverse real-world images improves transferability to unseen scenarios.
    Furthermore, inspired by the success of task-specific fine-tuning in foundation models, we introduce a novel fine-tuning strategy to adapt pre-trained vision encoders for the task of object discovery. 
    We find that the proposed approach results in state-of-the-art performance for unsupervised object discovery, exhibiting strong zero-shot transfer to unseen datasets.
  
\end{abstract}

\section{Introduction}

In the past decade, deep learning-based approaches have become ever more general, culminating in models that exhibit broad and flexible vision~\citep{Dehghani2023ViT22B,oquab2023dinov2} and language understanding~\citep{brown2020language,openai2024gpt4}.
These so-called foundation models can be applied to a variety of tasks, either in a zero-shot manner~\citep{brown2020language}, or via task-specific finetuning~\citep{Ziegler2019RLHF}. %
An open challenge is how to equip these models with the ability to robustly reason about visual inputs, that is, in a manner that supports compositional generalization and causal inference~\citep{scholkopf2021toward, goyal2022inductive}.
Evidence suggests that human cognition deals with these problems by dynamically binding raw perceptual features into symbol-like entities that can be flexibly composed together and reasoned over~\citep{Pinker1984VisualCognition, Spelke1990ObjPerception, Spelke2000CoreKnowledge}.
Inspired by these findings, the field of \emph{object-centric representation learning} aims to replicate these abilities in deep learning models~\citep{greff2020binding}.
By mirroring the compositional generative process of the world~\citep{Brady23provablylearning}, these methods learn to decompose visual scenes into structured representations capturing the objects in the scene in a fully unsupervised way.
Not only can object-centric representations provably exhibit compositional generalization~\citep{wiedemer2024provable}, they also support a diverse set of downstream tasks such as world modeling~\citep{ke2021systematic,wu2023slotformer}, robotic control~\citep{zadaianchuk2020smorl,haramati2024entitycentric,driess2023palme,didolkar2024cycle}, visual question answering~\citep{Xu2024SlotVLMSS,mamaghan2024exploringVQA}, and compositional generation in 2D~\citep{Singh2022SLATE,wu2023slotdiffusion,jiang2023object} and 3D~\citep{sajjadi2022osrt,Jabri2023DORSal}.

While long confined to simplistic synthetic datasets~\citep{eslami2016attend,greff2019multi,Engelcke2020GENESIS,Locatello2020SlotAttention}, recent progress has scaled object-centric representations to complex real-world image~\citep{seitzer2023bridging,wu2023slotdiffusion,jiang2023object,kakogeorgiou2023spot,lowe2024rotating}, and video datasets~\citep{zadaianchuk2023videosaur,aydemir2023self}.
This opens the door to evaluate these methods in terms of their zero-shot transferability to new data, \ie their ability to discover objects in unseen scenarios.
While this is common practice in other areas of deep learning~\citep{brown2020language, openai2024gpt4, bubeck2023sparks}, object-centric models have so far not been studied under this lens.
To close this gap, in this work, we focus on \emph{zero-shot object-centric representation learning}.

In particular, we introduce a benchmark consisting of 8 datasets comprising a diverse range of synthetic and real-world scenes. 
Using this benchmark, we 1) seek to understand the zero-shot transfer capabilities of existing models, and 2) study the properties of training datasets that influence generalization.
The general conclusion we draw from this benchmark is that object-centric models which are trained on naturalistic datasets consisting a variety of objects --- such as \COCO ~\citep{Lin2014COCO} --- usually exhibit decent zero-shot generalization.

Equipped with this knowledge, we aim to build a strong general-purpose object-centric model. 
To achieve this, we first make the observation that current approaches for real-world object-centric learning~\citep{seitzer2023bridging,wu2023slotdiffusion,jiang2023object, zadaianchuk2023videosaur,aydemir2023self} use \emph{fixed} pre-trained encoders (\eg with the \DINO{} method~\citep{Caron2021DINO}) to encode the input.
This may be limiting as, while the pre-trained encoders offer good general-purpose features, they may not be optimal for the task of object discovery. 
Instead, we propose to finetune the encoder parameters for the target task; to this end, we introduce a suitable training recipe as well as a novel decoder that reduces the increased computational costs from finetuning.
Building on the \DINOSAUR model~\citep{seitzer2023bridging}, our proposed finetuning approach sets a new state-of-the-art for real-world object-centric learning on the \COCO dataset, as well as in the zero-shot setting.
Our method shows zero-shot transfer across a multitude of diverse datasets, often achieving and even surpassing the in-distribution performance on these datasets.

Our contributions are as follows:
{\setlength{\topsep}{0.0em}
\setlength{\itemsep}{0em}
\begin{itemize}
    \item We introduce a benchmark to evaluate the zero-shot generalization of object discovery methods (\cref{subsec:zero-shot-benchmark}).
    \item Using the benchmark, we analyze the zero-shot capabilities of object-centric models (\cref{subsec:evaluating-models}) and investigate dataset properties for training generalizable models (\cref{subsec:evaluating-data}).
    \item We propose a finetuning approach applied to \DINOSAUR, which allows to stably adapt the parameters of the pre-trained encoder for the task of object discovery (\cref{sec:finetuning}).
    \item Our method achieves state-of-the-art results across various in-distribution and out-of-distribution scenarios (\cref{sec:evaluation}).
\end{itemize}
}

\section{Related Work}
\label{sec:related-work}

\paragraph{Object-Centric Learning on Real-World Datasets}
Originally, object-centric methods were mostly applied to synthetic data with limited complexity~\citep{johnson2017clevr, Karazija2021clevrtex, Greff2021Kubric} and trained from scratch~\citep{eslami2016attend, burgess2019monet, lin2020space, Locatello2020SlotAttention, traub2023:Loci}.
Recently, there has been considerable interest~\cite{elsayed2022savi++, Singh2022STEVE, seitzer2023bridging, didolkar2024cycle, lowe2024rotating, zadaianchuk2023videosaur, wu2023slotdiffusion, kakogeorgiou2023spot} in scaling those methods to complex and unconstrained real-world image and video datasets like \COCO~\citep{Lin2014COCO} or YouTube-VIS~\citep{vis2019}. 
Current state-of-the-art techniques~\citep{seitzer2023bridging, wu2023slotdiffusion, jiang2023object, kakogeorgiou2023spot, lowe2024rotating, aydemir2023self, zadaianchuk2023videosaur} rely on applying slot attention~\citep{Locatello2020SlotAttention} to frozen vision transformers (ViT)~\citep{Dosovitskiy2021ViT} pre-trained with contemporary self-supervised representation learning methods~\citep{Caron2021DINO,He2022MAE,Chen2021MoCov3,Assran2022MSN}. 
Approaches differ by their learning objective; one line of models is based on \DINOSAUR~\citep{seitzer2023bridging} and utilizes a feature reconstruction objective~\citep{seitzer2023bridging,kakogeorgiou2023spot,zadaianchuk2023videosaur,aydemir2023self}, whereas others apply diffusion objectives~\citep{jiang2023object,wu2023slotdiffusion}.
Although these techniques confirm that object-centric representation learning \emph{is} possible for complex real-world inputs, they are also limited by the quality of self-supervised encoders, as those encoders remain frozen during object-centric training. 
In contrast, our method, while starting from self-supervised features, adapts them through object-centric finetuning, making them more suitable for the task of object-centric scene decomposition. 

\paragraph{Task-Specific Finetuning} 
The idea of finetuning a pretrained model for a specific task or dataset has been around for quite some time. 
Early works in deep learning demonstrated that finetuning a pretrained model on a specific task often leads to better performance as compared to training the model from scratch~\citep{huh2016makes, yosinski2014transferable, dai2015semi}. 
The advent of large pre-trained (foundation) models has further popularized this approach in recent years~\citep{devlin2018bert, radford2018improving, chen2020big, Caron2021DINO, He2022MAE, kirillov2023segment, Dehghani2023ViT22B, oquab2023dinov2}. 
The central idea is that a large model is first pre-trained on a large and diverse dataset to obtain strong general-purpose features. 
This large model is then adapted to specific tasks by finetuning the entire model \citep{Howard2018UniversalLM, devlin2018bert, sun2019fine} or parts of the model \citep{zhou2021closer, shen2021partial}.
Specifically for self-supervised vision representations~\citep{Caron2021DINO}, adapting features was studied for the tasks of unsupervised semantic segmentation~\citep{Ziegler2022SelfSupervisedLO, Zadaianchuk2022COMUS, Hamilton2022Stego} and multi-object tracking in videos~\citep{salehi2023time, tumanyan2024dino}.
To our knowledge, the only work which applies finetuning in the context of object-centric models is \SPOT~\citep{kakogeorgiou2023spot}, using a two-stage procedure that enables finetuning the final four layers of a pre-trained encoder during the second stage.
In comparison, we introduce a finetuning approach that adapts the full encoder to the task of object discovery, and empirically show that this leads to significantly stronger object discovery performance compared to \SPOT.

\paragraph{Zero-shot Generalization} 
A paradigm first introduced and formalized by \citet{larochelle2008zero}, zero-shot generalization enables models to perform well on tasks or datasets not seen during training.
Recently, zero-shot generalization has become more prevalent in deep learning due to the availability of large foundation models~\citep{openai2024gpt4, oquab2023dinov2, kirillov2023segment}; these models utilize large-scale pre-training to develop robust general-purpose abilities that can be applied zero-shot to various tasks and datasets. 
In the context of object-centric learning, \citet{dittadi2022generalization} conducted a systematic study of object-centric models under various types of distribution shifts, such as color changes, texture changes, and occlusions. 
In our work, we adopt a slightly different definition of zero-shot generalization for object-centric models, aligning more closely with the literature on foundation models. 
Specifically, we aim to obtain general-purpose object-centric features through finetuning, which can then be applied zero-shot to any new and unseen dataset.%

\section{What Matters for Zero-Shot Transfer of Object-Centric Representations?}
\label{sec:zero-shot}

In this section, our goal is to understand factors influencing zero-shot performance of current object-centric models.
We first introduce a benchmark for measuring zero-shot performance in \cref{subsec:zero-shot-benchmark} and compare different models
 in \cref{subsec:evaluating-models}. 
In \cref{subsec:evaluating-data}, we investigate the role of the training data.

\subsection{Benchmark}
\label{subsec:zero-shot-benchmark}

\begin{figure}
    \centering
    \begin{subfigure}[t]{0.333\textwidth}
        \centering
        \includegraphics{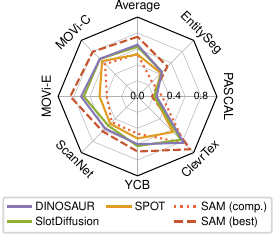}
        \caption{Varying models.}
        \label{fig:benchmark-methods}
    \end{subfigure}%
    \hfill%
    \begin{subfigure}[t]{0.333\textwidth}
        \centering
        \includegraphics{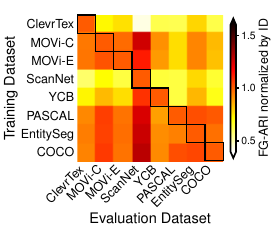}
        \captionsetup{width=.8\linewidth}
        \caption{Varying train datasets.}
        \label{fig:benchmark-datasets}
    \end{subfigure}%
    \hfill%
    \begin{subfigure}[t]{0.333\textwidth}
        \centering
        \includegraphics{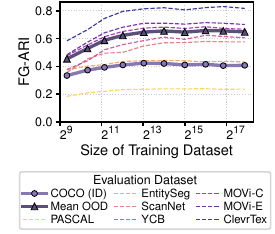}
        \captionsetup{width=.8\linewidth}
        \caption{Varying train dataset size.}
        \label{fig:benchmark-dataset-size}
    \end{subfigure}%
    \caption{\textbf{Evaluating zero-shot transfer of object-centric representations.}
    Performance given in FG-ARI, see \cref{app:fig:benchmark-mbo} for corresponding plots with mBO.
    (a): performance of current object-centric models trained on the \COCO dataset.
    (b): performance of the \DINOSAUR method~\citep{seitzer2023bridging} with different training datasets.
    (c): scaling behavior of \DINOSAUR training on differently sized subsets of \COCO.
    }
    \label{fig:benchmark}
\end{figure}

\paragraph{Datasets}
We argue that object-centric models should be able to discover and capture objects in a variety of conditions. 
Conveniently, the object-centric community has proposed many datasets of increasing complexity to challenge their models.
Thus, to obtain a test bed that robustly measures zero-shot performance, we take the evaluation splits of \ClevrTex~\citep{Karazija2021clevrtex}, \MOVi-C and \MOVi-E~
\citep{Greff2021Kubric}, \ScanNet and \YCB as used in~\citet{yang2022promising}, \PASCALVOC~\citep{Everingham2012PASCALVOC}, and \COCO 2017~\citep{Lin2014COCO}. 
Additionally, we add the challenging \EntitySeg dataset~\citep{qilu2023entityseg}, consisting of open-world real-world images with high-quality mask annotations.
In total, we gathered 8 datasets with a total of \numprint{25323} images.
For further details on the datasets, we refer to \cref{app:sec:datasets}.

Importantly, we do not specify the \emph{training data}; as a consequence, we can also evaluate the zero-shot behavior of supervised models such as Segment Anything (\cref{subsec:evaluating-models}).
Furthermore, this allows us to study the impact of different kinds of datasets for training (\cref{subsec:evaluating-data}).
Note that current object-centric models are sensitive to the \emph{number of objects}. 
As a concession to that, we evaluate the models with the number-of-slots parameter matching the expected complexity of the target dataset (mostly following prior work, see \cref{app:tab:dataset_properties}).
We leave it to future work to remove this limitation of the models.

\paragraph{Metrics}
We evaluate the quality of the object representation in terms of the masks associated with each object, using the instance mask annotations as reference.
To do so, we compute the commonly used \emph{foreground ARI} (FG-ARI)~\citep{rand1971objective, hubert1985comparing}, measuring how well the discovered objects follow the separation prescribed by the reference masks.
While previous work has argued against the use of FG-ARI~\citep{Karazija2021clevrtex,wu2023slotdiffusion,kakogeorgiou2023spot}, we think that it is still useful as the primary measure of the quality of object splitting.
In addition, we compute the \emph{mean best overlap} (mBO)~\citep{Arbelaez2014MCG}, measuring how well the discovered masks fit to objects.

On the \COCO dataset, which has a \emph{panoptic} labeling (object ``things'' and background ``stuff''), we additionally evaluate \emph{scene decomposition}.
This is sensible because on real-world images, there is no clear distinction between objects and background from the model's point-of-view.
To this end, we compute \emph{panoptic ARI} (P-ARI) and \emph{class-agnostic panoptic quality} (PQ), where the latter measures both mask quality and precision/recall \citep{kirillov2019panoptic}.
We refer to \cref{app:sec:metrics} for more information about metrics.
To aggregate results over different datasets, we compute the \emph{per-sample average}, normalizing by the dataset size (see \cref{app:tab:dataset_properties}).

Finally, we remark that there are other ways to evaluate the quality of object-centric representations, for instance inspecting the content of the learned representation or their use for downstream tasks.
We consider these as orthogonal to our mask-based evaluation and defer them to future work.

\subsection{Evaluating Models}
\label{subsec:evaluating-models}

We evaluate three recent state-of-the-art object methods capable of real-world object-centric learning:
\DINOSAUR~\citep{seitzer2023bridging}, using pre-trained self-supervised features as inputs and targets; \SPOT~\citep{kakogeorgiou2023spot}, which builds upon \DINOSAUR by improving the decoder; and SlotDiffusion~\citep{wu2023slotdiffusion}, which also utilizes pre-trained features, but uses a diffusion decoder.
We limit ourselves to pre-trained feature-based methods as other approaches have not shown scalability to real-world data.
All models are trained on the COCO dataset.
In addition, we evaluate the Segment Anything model (\SAM)~\citep{kirillov2023segment}, a supervised segmentation foundation model.
To showcase the gap between state-of-the-art supervised and unsupervised methods for object discovery, we use the largest available model (ViT-Huge), and pick the mask confidence threshold resulting in the best performance per-dataset (\SAM (best)); in contrast to current object-centric methods, this results in a variable number of masks per-image.
For better comparability, we also evaluate a baseline \SAM (comp.), using a ViT-Base encoder and a fixed number of masks.
Please refer to \cref{app:sec:method-details} for details about the models.

We present the results in \cref{fig:benchmark-methods}.
We find that SlotDiffusion and \DINOSAUR exhibit similar zero-shot FG-ARI performance while outperforming \SPOT on all datasets. 
\SAM (best) achieves the best FG-ARI performance on all but one dataset (\PASCALVOC). 
In terms of mBO (\cref{app:fig:benchmark-mbo}), we observe that SlotDiffusion and \SPOT both outperform \DINOSAUR with SlotDiffusion achieving the best performance. 
Again we find that \SAM (best) achieves the highest performance with the difference being even more pronounced than that for FG-ARI. 
We note that the superiority of \SAM (best) is expected and stems from being trained with extensive supervision, as well as its ability to output a variable number of masks; this allows \SAM to discover smaller objects not captured by methods with a fixed number of slots. 
If we remove that privilege, the SAM (comp.) baseline is on many datasets inferior to the unsupervised models, with the exception of \ClevrTex and \PASCALVOC.
Thus, we conclude that object-centric models already exhibit decent zero-shot generalization to unseen datasets. 
Moreover, even though all the methods are trained with a fixed number of 7 slots on \COCO, they are evaluated with a different number of slots on each target dataset (see \cref{app:tab:dataset_properties}).
This further speaks to the zero-shot transferability of existing unsupervised slot-based methods.

\subsection{Evaluating Training Data}
\label{subsec:evaluating-data}

So far, we used the same training data for comparing the models --- a natural question is how the \emph{training data} affects zero-shot behavior.
To answer it, we train \DINOSAUR on different training datasets and evaluate the zero-shot performance on our benchmark.
We organize our experiments into two groups: 1) varying the data distribution, and 2) varying the amount of samples from a particular data distribution. 
The former allows us to identify properties of the data that influence zero-shot behavior, whereas the latter investigates how models scale with data.

\paragraph{Properties of the Data Distribution}
To obtain training datasets with different properties, we utilize the training splits belonging to the benchmark datasets listed in \cref{subsec:zero-shot-benchmark}.
We characterize the training datasets along three dimensions: \textbf{realism} ---  in terms of the three categories synthetic (\ClevrTex), hybrid (\MOVi, \ScanNet, \YCB) and natural (\PASCALVOC, \COCO, \EntitySeg); \textbf{diversity} --- on a spectrum from narrow to broad (roughly \ClevrTex $\ll$ \ScanNet, \YCB, \PASCALVOC $\ll$ \MOVi $\ll$ \COCO $\ll$ \EntitySeg); and the \textbf{amount of objects} --- ranging from few (\PASCALVOC) to moderate (up to 6; \ClevrTex, \ScanNet, \YCB) to many (\MOVi, \COCO, \EntitySeg).
The results are shown in \cref{fig:benchmark-datasets}.\looseness-1

First, we find that training and evaluating in-distribution, \ie on matching datasets, unsurprisingly performs best in general. 
Training on \emph{synthetic} and \emph{hybrid} datasets transfers well to datasets in those categories, but not to natural data; conversely, \emph{natural} data transfers well to synthetic and hybrid data. 
Next, we find that the zero-shot performance is fairly similar when trained on \COCO and \EntitySeg (high diversity, many objects), or \PASCALVOC (less diversity, few objects). 
This shows that having complex natural data is more important for zero-shot performance compared to data diversity.
Moreover, even when trained on natural data with few objects (\eg \PASCALVOC), the model transfers well to datasets with more objects such as \MOVi, \COCO, and \EntitySeg.
Overall, we can conclude that training on natural data leads to strong zero-shot performance of current object-centric models.
We pick \COCO as the main training dataset for zero-shot object-centric models for the remainder of this work.

\paragraph{Effect of Data Scale}
We now investigate the effect of the number of training data points.
To do so, we train \DINOSAUR on differently sized subsets of the \COCO dataset (up to 240k samples when including the \emph{``unlabeled''} split).
From the results in \cref{fig:benchmark-dataset-size}, we find that in-distribution performance plateaus around \numprint{8192} ($2^{13}$) samples, and zero-shot performance around \numprint{16182} ($2^{14}$) samples. 
Intriguingly, this shows that current object-centric models can be very sample efficient in obtaining decent in-distribution and competitive zero-shot generalization.
However, we do not find evidence of favorable data scaling laws.

\subsection{Summary}

In summary, we find that existing object-centric models (\DINOSAUR, \SPOT, SlotDiffusion) already exhibit decent zero-shot transfer to unseen domains. 
Their success can potentially be attributed to using pre-trained general-purpose encoders as a base for object discovery. 
Furthermore, our experiments show that training on complex natural data is an important component for zero-shot transfer which can be attributed to the inherent complexities associated with such data. 
In addition, real-world datasets offer a significantly larger catalog of objects and instances to train on compared to synthetic or hybrid datasets. 

Equipped with this knowledge, we shift our focus to enhancing the performance of unsupervised object-centric models. 
Specifically, the question we ask is: \textit{Can we improve object discovery by finetuning pre-trained encoders specifically for the task of object discovery?}

\section{Object-Centric Finetuning}
\label{sec:finetuning}

\begin{figure}
    \includegraphics[width=\textwidth]{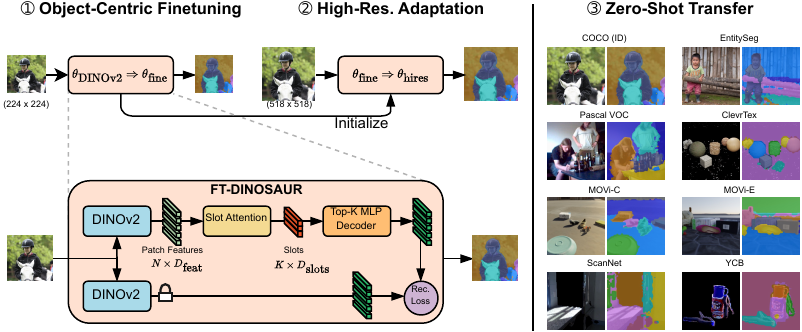}
    \caption{
        \textbf{Overview of our method ``\method{}''.} 
        \textbf{\ding{192}~Object-Centric Finetuning}: starting from \DINO{}v2, the encoder is finetuned for the task of object discovery on the \COCO dataset. 
        \textbf{\ding{193}~High-Res Adaptation}: the model is further adapted to high-resolution images.
        \textbf{\ding{194}~Zero-Shot Transfer}: at test time, we apply the trained model to 8 datasets from our proposed zero-shot benchmark (\cref{subsec:zero-shot-benchmark}).
    }   
    \label{fig:method}
\end{figure}

Current methods for real-world object-centric learning~\citep{seitzer2023bridging,jiang2023object,wu2023slotdiffusion,kakogeorgiou2023spot,zadaianchuk2023videosaur,aydemir2023self,lowe2024rotating} are all based on pre-trained self-supervised features~\citep{Caron2021DINO,He2022MAE,oquab2023dinov2}.
While those features offer good performance for many downstream tasks out of the box, they are not explicitly designed for the \emph{task of object discovery}. 
We conjecture that this gap between training and downstream objective leads to sub-optimal transfer performance. 
Thus, we propose to adapt the pre-trained features by \emph{task-specific finetuning} ---
\cref{fig:method} shows our approach.\looseness-1

\subsection{Finetuned \DINOSAUR} 
\label{sec:method}  %

\paragraph{Finetuning}
We first describe how we adapt the \DINOSAUR architecture~\citep{seitzer2023bridging} for finetuning.
\DINOSAUR uses a pre-trained ViT as the encoder that is kept fixed during training. 
The original work reported that unfreezing the encoder leads to a collapse; this is because the encoder features are simultaneously used as the model's prediction targets.
To sidestep this problem, we add a \emph{target encoder} that is initialized to be a copy of the original encoder, but kept fixed throughout training. 
This allows us train the full model end-to-end without collapse.

We found that the encoder would initially drift away from its pre-trained initialization, likely induced by the noisy gradients from the randomly initialized slot attention module.
To reduce the effect of this, we introduce blockwise exponentially decaying learning rates~\citep{Howard2018UniversalLM} for the encoder.
Furthermore, we found an improved set of hyperparameters, namely a lower learning rate, switching to a cosine learning rate schedule~\citep{Loshchilov2017SGDRSG}, lower gradient clipping, weight decay on the encoder and a higher batch size.
Showing the efficacy of this improved setup, we find that we can now also train the ViT encoder from a \emph{random initialization} (42.3 FG-ARI, 27.3 mBO), a scenario which was previously reported as leading to collapse by \citet{seitzer2023bridging}.
We detail the exact settings in \cref{app:subsec:imp-hps}.
We also experimented with an EMA student-teacher setup to continuously adapt the targets throughout training, but found that this leads to worse results (see \cref{app:subsec:ema}).

\paragraph{High Resolution Adaptation}
A further way to make more effective usage of data is to increase the image resolution.
Standard ViTs use a relatively low resolution of $224 \times 224$ pixels, leading to a patch resolution of $16 \times 16$ when trained with patch size $14$.
This hides details, inhibits capturing smaller objects, and leads to coarser objects masks.
Thus, after training at $224 \times 224$ resolution, we add a short second stage of training, in which the model is adapted to image resolution to $518 \times 518$ (\ie $37 \times 37$ patches) over \numprint{10000} steps.
This is similar \DINO{}v2's training strategy~\citep{oquab2023dinov2}, and adds significant improvements (\cref{subsec:method-ablation}) without a high computational burden.

\paragraph{Efficient Top-K Decoding}
\label{subsec:topk-decoding}
Finetuning the encoder and high resolution adaptation both significantly increase the costs in terms of computation and memory.
To mitigate this, we introduce a novel efficient decoding approach based on the MLP decoder introduced by \citet{seitzer2023bridging}, which we call \emph{top-k decoding}.
For each of $N$ patches, the MLP decoder produces an output by combining the predictions over $C$ slots using a slot-wise weighted average, resulting in a computational cost of $\mathcal{O}(N \cdot C)$.
Our insight is that \emph{most of this computation is wasted}, as slots are localized and mostly sparsely distributed across the image --- instead, it suffices to decode the \emph{$k$ most likely slots} occupying a patch, reducing the costs to $\mathcal{O}(N)$ for constant $k$.
While we do not have access to the true occupation probabilities apriori, empirically we found that the masks from slot attention can serve as a good proxy.
We refer to \cref{app:subsec:topk} for more details.

\subsection{Analysis}
\label{subsec:method-analysis}

\begin{figure}
    \centering
    \includegraphics[width=0.485\textwidth]{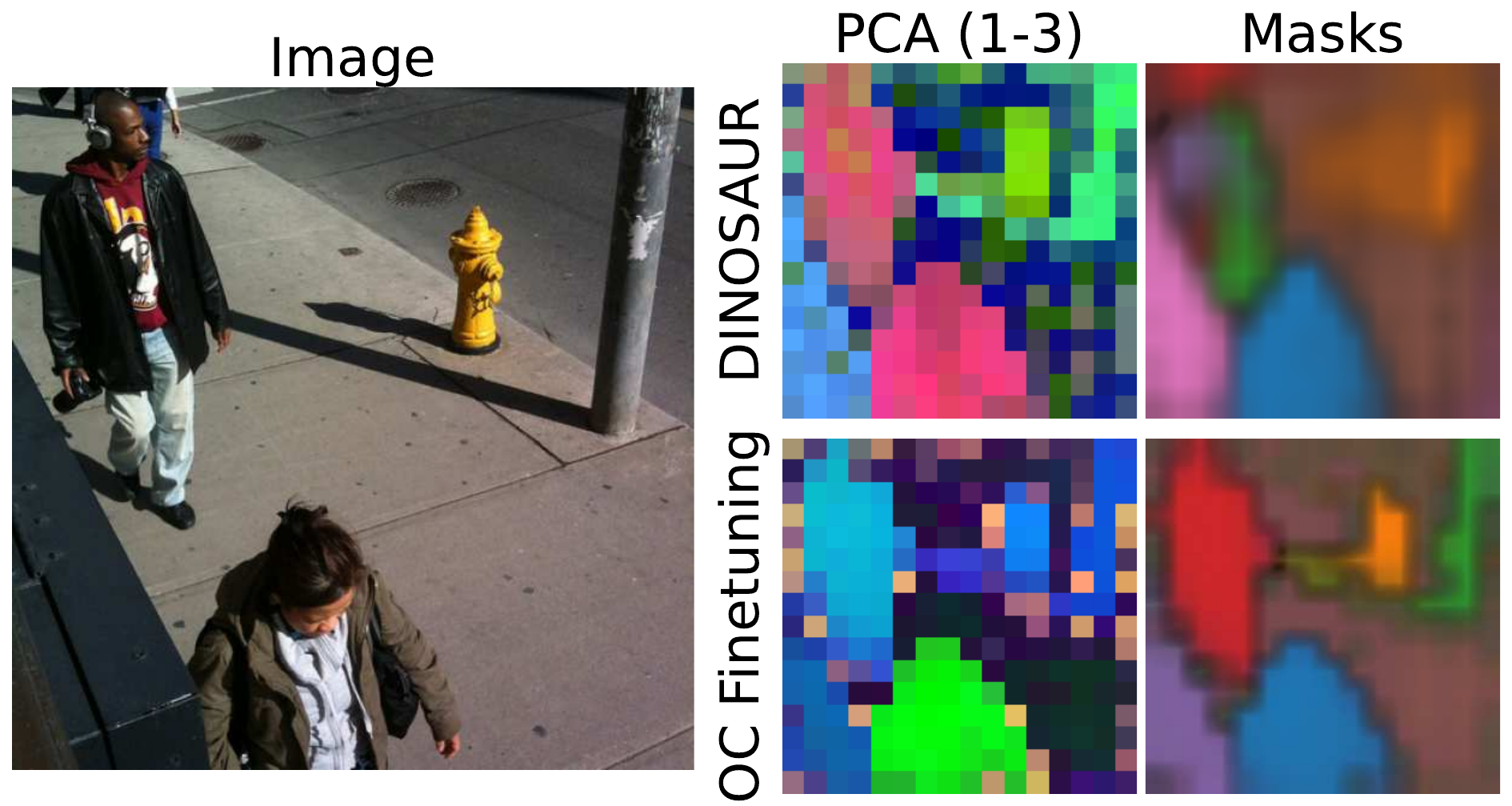}\hfill
    \includegraphics[width=0.485\textwidth]{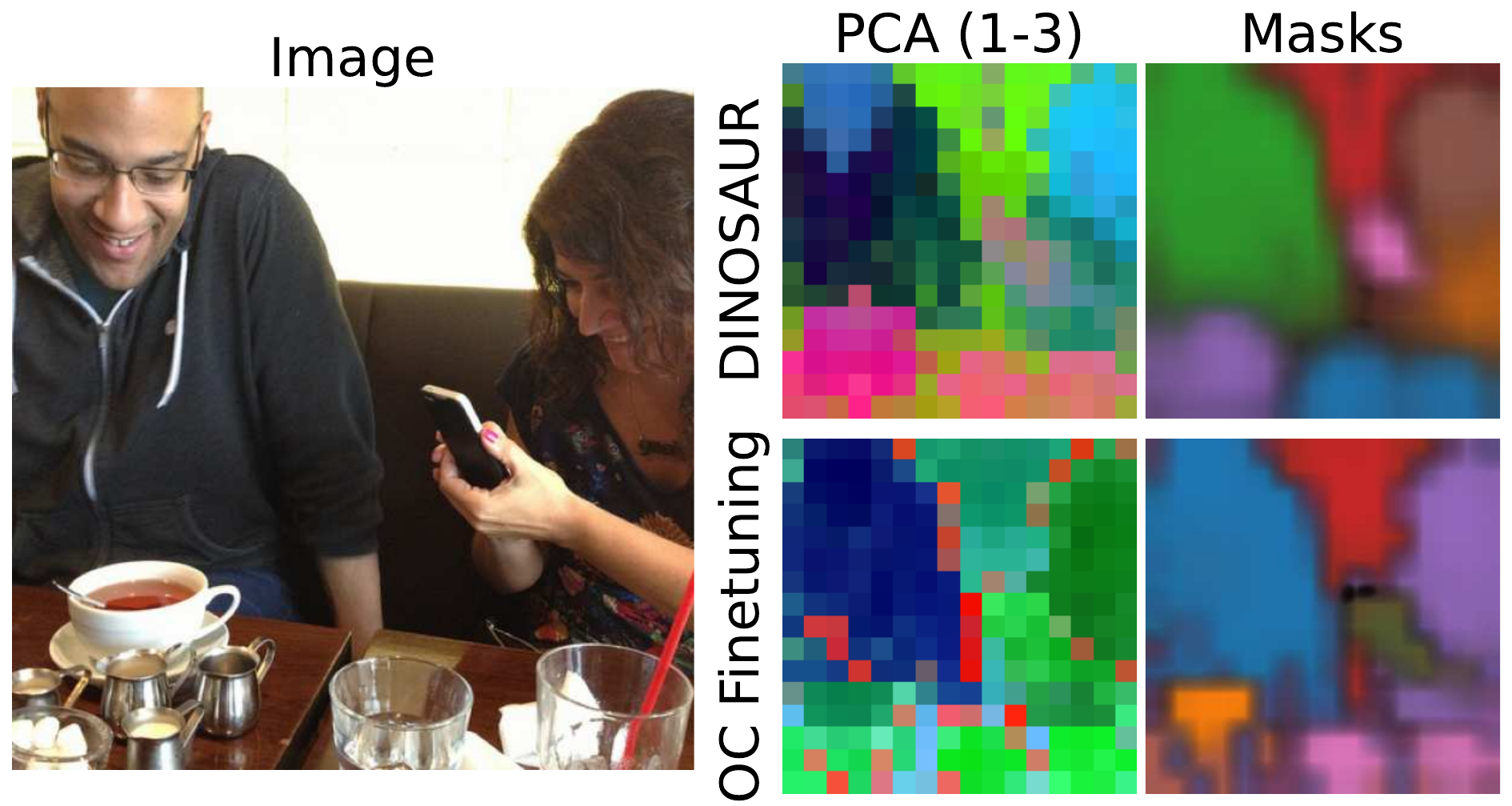}
    \caption{\textbf{Visualization of encoder features in \DINOSAUR (frozen \DINO{}v2 features) and for features adapted with object-centric finetuning.} 
    We show the 1st to 3rd PCA components visualized by different RGB channels~(second column).
    The last column shows scene decomposition masks by each method. 
    More examples and additional PCA components are shown in \cref{app:fig:analysis-appendix}.}
    \label{fig:analysis} 
\end{figure}

Object-centric finetuning adapts the pre-trained encoder such that the original \DINO{}v2 features can be predicted better, with the slot representations acting as a bottleneck.
To better understand the effect of this procedure, we study how the encoder representations change after finetuning.
In \cref{fig:analysis} and \cref{app:fig:analysis-appendix}, we show the first PCA components obtained from \DINO{}v2 features (used by \DINOSAUR) and features after object-centric finetuning. 
\DINO{}v2 features mainly exhibit semantic similarity, \ie one component often corresponds to several different objects or parts of the same category (such as human heads).
In contrast, after object-centric finetuning, PCA components are noticeably object-centric, splitting instances of the same category and grouping together different object parts into one component. 
To confirm this observation quantitatively, we apply per-image k-means clustering to the two types of features.
On \COCO, we find that the clustering of features from object-centric finetuning corresponds better to object instances, reaching 34.0 FG-ARI and 28.7 mBO in contrast to 27.4 FG-ARI and 24.7 mBO for the original \DINO{}v2 features.

\subsection{Ablations}
\label{subsec:method-ablation}

\begin{table}
    \caption{\textbf{Ablation study on \COCO}. Starting from \DINOSAUR~\citep{seitzer2023bridging} (first row), we ablate the impact of switching to \DINO{}v2, finetuning the encoder (FT), improving general (G-HP) and encoder hyperparameters (E-HP), adding top-k decoding and high-resolution adaptation. 
    Results averaged over 3 random seeds besides last two rows, which use 5 seeds.
    }\vspace{.3em}
    \label{tab:method-ablation-coco}
    \begin{minipage}{0.58\textwidth}
    {
        \setlength{\tabcolsep}{2pt}
        \small
        \centering
        \begin{tabular}{@{}l@{\hskip1ex}ccccccc@{}}
            \toprule
            Model & FT & G-HP & E-HP & FG-ARI & mBO & P-ARI & PQ \\
            \midrule
            \DINO{} ViT-B/16   & \xmark & \xmark & \xmark & 40.3 & 27.2 & 37.1 & 14.4 \\
            \midrule
            \DINO{}v2 ViT-S/14 & \xmark & \xmark & \xmark & 42.5 & 28.8 & 39.5 & 16.3 \\
            & \xmark & \cmark & \xmark & 42.9 & 29.1 & 39.8 & 16.8 \\
            & \cmark & \xmark & \xmark & 46.5 & 29.8 & 42.2 & 17.9 \\
            & \cmark & \cmark & \xmark & 48.0 & 30.6 & 42.8 & 18.8 \\
            & \cmark & \cmark & \cmark & 48.5 & 30.7 & 42.6 & 19.0 \\
            +Top-k & \cmark & \cmark & \cmark & 46.4 & 32.0 & 43.5 & 19.5 \\
            +Top-k, +Hi-Res & \cmark & \cmark & \cmark & 46.6 & 35.6 & 49.6 & 23.6  \\
            \bottomrule
        \end{tabular}%
    }
    \end{minipage}%
    \hfill%
    \begin{minipage}{0.385\textwidth}
    {
        \centering \vspace*{-1.75em}  %
        \newcommand{\myig}[1]{\includegraphics[width=0.19\textwidth,valign=c]{images/#1}}
        \newcommand{\mycaption}[1]{{\scriptsize#1}}
        \renewcommand{\arraystretch}{2.5}
        \setlength{\tabcolsep}{1pt}
        \begin{tabular}{@{}ccccc@{}}
            \mycaption{\DINO{}} & \mycaption{+\DINO{}v2} & \mycaption{+FT} & \mycaption{+Hi-Res} & \\[-0.7em]
            \myig{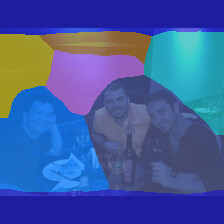} &
            \myig{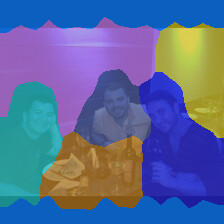} & \myig{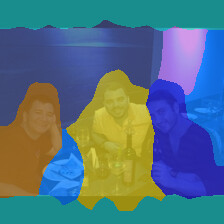} &
            \myig{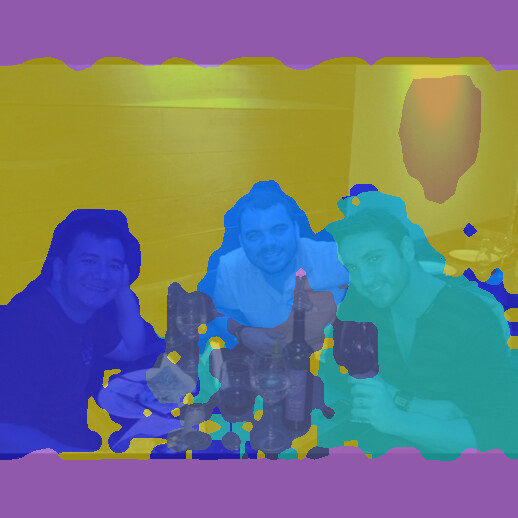} &
            \myig{images/entityseg/001} \\
            \myig{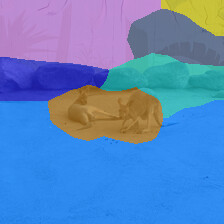} &
            \myig{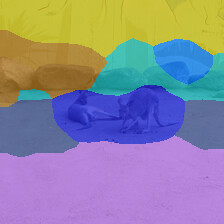} & \myig{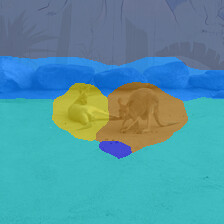} &
            \myig{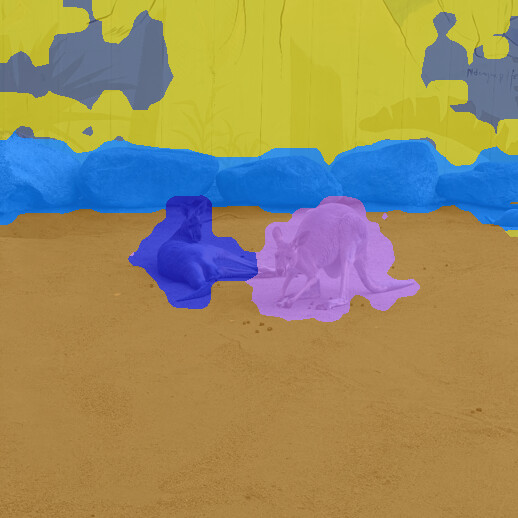} & 
            \myig{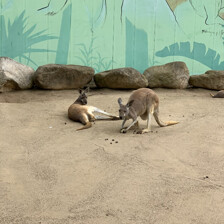} \\
            \myig{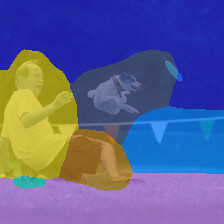} &
            \myig{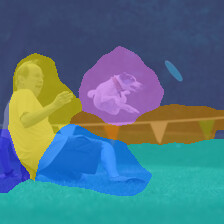} & \myig{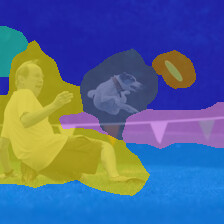} &
            \myig{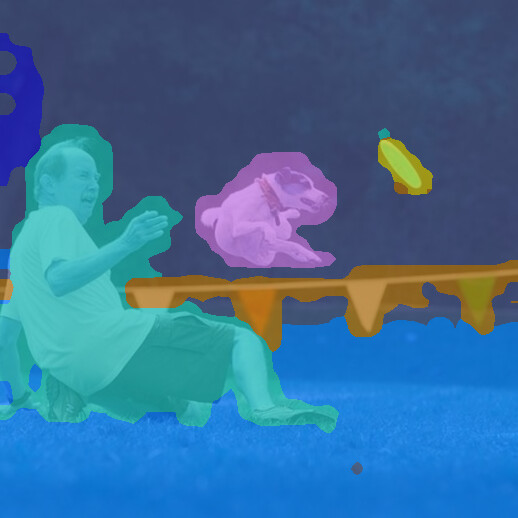} & 
            \myig{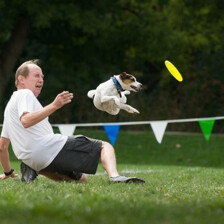} \\
        \end{tabular}
    }
    \end{minipage}
\end{table}

In \cref{tab:method-ablation-coco}, we analyze the contribution of different components of our model on the \COCO dataset, starting from the original \DINOSAUR model and ending with our final model.
First, we find that \emph{switching from \DINO{} to \DINO{}v2} leads to moderate improvements (+2.2 FG-ARI and +1.6 mBO). 
Adding \emph{finetuning} results in a strong improvement of FG-ARI (+4.0), demonstrating the importance of task-specific adaptation.
To evaluate our \emph{hyperparameter changes}, we split them into two groups: general hyperparameters (cosine schedule, lower learning rate, lower gradient clipping), and encoder hyperparameters (blockwise learning rates, lower encoder learning rate, encoder weight decay).
The changes to the general hyperparameters result in moderate improvements (+1.5 FG-ARI, +0.8 mBO, +0.9 PQ), with the changed encoder hyperparameters contributing further small improvements. 
Introducing \emph{top-k decoding} reduces FG-ARI (-2.1), but increases the other metrics (\eg +1.3 mBO).
Finally, \emph{high-resolution adaptation} results in further strong boosts (+3.6 mBO, +6.1 P-ARI, +4.1 PQ).\looseness-1

\newpage
\section{Evaluation}
\label{sec:evaluation}

To evaluate our approach, we use our benchmark to answer the following three questions:
{\setlength{\topsep}{0.0em}
\begin{itemize}[left=1em]\setlength{\itemsep}{0.0em}
    \item How does our proposed finetuning methodology work on diverse datasets (\cref{subsec:eval-in-distribution})?
    \item How does our method compare to prior methods for real-world object-centric learning (\cref{subsec:eval-comparison-prior-work-coco})?
    \item How does our method perform on the introduced zero-shot benchmark (\cref{subsec:eval-zero-shot})?
\end{itemize}}

\subsection{Evaluation of Object-Centric Finetuning}
\label{subsec:eval-in-distribution}
\begin{figure}
    \begin{minipage}[t]{0.48\textwidth}
        \centering
        \vspace{0pt}
        \includegraphics{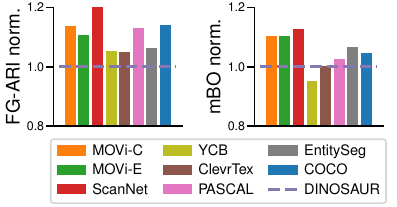}
        \caption{
            \textbf{Normalized performance when adding \emph{finetuning} to \DINOSAUR for \emph{in-distribution} training}, using a ViT-S/14 \DINO{}v2 encoder.
            Finetuning shows strong gains on all datasets.
            Numerical results in \cref{app:tab:eval-finetuning-in-distribution}.
        }
        \label{fig:eval-finetuning-in-distribution}
    \end{minipage}%
    \hfill%
    \begin{minipage}[t]{0.48\textwidth}
        \small
        \vspace{0pt}
        \centering
\setlength{\tabcolsep}{3pt}
\begin{tabular}{@{}l@{\hskip1ex}cccc@{}}
    \toprule
    Model & FG-ARI & mBO & P-ARI & PQ \\
    \midrule
    \DINOSAUR~\citep{seitzer2023bridging} & 40.5 & 27.7 & 37.1 & 14.4 \\
    SlotDiffusion$^\dagger$~\citep{wu2023slotdiffusion} & 37.3 & 31.4 & 47.6 & 21.0 \\
    \SPOT$^\dagger$~\citep{kakogeorgiou2023spot} & 37.0 & 34.8 & \textbf{52.4} & 21.3 \\
    \methodshort{} & \textbf{48.8} & \textbf{36.3} & 49.4 & \textbf{23.9} \\
    \midrule
    \textcolor{samcolor}{\SAMcomp{}$^\dagger$}~\citep{kirillov2023segment} & 12.1 & 19.0 & 10.8 & 9.4 \\
    \textcolor{samcolor}{\SAMbest{}$^\dagger$}~\citep{kirillov2023segment} & 44.9 & 56.9 & 54.4 & 10.9 \\
    \bottomrule
\end{tabular}
\captionof{table}{
    \textbf{Comparison to prior work on \COCO.}
    We use a ViT-B/14 encoder with top-k decoding and hi-res.~adaptation.
    Results for \mbox{(FT-)}\textsc{Dinosaur} averaged over 3 seeds.
    Results marked $\dagger$ evaluate official checkpoints, supervised models in gray.
    We compare to more baselines in \cref{app:tab:eval-comparison-prior-work-coco-extended}.
}
\label{tab:eval-comparison-prior-work-coco}

    \end{minipage}
\end{figure}

We first validate our proposed finetuning approach as a general methodology by training on diverse datasets.
In particular, we train a \DINOSAUR model using a \DINO{}v2 backbone with and without finetuning on the training splits of all 8 datasets included in our zero-shot benchmark, and evaluate \emph{in-distribution}.
The results are listed in \cref{fig:eval-finetuning-in-distribution}.
We find that adding finetuning results in strong improvements on both FG-ARI (up to 11 points) and mBO (up to 5 points) across all 8 datasets.
First, this demonstrates that finetuning, when using our training recipe, is a general strategy to improve performance of slot attention-based object-centric models with pre-trained backbones.
This is in contrast to~\citet{seitzer2023bridging}'s findings, who reported collapsing slots when finetuning the pre-trained ViT encoder.
Second, this shows that while pre-trained features obtained from self-supervised methods like \DINO{}v2 are powerful, it is possible to improve upon them with task-specific finetuning.
Interestingly, even though the model's objective is to \emph{predict} \DINO{}v2 features, the optimal input to slot attention are \emph{not} those exact features.
Following our analysis in \cref{subsec:method-analysis}, we conjecture that finetuning adapts the features to simplify grouping under the inductive biases of the model.

\subsection{Comparison to Prior Work on Real-World Object-Centric Learning}
\label{subsec:eval-comparison-prior-work-coco}

Second, in \cref{tab:eval-comparison-prior-work-coco}, we compare our full approach with prior work on the \COCO dataset. 
See \cref{app:fig:examples-coco} for example predictions.
We find that our method sets a new state-of-the-art on \COCO, achieving better results than all previous unsupervised object-centric methods, except being slightly worse than \SPOT on the panoptic ARI metric.
Moreover, our method also outperforms the \SAMcomp{} baseline (ViT-Base encoder, same number of masks) on all metrics.
In particular, our method has strongly improved FG-ARI (+9), indicating much better object discovery capabilities --- it even achieves higher FG-ARI than the \SAMbest{} baseline (ViT-Huge encoder, variable number of masks).
However, there is still a large gap to \SAM's mBO, which we attribute to 1) \SAM's generally higher mask quality, and 2) its ability to capture a variable number of objects, which in particular leads to finding more small objects. 

\subsection{Zero-Shot Evaluation}
\label{subsec:eval-zero-shot}
\begin{figure}
    \begin{minipage}[t]{0.48\textwidth}
        \centering
        \includegraphics{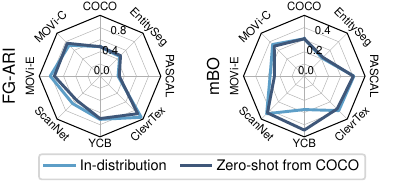}
        \caption{
            \textbf{Comparing \emph{in-distribution training} \versus \emph{zero-shot transfer} from \COCO for our finetuning approach.}
            Overall, performance is similar.
            Numerical results in \cref{app:tab:eval-finetuning-in-distribution}.
        }
        \label{fig:eval-comparison-id-vs-ood}
    \end{minipage}%
    \hfill%
    \begin{minipage}[t]{0.48\textwidth}
        \centering
        \includegraphics{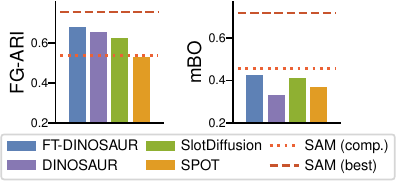}
        \caption{
            \textbf{Zero-shot performance averaged over datasets.}
            \methodshort performs best both in FG-ARI and mBO.
            Results per datasets available in \cref{app:tab:eval-comparison-ood}.}
        \label{fig:eval-comparison-ood-average}
    \end{minipage}
\end{figure}

Finally, we evaluate our method in terms of its zero-shot performance.
First, in \cref{fig:eval-comparison-id-vs-ood}, we compare the zero-shot performance of our model finetuned \emph{on \COCO} (without top-k decoding or high-res.~adaptation) to the performance of our model finetuned \emph{in-distribution}. 
We find that transferring from \COCO yields comparable results to training in-distribution on most datasets (\ScanNet, \PASCALVOC, \EntitySeg), and even surpasses \emph{in-distribution} training on some datasets (\MOVi-C, \YCB) --- surprisingly, object-centric finetuning does not hurt generalization (\eg by overfitting), indicating that it adapts the model to the \emph{task} rather than the \emph{data}.
Overall, this shows that task-specific finetuning on diverse real-world data is a viable path to obtain zero-shot object-centric models.

Second, in \cref{fig:eval-comparison-ood-average}, we compare the zero-shot performance of our full model (including top-k decoding and high-resolution adaptation) to prior work. 
Averaged over all datasets, our approach achieves both the highest FG-ARI and mBO, while previous work generally trades off high FG-ARI with low mBO (\DINOSAUR), or high mBO with low FG-ARI (SlotDiffusion, \SPOT). 
On top of finetuning, we ascribe this to our usage of the MLP decoder (higher FG-ARI) in combination with high-resolution training (higher mBO). 

Last, we compare our model to \SAM.
\SAMcomp{} generally performs \emph{worse than our model}, showing the difficulty of unsupervised scene decomposition in the absence of task-specific information.
\SAMbest{} achieves an FG-ARI of 76.1, compared to 67.8 for our approach.
In terms of mBO, there is still a large difference between \SAM and our approach (42.5 \versus 73.2).
Taken together, these results show that unsupervised object-centric models are \emph{closing the gap to supervised methods in terms of zero-shot object discovery}.
This is astonishing, given that \SAM was trained on 10 million images with over 1 billion mask annotations.
Moreover, a principal advantage of object-centric models over \SAM is that they come equipped with explicit object representations.
While mask quality as measured by mBO is lacking behind \SAM, we are hopeful that this gap is addressable by training on even higher resolution images and introducing innovations for variable number of slots. 
We present a comparison of the masks obtained from the proposed approach and all baselines in \cref{app:sec:examples}.

\section{Conclusion}
\label{sec:conclusion}

In this work, we have introduced a benchmark of diverse real-world and synthetic datasets to study the zero-shot capabilities of object-centric representation learning models. 
Our findings indicate that object-centric models using pre-trained encoders already exhibit notable zero-shot capabilities when trained on real-world data.
We then presented a finetuning procedure for adapting pre-trained encoders to the task of object discovery, demonstrating that this approach achieves state-of-the-art results across 8 datasets in both in-distribution and out-of-distribution scenarios.
We believe that our contributed tools --- the zero-shot benchmark and stable finetuning --- are important stepping stones towards an object-centric foundation model.

Our benchmark showed the importance of the type of training data for zero-shot transfer.
Our experiments indicate that training on complex natural data is important, suggesting an exciting direction to design curated datasets for zero-shot object-centric learning.
Moreover, our benchmark revealed that current object-centric models are highly sample-efficient but fail to leverage larger datasets to improve performance at current model sizes.
This result is significant because it suggests that, unlike other deep learning domains, stronger object-centric models cannot be achieved simply by scaling up data alone. 
We hope our findings will encourage the community to develop object-centric models that scale effectively with both data and model size.

For general-purpose object-centric models, an important property is the usefulness of the learned object-centric representation for downstream tasks. 
While downstream applicability has been explored in various forms~\citep{yoon2023investigation, wu2023slotformer, wu2023slotdiffusion, Xu2024SlotVLMSS,mamaghan2024exploringVQA}, the zero-shot scenario has not been comprehensively studied so far.
An exciting direction for future work is to extend our benchmark to include zero-shot downstream tasks and to consider other dimensions of scaling.

\clearpage
\begin{ack}
This work was supported by the ERC - 101045454 REAL-RL and funded by EXC number 2064/1 – Project number 390727645.
We acknowledge the support from the German Federal Ministry of Education and Research (BMBF) through the Tübingen AI Center (FKZ: 01IS18039B).
Andrii Zadaianchuk is funded by the European Union (ERC, EVA, 950086). Views and opinions expressed are, however, those of the author only and do not necessarily reflect those of the European Union or the European Research Council Executive Agency. Neither the European Union nor the granting authority can be held responsible for them.
The authors thank the International Max Planck Research School for Intelligent Systems (IMPRS-IS) for supporting Maximilian Seitzer. 
Aniket Didolkar would like to thank Mila for providing part of the compute resources used in this work. 
\end{ack}

\section*{Contributions}
This project was initiated by AD and MS, and MS had the role of project lead.
AD and MS contributed equally.
AZ joined the project from the start and had critical input at all stages.
AD, AZ, and MS shaped the project direction, with advise from AG, MM, YB, and GM.
MS implemented most of the code, with contributions from AD.
AD and MS performed the exploratory experiments.
AD performed most of the final experiments and evaluations, with some experiments ran by MS.
AZ performed the analysis of encoder features and created the corresponding figures (\cref{fig:analysis}, \cref{app:fig:analysis-appendix}), AD created the model figure (\cref{fig:method}), and MS created the remaining figures.
The first draft was written by AD, AZ and MS, with AG and GM contributing to the final version.

\bibliography{main}

\clearpage
\appendix

\setcounter{figure}{0}
\renewcommand\thefigure{\thesection.\arabic{figure}}
\setcounter{table}{0}
\renewcommand\thetable{\thesection.\arabic{table}}

{
    \begin{center}
        \LARGE
        \textsc{Appendix}
    \end{center}
    \FloatBarrier
}

\section{Additional Experiments}
\label{app:sec:additional-experiments}

\subsection{Zero-Shot Benchmark}

\begin{figure}
    \centering
    \begin{subfigure}[t]{0.333\textwidth}
        \centering
        \includegraphics{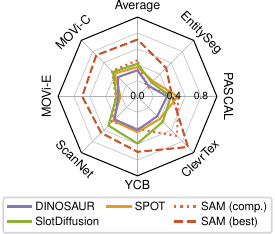}
        \caption{Varying models.}
    \end{subfigure}%
    \hfill%
    \begin{subfigure}[t]{0.333\textwidth}
        \centering
        \includegraphics{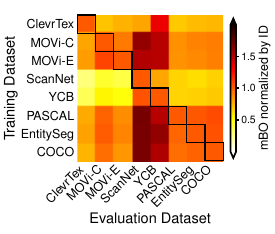}
        \captionsetup{width=.8\linewidth}
        \caption{Varying train datasets.}
    \end{subfigure}%
    \hfill%
    \begin{subfigure}[t]{0.333\textwidth}
        \centering
        \includegraphics{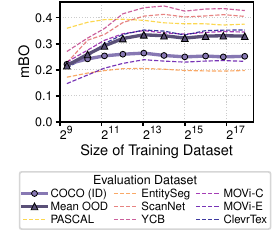}
        \captionsetup{width=.8\linewidth}
        \caption{Varying train dataset size.}
    \end{subfigure}
    \caption{Evaluating zero-shot transfer of object-centric representations. Corresponds to \cref{fig:benchmark}, but shows mBO instead of FG-ARI.}
    \label{app:fig:benchmark-mbo}
\end{figure}

We show additional results complementary to the results in the main part.
\Cref{app:fig:benchmark-mbo} shows benchmark results in terms of varying models, training data distribution, and training dataset size, but with the mBO metric instead of the FG-ARI metric.
The results largely mirror those in \cref{fig:benchmark}; it can be seen that \DINOSAUR generally has worse mBO than \SPOT and SlotDiffusion, whereas with FG-ARI, this trend is reversed.

\begin{figure}
    \centering
    \begin{subfigure}[t]{\textwidth}
        \includegraphics[width=0.333\textwidth]{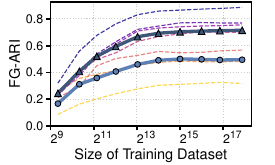}\hfill%
        \includegraphics[width=0.333\textwidth]{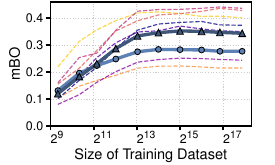}\hfill%
        \includegraphics[width=0.333\textwidth]{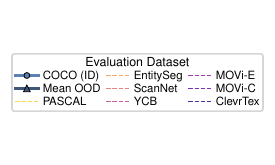}
        \caption{Scaling behavior of \method, showing OOD performance on individual datasets.}
        \label{app:subfig:benchmark-dataset-size-ours-individual}
    \end{subfigure}

    \vspace{0.5em}
    \begin{subfigure}[t]{\textwidth}
        \includegraphics[width=0.333\textwidth]{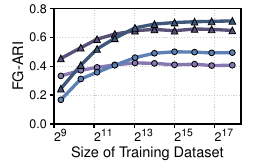}\hfill%
        \includegraphics[width=0.333\textwidth]{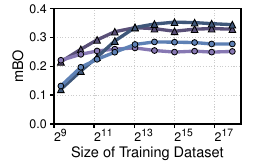}\hfill%
        \includegraphics[width=0.333\textwidth]{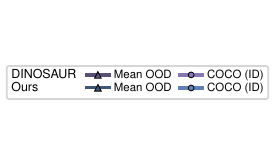}
        \caption{Scaling behavior of \method \versus \DINOSAUR.}
        \label{app:subfig:benchmark-dataset-size-ours-vs-dinosaur}
    \end{subfigure}
    \caption{Scaling behaviour of \method on trained on differently sized subsets of \COCO. 
    Our method uses a ViT-S/14 with \DINO{}v2 with finetuning, but no top-k decoding and hi-res adaptation.}
    \label{app:fig:benchmark-dataset-size-ours}
\end{figure}

\Cref{app:fig:benchmark-dataset-size-ours} shows the data scaling behavior of our \method method trained on different subsets of the \COCO dataset, showing performance on the individual datasets in \cref{app:subfig:benchmark-dataset-size-ours-individual}, and comparing the aggregated performance to \DINOSAUR in \cref{app:subfig:benchmark-dataset-size-ours-vs-dinosaur}.
While \DINOSAUR is better in the very-low sample regime (less than ~\np{5000} samples), \method overall shows better scaling behavior.
In particular, \method exhibits a slightly upward trending scaling curve for OOD evaluation with FG-ARI; while the effect is too weak to conclude that \method scales well with data, it would be interesting to extend this experiment to include 1--2 magnitudes more data.

\subsection{Object-Centric Finetuning}
\begin{wrapfigure}[8]{r}{0.35\textwidth}
    \vspace{-\intextsep}
    \captionof{table}{Analysis of targets. $\tau$ is the momentum for teacher updates.}
    \label{app:tab:method-analysis-targets}
    \vspace{-.8em}
    \small
    \begin{center}
        \begin{tabular}{@{}rcc@{}}
            \toprule
            $\tau$ & FG-ARI & mBO \\
            \midrule
            $0$ & 0.09 & 13.6 \\
            $0.999$ & 23.6 & 16.4 \\
            $0.9999$ & 37.8 & 21.0 \\
            $1$ & 48.5 & 30.7 \\
            \bottomrule
        \end{tabular}
    \end{center}
\end{wrapfigure}
\paragraph{Targets from EMA teacher}
\label{app:subsec:ema}
We can also frame our setup as a variant of the student-teacher framework common in self-supervised methods~\citep{Grill2020BYOL,Caron2021DINO,oquab2023dinov2}.
There, the weights of the teacher model are continuously updated from the student's weights through an exponential moving average (EMA), with a momentum parameter $\tau \in [0, 1]$ controlling the speed of adaptation.
Through this lens, our approach uses $\tau=1$, corresponding to not updating the teacher.
This view suggests to use $\tau < 1$ to improve the targets throughout training.

In \cref{app:tab:method-analysis-targets}, we analyze the effect of introducing student-teacher style EMA updates.
Directly using the features of the student as the targets ($\tau=0$) leads to collapse, as reported previously~\citep{seitzer2023bridging}.
With momentum updates, we still find a high value for $\tau$ to be necessary to stabilize training.
Using fixed targets ($\tau=1$) gives the best results.
We speculate this is because there is no missing information in the auto-encoder setup, leading to a gradual loss of information.

\begin{figure}
\centering
    \includegraphics[width=0.8\textwidth]{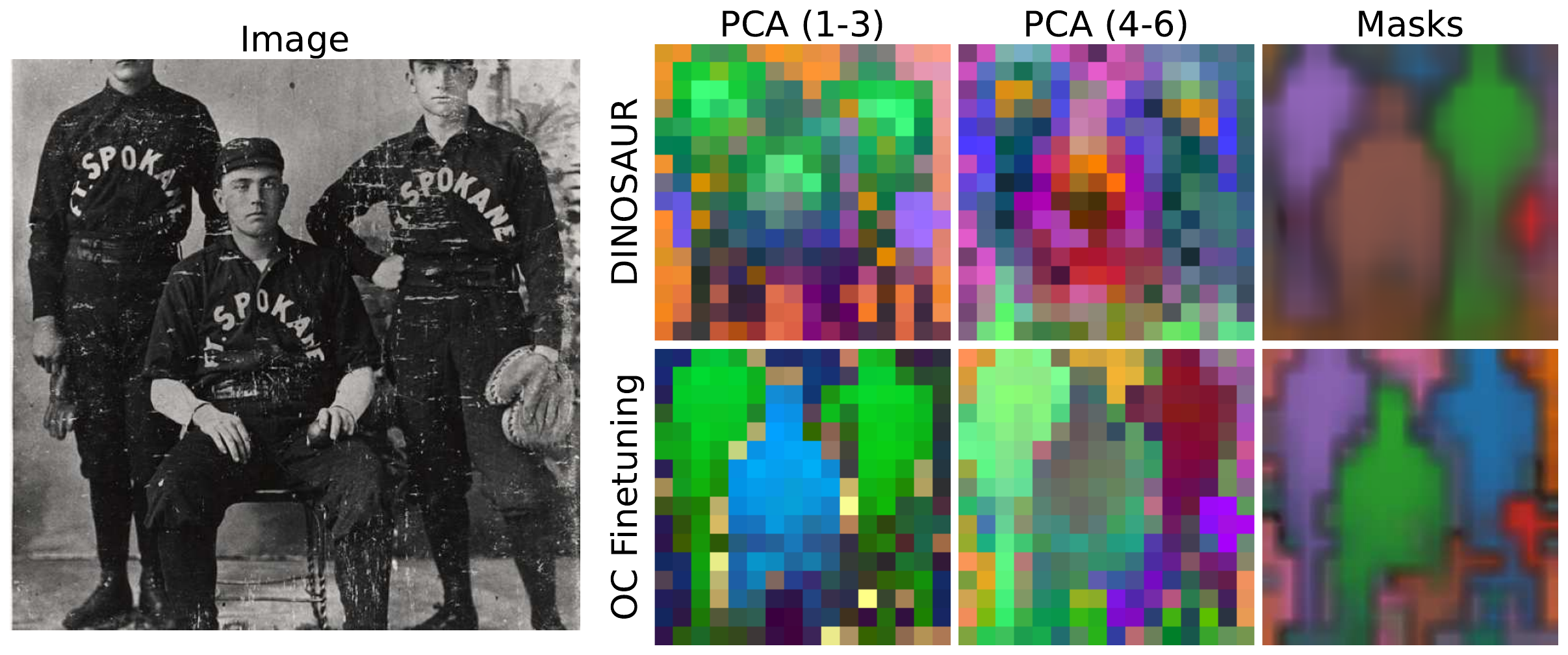}
    \includegraphics[width=0.8\textwidth]{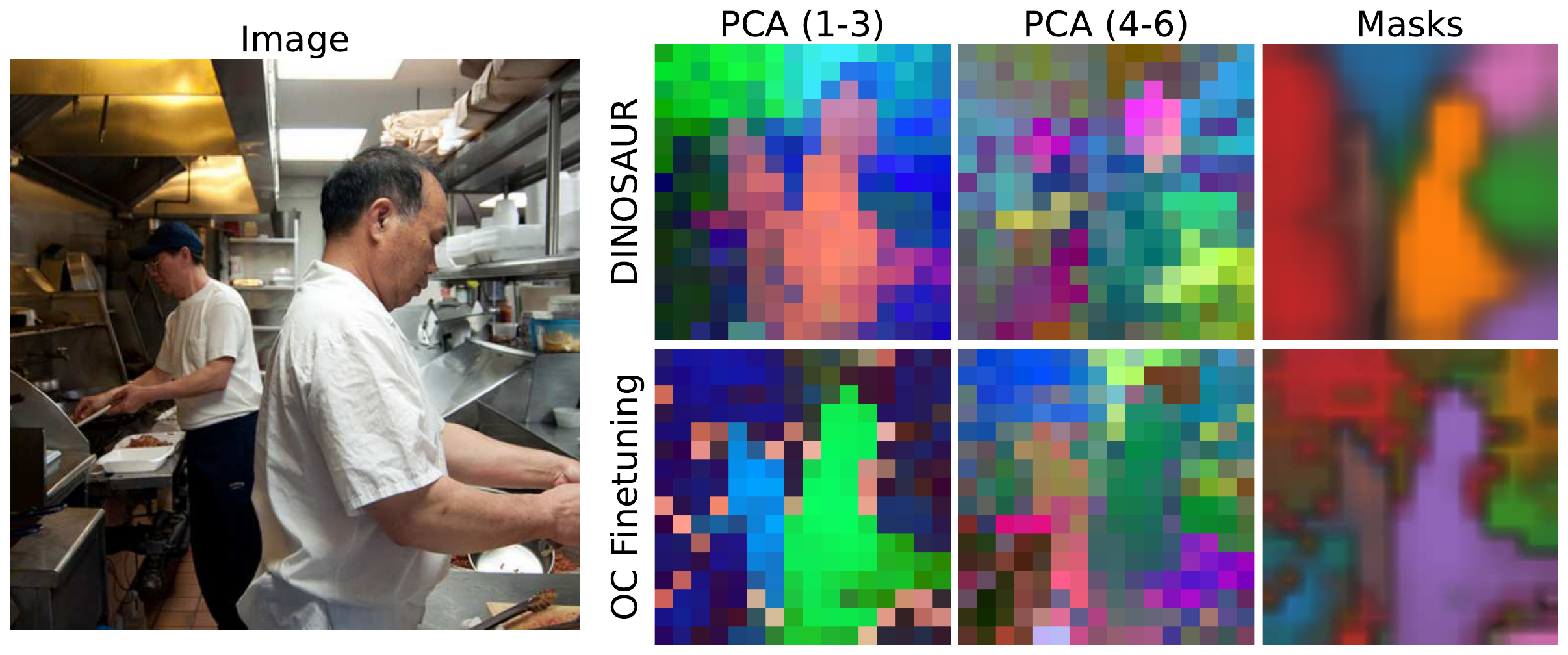}
    \includegraphics[width=0.8\textwidth]{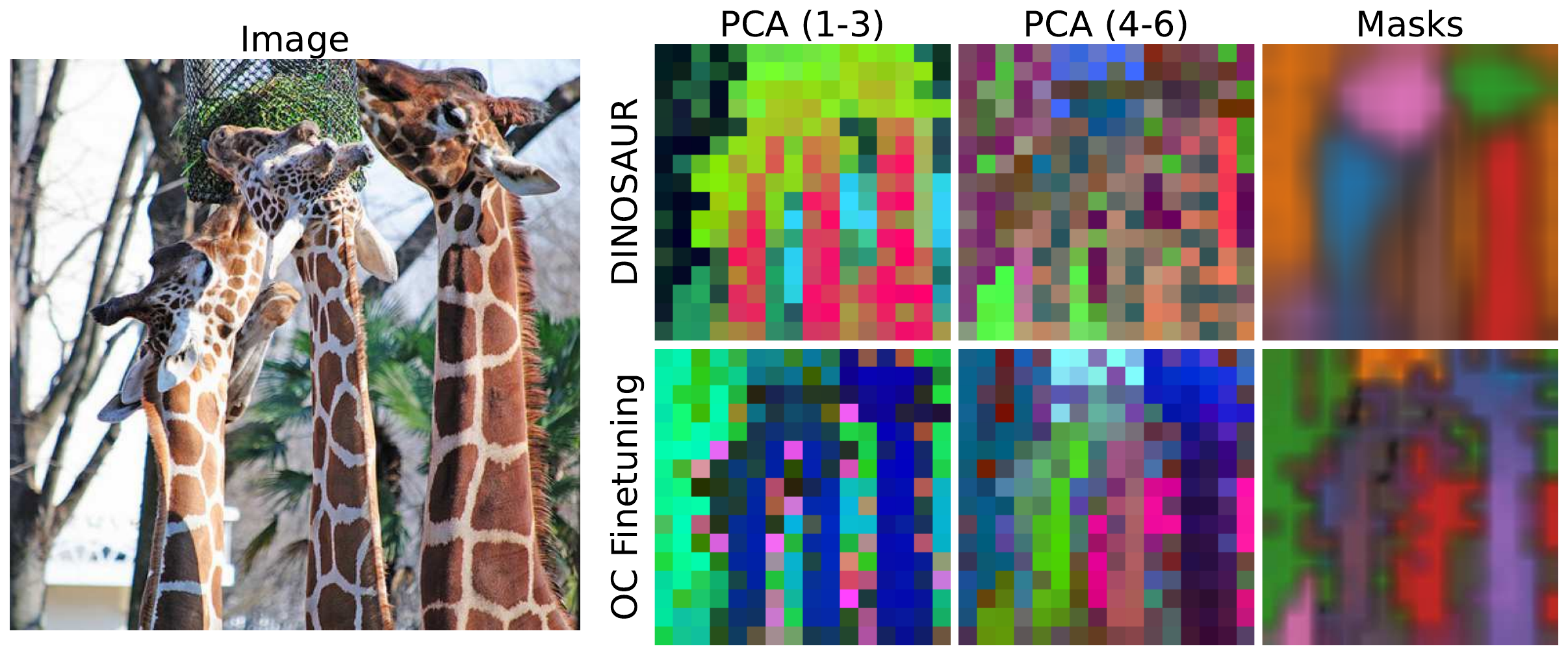}
    \includegraphics[width=0.8\textwidth]{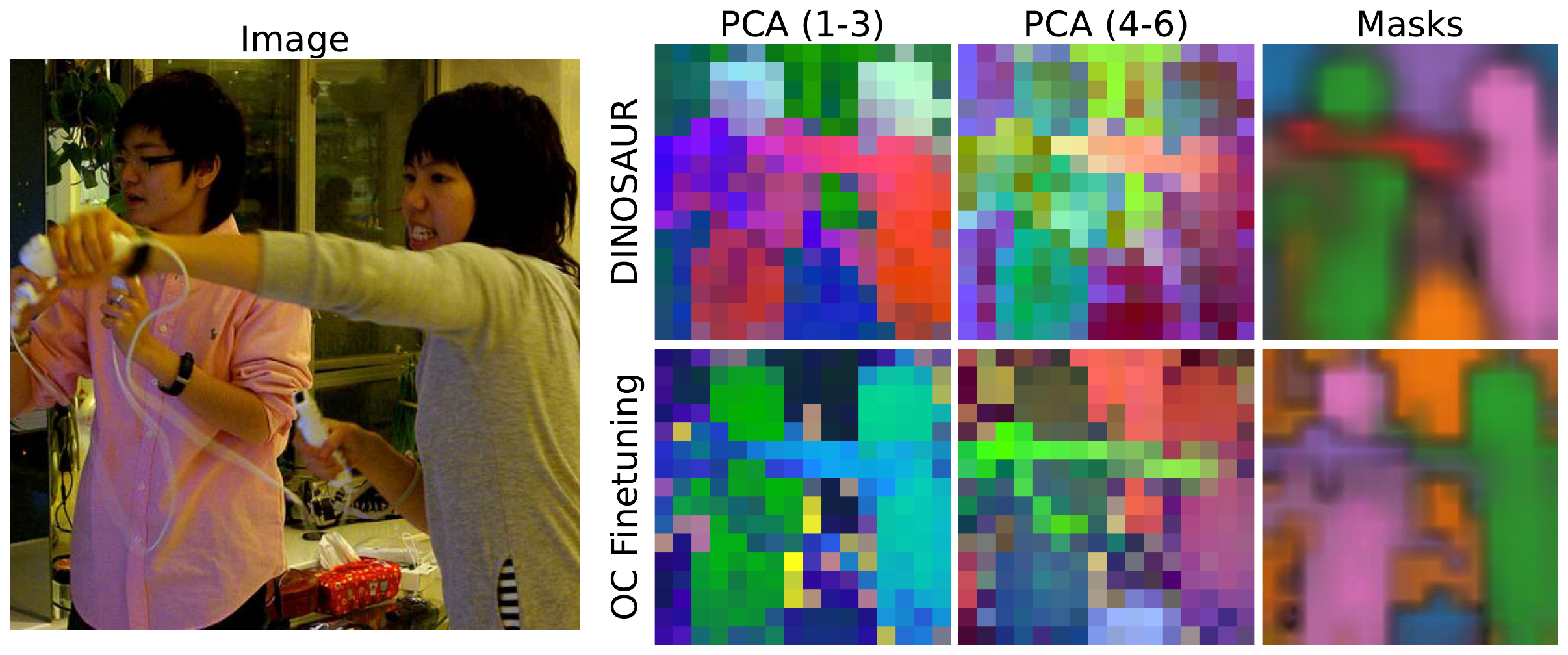}
    \caption{Visualization of encoder features in \DINOSAUR (frozen \DINO{}v2 features) and for encoder features adapted with object-centric finetuning, similar to \cref{fig:analysis} in the main paper.
    The second column shows 1st to 3rd PCA components, and the third column shows 4th to 6th PCA components grouped in one image by using different RGB channels. 
    The last column shows object discovery masks by each method.
    }
    \label{app:fig:analysis-appendix} 
\end{figure}

\paragraph{Analysis of Finetuned Features}
In \cref{app:fig:analysis-appendix}, we show additional examples for visualizing the PCA on the finetuned features compared to \DINO{}v2 features (similar to \cref{fig:analysis}).
Similar to the discussion in \cref{subsec:method-analysis}, we find that after finetuning, the encoder features are noticeably more object-centric.
For example, in the first and last examples, \DINO{}v2 features show a part-based split of the shown persons in the dominant PCA components; the finetuned features highlight the whole persons better.
In the second example, \DINO{}v2 features group semantic instances (human) together in the dominant components; after finetuning, the features clearly split the persons. 
However, note that is not necessary that the features highlight the instances in the dominant components to derived an instance-based grouping; in all examples, the masks discovered by \DINOSAUR (last column) feature a correct instance split (while also splitting further into parts in the last two examples).
This may be because the necessary information for the correct split is contained in the less dominant components of the features (\eg in PCA dimensions 4--6).
However, we conjecture that the finetuned features simplify the grouping task for slot attention, leading to better and more consistent object discovery.

\subsection{Evaluation}

We include the following additional results for the evaluation of \method conducted in \cref{sec:evaluation} of the main paper:
\begin{itemize}
 \item \textbf{Finetuning In-Distribution (\cref{subsec:eval-in-distribution})}: we show the numeric values corresponding to \cref{fig:eval-finetuning-in-distribution} in the main part in \cref{app:tab:eval-finetuning-in-distribution}.
 This table also shows the results for the in-distribution \versus zero-shot comparison in \cref{fig:eval-comparison-id-vs-ood}.
 \item \textbf{Extended Comparison To Prior Work on Real-World Object-Centric Learning (\cref{subsec:eval-comparison-prior-work-coco})}: we conduct an extended comparison to prior work for real-world object-centric learning on the \COCO dataset in \cref{app:tab:eval-comparison-prior-work-coco-extended}.
 \item \textbf{Zero-Shot Evaluation (\cref{subsec:eval-zero-shot})}: we show the full results over all datasets of the zero-shot benchmark, corresponding to \cref{fig:benchmark-methods} and \cref{fig:eval-comparison-ood-average} in the main part in \cref{app:tab:eval-comparison-ood}.
\end{itemize}

\begin{table}
    \caption{Evaluation of adding \emph{finetuning} to \DINOSAUR when training \emph{in-distribution}, using a ViT-S/14 DINOv2 backbone.
    Finetuning shows strong performance improvements on all eight datasets. 
    We also show zero-shot transfer when finetuning on \COCO, which performs comparable or better to training in-distribution on 5 out of 7 datasets. %
    Results corresponding to experiment in \cref{fig:eval-finetuning-in-distribution} in the main paper.
    }
    \label{app:tab:eval-finetuning-in-distribution}
    \small\vspace{.3em}
    \centering
    \begin{tabular}{@{}lcc ccc ccc ccc@{}}
        \toprule
        & \multicolumn{2}{c}{\textbf{\MOVi-C}} & {} & \multicolumn{2}{c}{\textbf{\MOVi-E}} & {} & \multicolumn{2}{c}{\textbf{\ScanNet}} & {} & \multicolumn{2}{c}{\textbf{\YCB}} \\
        \cmidrule{2-3} \cmidrule{5-6} \cmidrule{8-9} \cmidrule{11-12}
        & FG-ARI & mBO & \phantom & FG-ARI & mBO  & \phantom & FG-ARI & mBO  & \phantom & FG-ARI & mBO \\
        \midrule
        \DINOSAUR & 63.1 & 33.3 && 74.0 & 25.5 && 52.8 & 38.8 && 67.5 & 28.9 \\
        +Finetuning & 71.9 & 36.8 && 82.0 & 28.1 && 63.8 & 43.8 && 71.1 & 27.5 \\
        \midrule
        Zero-shot (\COCO) & 76.4 & 34.9 && 75.1 & 24.5 && 55.4 & 42.8 && 69.9 & 44.4  \\
        \bottomrule
        \\[-0.5em]
        \toprule
        & \multicolumn{2}{c}{\textbf{\ClevrTex}} & {} & \multicolumn{2}{c}{\textbf{\PASCALVOC}} & {} & \multicolumn{2}{c}{\textbf{\EntitySeg}} & {} & \multicolumn{2}{c}{\textbf{\COCO}} \\
        \cmidrule{2-3} \cmidrule{5-6} \cmidrule{8-9} \cmidrule{11-12}
        & FG-ARI & mBO & \phantom & FG-ARI & mBO  & \phantom & FG-ARI & mBO  & \phantom & FG-ARI & mBO \\
        \midrule
        \DINOSAUR & 91.3 & 40.1 && 26.2 & 40.1 && 43.0 & 19.6 && 42.4 & 29.4 \\
        +Finetuning & 95.8 & 40.1 && 29.7 & 40.8 && 45.7 & 20.9 && 48.5 & 30.7  \\
        \midrule
        Zero-shot (\COCO) & 87.5 & 38.2 && 31.7 & 40.4 && 47.7 & 21.8 && 48.5 & 30.7 \\
        \bottomrule
    \end{tabular}
\end{table}

\begin{table}
    \caption{Extended comparison of our method (\methodshort{}) to prior work on the \COCO dataset, corresponding to \cref{tab:eval-comparison-prior-work-coco} in the main paper.
    For our proposed approach, we average results across 5 seeds for \methodshort{}, ViT-S/14 and across 3 seeds for \methodshort{}, ViT-B/14.
    Results marked with $\dagger$ are from evaluating official checkpoints; results marked with $\ast$ are taken from the respective papers.
    Supervised models (\textsc{Sam}) colored in gray.}
    \label{app:tab:eval-comparison-prior-work-coco-extended}
    \centering
    \begin{tabular}{@{}l@{\hskip1ex}cccc@{}}
        \toprule
        Model & FG-ARI & mBO & P-ARI & PQ \\
        \midrule
        Slot Attention~\citep{Locatello2020SlotAttention}, \citep{wu2023slotdiffusion}$^\ast$ & 21.4 & 17.2 & & \\
        \textsc{Slate}~\citep{Singh2022SLATE}, \citep{wu2023slotdiffusion}$^\ast$ & 32.5 & 29.1 & & \\ 
        \DINOSAUR (MLP Dec.)~\citep{seitzer2023bridging} & 40.5 & 27.7 & 37.1 & 14.4 \\
        \DINOSAUR (TF.~Dec.)$^\ast$~\citep{seitzer2023bridging} & 34.1 & 31.6\\
        Stable-LSD$^\ast$~\citep{jiang2023object} & 35.0 & 30.4 \\
        SlotDiffusion$^\dagger$~\citep{wu2023slotdiffusion} & 37.3 & 31.4 & 47.6 & 21.0 \\
        \SPOT$^\dagger$~\citep{kakogeorgiou2023spot} & 37.0 & 34.8 & \textbf{52.4} & 21.3 \\
        \methodshort{}, ViT-S/14  & 46.6 & 35.6 & 49.7 & 23.5 \\
        \methodshort{}, ViT-B/14 & \textbf{48.8} & \textbf{36.3} & 49.4 & \textbf{23.9} \\
        \midrule
        \textcolor{samcolor}{\SAMcomp{}$^\dagger$}~\citep{kirillov2023segment} & 12.1 & 19.0 & 10.8 & 9.4 \\
        \textcolor{samcolor}{\SAMbest{}$^\dagger$}~\citep{kirillov2023segment} & 44.9 & 56.9 & 54.4 & 10.9 \\
        \bottomrule
    \end{tabular}
\end{table}

\begin{table}
    \caption{Per-dataset zero-shot performance, corresponding to \cref{fig:eval-comparison-ood-average} in the main paper.
    All unsupervised object-centric methods (\DINOSAUR, SlotDiffusion, \SPOT, \methodshort{}) are trained on the \COCO dataset.
    Furthermore, we compare with the supervised Segment Anything model (\SAM).
    Average is computed as a weighted average normalizing using the size of the evaluation datasets (\cf \cref{app:tab:dataset_properties}).
    Results marked $\dagger$ are from evaluating official checkpoints.
    Supervised models (\textsc{Sam}) colored in gray.
    }
    \label{app:tab:eval-comparison-ood}
    \small \vspace{.3em}
    \centering
    \setlength{\tabcolsep}{4pt}
    \begin{tabular}{@{}lcc ccc ccc ccc ccc ccc ccc ccc@{}}
        \toprule
        & \multicolumn{2}{c}{\textbf{\MOVi-C}} & {} & \multicolumn{2}{c}{\textbf{\MOVi-E}} & {} & \multicolumn{2}{c}{\textbf{\ScanNet}} & {} & \multicolumn{2}{c}{\textbf{\YCB}} \\
        \cmidrule{2-3} \cmidrule{5-6} \cmidrule{8-9} \cmidrule{11-12}
        & FG-ARI & mBO & \phantom & FG-ARI & mBO  & \phantom & FG-ARI & mBO  & \phantom & FG-ARI & mBO \\
        \midrule
        \DINOSAUR~\citep{seitzer2023bridging} & 67.0 & 34.5 && \textbf{71.1} & 24.2 && \textbf{57.4} & 40.8 && 60.2 & 42.2  \\
        SlotDiffusion$^\dagger$~\citep{wu2023slotdiffusion} & 66.9 & 43.6 && 67.6 & 26.4 && 52.0 & \textbf{51.7} && 62.5 & \textbf{59.2} \\
        \SPOT$^\dagger$~\citep{kakogeorgiou2023spot} & 63.0 & 40.8 && 47.8 & 21.5 && 48.6 & 43.2 && 52.9 & 45.1 \\
        \methodshort{} (ViT-S/14) & 71.3 & \textbf{44.2} && \textbf{71.1} & \textbf{29.9} && 54.8 & 48.4 && 67.4 & 54.5 \\
        \methodshort{} (ViT-B/14) & \textbf{73.3} & 42.9 && 69.7 & 27.9 && 55.8 & 48.6 && \textbf{70.1} & 54.1 \\
        \midrule
        \textcolor{samcolor}{\SAMcomp{}$^\dagger$}~\citep{kirillov2023segment} & 57.6 & 45.3 && 38.5 & 27.4 && 45.8 & 45.5 && 46.9 & 40.9 \\
        \textcolor{samcolor}{\SAMbest{}$^\dagger$}~\citep{kirillov2023segment} & 79.7 & 73.5 && 84.7 & 69.7 && 62.2 & 64.7 && 69.4 & 69.8 \\
        \bottomrule
        \\[-0.5em]
        \toprule
        & \multicolumn{2}{c}{\textbf{\ClevrTex}} & {} & \multicolumn{2}{c}{\textbf{\PASCALVOC}} & {} & \multicolumn{2}{c}{\textbf{\EntitySeg}} & {} & \multicolumn{2}{c}{\textbf{Average}} \\
        \cmidrule{2-3} \cmidrule{5-6} \cmidrule{8-9} \cmidrule{11-12}
        & FG-ARI & mBO & \phantom & FG-ARI & mBO  & \phantom & FG-ARI & mBO  & \phantom & FG-ARI & mBO \\
        \midrule
        \DINOSAUR~\citep{seitzer2023bridging} & 82.5 & 35.2 && 24.0 & 37.2 && 43.5 & 19.4 && 65.2 & 33.0 \\
        SlotDiffusion$^\dagger$~\citep{wu2023slotdiffusion} & 77.0 & 45.0 && 21.1 & 42.0 && 43.7 & 25.1 && 62.6 & 41.3 \\
        \SPOT$^\dagger$~\citep{kakogeorgiou2023spot} & 63.3 & 40.0 && 21.2 & \textbf{50.6} && 41.7 & 27.4 && 53.0 & 37.0 \\
        \methodshort{} (ViT-S/14) & \textbf{86.0} & \textbf{50.1} && 24 & 37.6 && 48.1 & 28.4 &&  67.8 & \textbf{42.5} \\
        \methodshort{} (ViT-B/14) & 83.9 & 45.9 && \textbf{25.9} & 37.8 && \textbf{49.7} & \textbf{29.0} && \textbf{68.0} & 40.8 \\
        \midrule
        \textcolor{samcolor}{\SAMcomp{}$^\dagger$}~\citep{kirillov2023segment} & 82.9 & 70.3 && 31.0 & 51.5 && 25.9 & 16.5 && 53.5 & 45.7 \\
        \textcolor{samcolor}{\SAMbest{}$^\dagger$}~\citep{kirillov2023segment} & 94.0 & 90.0 && 31.1 & 64.2 && 53.4 & 51.0 && 76.1 & 73.2 \\
        \bottomrule
    \end{tabular}
\end{table}

\section{Limitations}
\label{app:sec:limitations}
The proposed zero-shot benchmark and \method model have several limitations that we cover in this section.

\subsection{Benchmark Limitations} 
While our benchmark focuses on a broad range of datasets, including fully OOD images from synthetic data like \ClevrTex, and open-world natural images from the \EntitySeg dataset, object-centric representation could also be extracted from to non-natural domains such as medical, microscopy, or satellite imagery. 
Thus, extending our benchmark to include such domains to evaluate zero-shot transfer performance is an interesting future direction. 
Furthermore, in our current benchmark, we focus only on unsupervised scene decomposition and object discovery. 
Although this provides valuable insights into the models’ localization abilities, it does not fully evaluate the content of the learned representations. 
Hence, the benchmark could be extended by incorporating additional downstream tasks, such as object category and attribute prediction.

\subsection{Model Limitations} 
\begin{figure}
    \newcommand{\myig}[1]{\includegraphics[width=0.17\textwidth,valign=c]{images/#1}}
    \newcommand{\mycaption}[1]{{#1}}
    \renewcommand{\arraystretch}{5.2}
    \setlength{\tabcolsep}{2pt}
    \centering
    \begin{subfigure}[t]{\textwidth}
        \centering
        \begin{tabular}{@{}cccc@{}}
            \myig{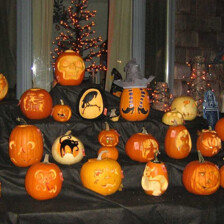} &
            \myig{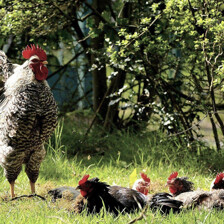} &
            \myig{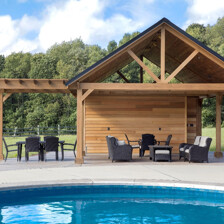} & 
            \myig{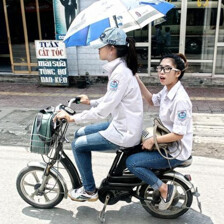} \\
            \myig{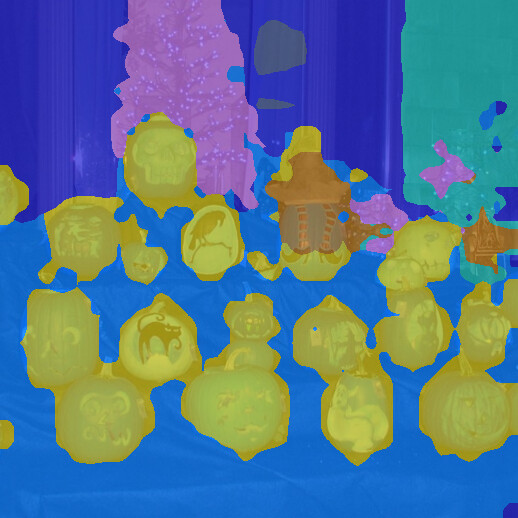} &
            \myig{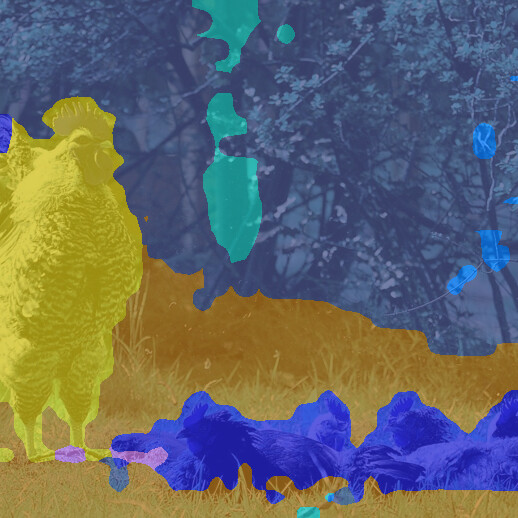} &
            \myig{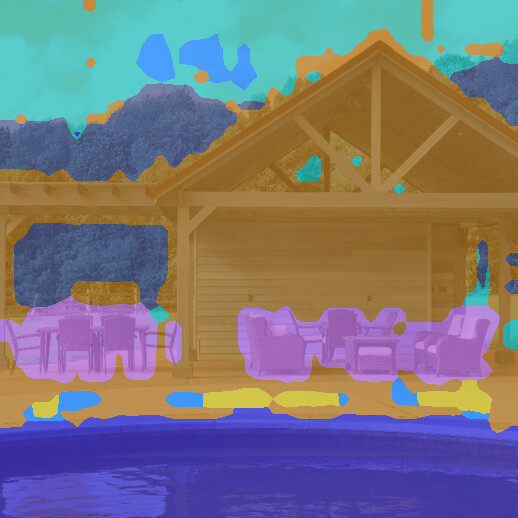} & 
            \myig{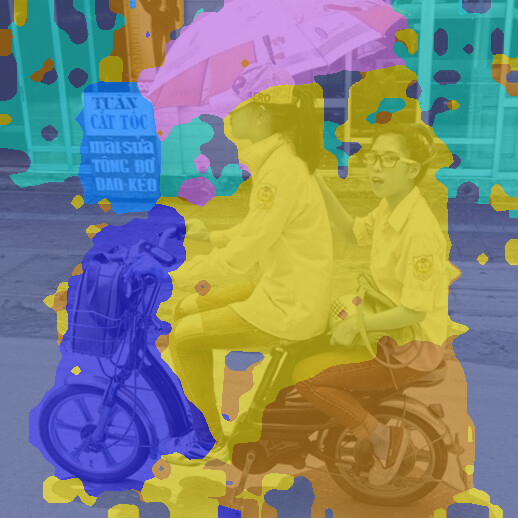}
        \end{tabular}
        \caption{Joining semantically-related objects together.}
        \label{app:subfig:examples-failures-overgrouping}
    \end{subfigure}
    
    \begin{subfigure}[t]{\textwidth}
        \centering
        \begin{tabular}{@{}cccc@{}}
            \myig{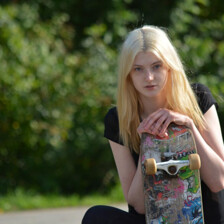} &
            \myig{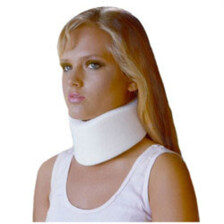} &
            \myig{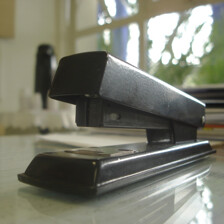} & 
            \myig{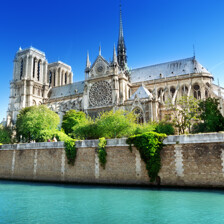} \\
            \myig{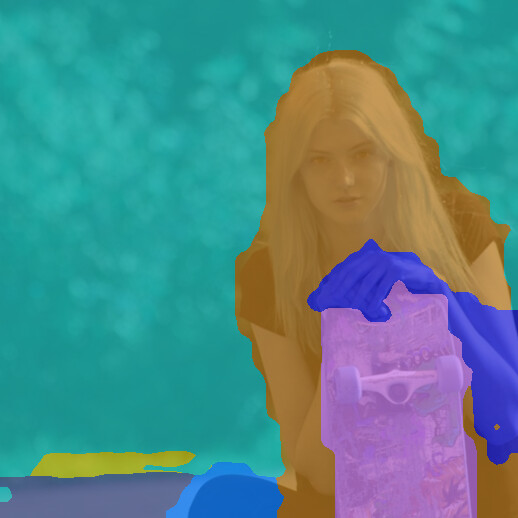} &
            \myig{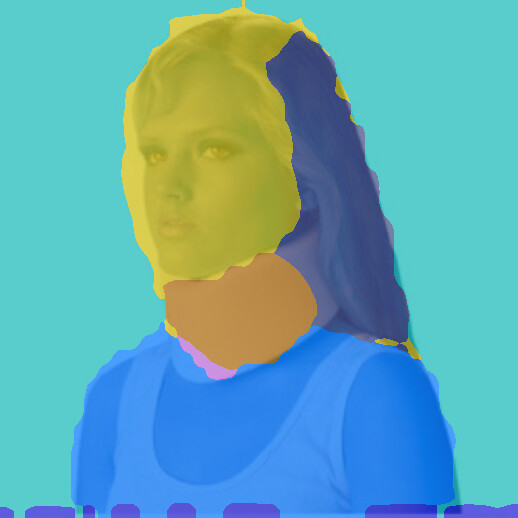} &
            \myig{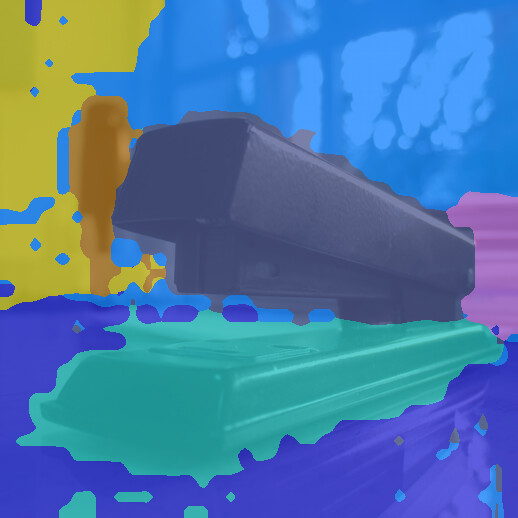} & 
            \myig{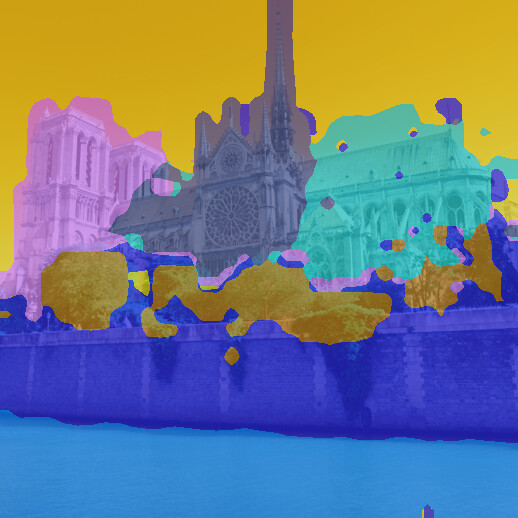}
        \end{tabular}
        \caption{Splitting objects into parts.}
        \label{app:subfig:examples-failures-parts}
    \end{subfigure}
    
    \begin{subfigure}[t]{\textwidth}
        \centering
        \begin{tabular}{@{}cccc@{}}
            \myig{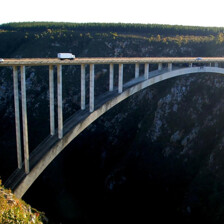} &
            \myig{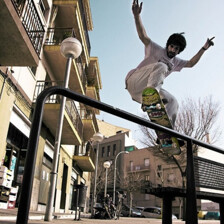} &
            \myig{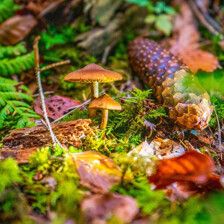} & 
            \myig{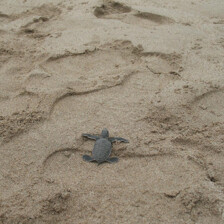} \\
            \myig{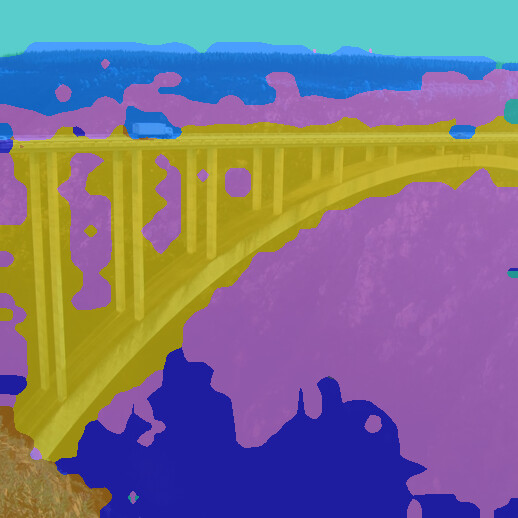} &
            \myig{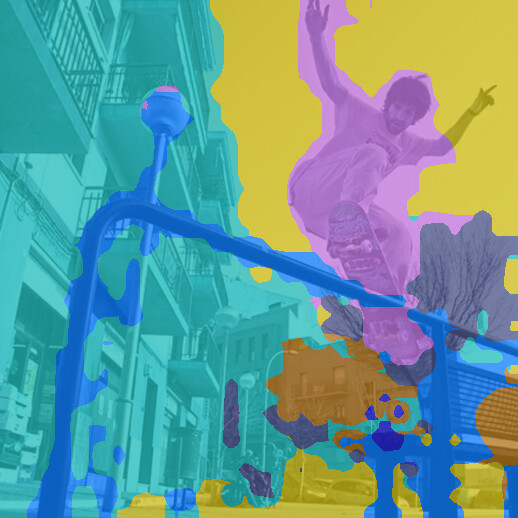} &
            \myig{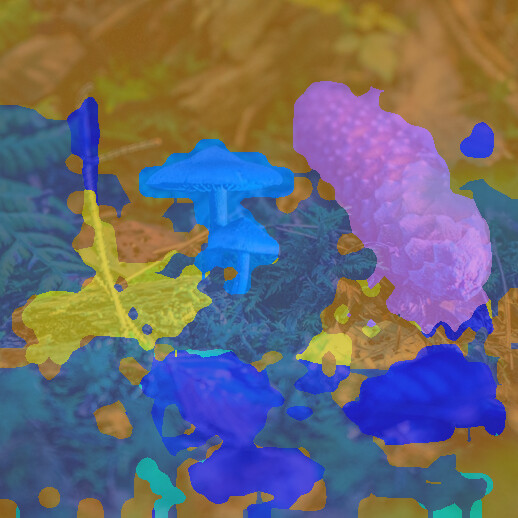} & 
            \myig{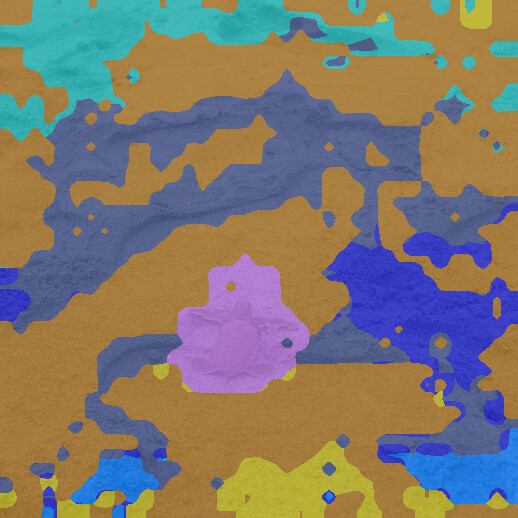}
        \end{tabular}
        \caption{Complex or unusual images: (1) tiny objects, (2) incorrect 3D inference, (3, 4) OOD scenes.}
        \label{app:subfig:examples-failures-complex}
    \end{subfigure}%
    \caption{Failure modes of \method.
    We show typical failure cases grouped into three categories: 
    (a) joining semantically related objects into a single object;
    (b) splitting objects into parts; and
    (c) incorrect decomposition of complex or unusual scenes.
    Note that the model's decompositions in (a) and (b) arguably are correct but do not correspond to the labeling prescribed by the ground truth annotations; without knowledge of the intended downstream task, the ``correct'' grouping is ambiguous.
    We use the model from \cref{tab:eval-comparison-prior-work-coco}; it uses a ViT-B/14 encoder with hi-res adaptation and is trained on the \COCO dataset.
    All images show zero-shot predictions on the \EntitySeg dataset.
    }
    \label{app:fig:examples-failures}
\end{figure}

Even though \method brings large improvements over \DINOSAUR, the model still exhibits problems with certain types of scenes.
In \cref{app:fig:examples-failures}, we show several examples of such failure cases, grouped into modes of failure.
Two typical categories of failure are the overgrouping of semantically-related objects (\cref{app:subfig:examples-failures-overgrouping}), and the split of objects into parts (\cref{app:subfig:examples-failures-parts}).
Both problems are primarily caused by the model using the wrong number of slots.
But note that even having access to the ``correct'' number of slots per image can not resolve all problems, as the model may still allocate the slots in undesirable ways.
Consider the last example in \cref{app:subfig:examples-failures-overgrouping}: here, the model could correctly split the two persons into individual slots if the motorbike is grouped as one object instead of as parts.
A third category of failure broadly stems from difficult or unusual images (\cref{app:subfig:examples-failures-complex}): for example, grouping tiny objects together with the background (cars on bridge); incorrect 3D inference  due to unusual camera perspective (rail and light post); sub-optimal decompositions in OOD scenes (grass stalk in forest, sand patterns).

\emph{How could the failure modes regarding overgrouping and oversplitting be resolved?}
First, like all slot attention/\DINOSAUR-based methods, \method decomposes the scene into a fixed number of regions/objects. 
However, especially on real-world images, the number of objects varies significantly from image to image. 
Therefore, it is important to develop methods that infer a suitable number of objects for an image; however, further innovations are needed to deal with the slot allocation problem we have alluded to before.
Second, unsupervised scene decomposition is inherently an ill-defined task on real-world data as scenes can be split in numerous ways (\cf \cref{app:subfig:examples-failures-overgrouping,app:subfig:examples-failures-parts}).
Thus, predicting only a single set of masks might ultimately be insufficient. 
Instead, it may be beneficial to model the full \emph{part-whole hierarchy}, producing various decompositions of different granularity. 
Such models could further allow \emph{control} over the level of granularity through external conditioning variables or text. 
However, the examples in \cref{app:subfig:examples-failures-overgrouping,app:subfig:examples-failures-parts} also demonstrate the limitations of current evaluation techniques.
Arguably, these are not failures of the model, but are treated as such by the evaluation metrics.
This is because current datasets have annotations that prescribe a single ground truth labeling for each image.
Instead, datasets should be annotated with multi-level labelings, \eg by including parts of objects, or further splitting the background into specific elements (\eg splitting the background "tree" class into particular trees).
To evaluate methods that model the full part-whole hierarchy, such annotations even become a necessity.

\section{Method Details}

\subsection{Improved Hyperparameters}
\label{app:subsec:imp-hps}
As discussed in \cref{sec:method} in the main paper, we found an improved set of hyperparameters that work well for finetuning the pre-trained ViT encoder.
We split these into general hyperparameters (G-HPs), affecting all modules of the model, and encoder hyperparameters (E-HPs), only affecting the finetuning of the encoder (see also \cref{app:tab:hps}).
We ablate the effect of these groups of hyperparameters in \cref{tab:method-ablation-coco} in the main paper.

The general hyperparameter changes are as follows:
\begin{itemize}
    \item Increasing \emph{batch size} from 64 to 128.
    \item Decreasing \emph{base learning rate} from 0.0004 to 0.0003.
    \item Switching from an exponential decay \emph{learning rate schedule} to a cosine schedule.
    \item Lowering \emph{gradient clipping} from 1.0 to 0.1.
\end{itemize}
The hyperparameter for encoder finetuning are as follows:
\begin{itemize}
    \item Lowering the \emph{base learning rate} for the encoder by a factor of 0.5 from 0.0003 to 0.00015.
    \item Introducing \emph{blockwise learning rate decay} with a decay rate of 0.85.
    \item Adding \emph{weight decay} of 0.01 to the encoder parameters in conjunction with the AdamW optimizer.
\end{itemize}

Note that these changes resulted from a joined hyperparameter search over all individual hyperparameters and it is highly likely that (1) not all of these parameters changes are necessary, and (2) an even better set of hyperparameters can be found.

\subsection{Top-K Decoding}
\label{app:subsec:topk}

We first describe the MLP decoder from \citet{seitzer2023bridging}.
For $N$ patches and $K$ slots, the MLP decoder produces a reconstruction $\hat{\vy} \in \mathbb{R}^{N \times K \times D}$, as well as an alpha mask $\bm{\alpha} \in \mathbb{R}^{N \times K}$ that shows how active each slot is at each patch.
The final reconstruction $\vy \in \mathbb{R}^{N \times D}$ is then given by taking a weighted average over the slots, that is, the reconstruction $\vy_i$ for patch $i$ is given by
\begin{equation}
\vy_i = \sum_{\kappa=1}^K \hat{\vy}_{i,\kappa} \odot \vm_{i,\kappa},\quad\quad \vm_{i,\kappa} = \left(\mysoftmax_{j} \bm{\alpha}_{i,j}\right)_\kappa.
\end{equation}
With \textbf{top-k decoding}, we only take the $k \in \gK_i$ most active slots into account for each patch $i$, as determined by the slot attention mask $\va \in [0, 1]^{N \times K}$:
\begin{equation}
\vy_i = \sum_{\kappa \in \gK_i} \hat{\vy}_{i,\kappa} \odot \vm_{i,\kappa},\quad\quad \vm_{i,\kappa} = \left(\mysoftmax_{j \in \gK^i} \bm{\alpha}_{i,j}\right)_\kappa, \quad\quad \gK_i = \topk\left(\va_i, k\right),
\end{equation}
where $\topk(\vx, k) = \argmax_{I \subseteq \{1,\ldots,n\}: |I| = k } \sum_{i \in I} \vx_i$ is the function that selects the indices of the $k$ highest values of the vector $\vx \in \mathbb{R}^n$.
In practice, we can efficiently implement the decoding step by first broadcasting slots to patches and adding the positional encoding, then packing the top-k slots for each position together using a gather operation, directly resulting in reconstructions $\hat{\vy} \in \mathbb{R}^{N \times k}$ and alpha masks $\bm{\alpha} \in \mathbb{R}^{N \times k}$.

\begin{table}
\caption{
    Hyperparameters for the \DINOSAUR and \method models displayed in \cref{tab:method-ablation-coco}.
    The second column (\DINOSAUR +Enc.~Train (random init.)) also lists hyperparameters for training with random encoder initialization, as discussed in \cref{sec:method}.
    Results in \cref{fig:eval-finetuning-in-distribution} use the settings in the fourth column (\DINOSAUR +FT w/G-HP's w/E-HP's).
    Results in \cref{tab:eval-comparison-prior-work-coco,fig:eval-comparison-ood-average} use the settings in the last column, but with a ViT-B/14 encoder.
    See also \cref{app:subsec:imp-hps} for a concise description of the improved hyperparameters for finetuning (G-HP's and E-HP's).
}
\label{app:tab:hps}
\vspace{0.25em}
\centering
\setlength{\tabcolsep}{0.2em}
\renewcommand{\arraystretch}{1.3}
\adjustbox{max width=\textwidth}{
\begin{tabular}{@{}l@{\hspace{-1em}}p{13em}@{\hspace{-3em}}r|r|r|r|rr@{}}
\toprule
\textbf{Models}                            &                           & \textbf{\DINOSAUR}  & \textbf{\DINOSAUR}    & \textbf{\DINOSAUR} &  \textbf{\DINOSAUR} & \textbf{\DINOSAUR}  \\ 
                            &                           &  & \textbf{+Enc. Train}    & \textbf{+FT.} &  \textbf{+FT} & \textbf{+FT}  \\
                            &                           &  & \textbf{(random init.)}    & \textbf{w/G-HP's} &  \textbf{w/G-HP's} & \textbf{+Top-k }  \\ 
                            &                           &  &    &  &  \textbf{w/E-HP's} & \textbf{+High-Res. }  \\

                            \midrule
                            
Training Steps                     &                           & 300k       & 300k       & 300k         & 300k   & 10k   \\
Batch Size                         &                           & 64         & 128         & 128           & 128     & 64           \\
\midrule

Image/Crop Size                                                                                     &                           & 224        & 224        & 224 & 224  & 518 \\
Cropping Strategy                                                                              &                           & Random         & Random         & Random       & Random  & Random           \\
Augmentations                                                                             &                           & --         & --         &  -- & -- & --     \\
Image Tokens                                                                                   &                           & 784/256      & 256       & 256  & 256   & 1369  \\ 
\midrule

\multirow{4}{*}{LR}& LR Warmup Steps                                            & 10000      & 10000      & 10000        & 10000  & 333     \\

    &  Base LR \big(Total LR = $\text{Base LR}\cdot \sqrt(\frac{\text{Batch Size}}{64})$\big)                                               & 0.0004     & 0.0003     & 0.0003       & 0.0003 & 0.0001           \\
& Exp.~Decay Half-Life               & 100k       & 100k       & 100k       & 100k     & 10k     \\
& Schedule  & Exponential       &  Cosine      &  Cosine      &  Cosine    & Cosine    \\
\midrule

\multirow{4}{*}{Encoder LR} & Blockwise LR & \xmark &  \cmark      &  \xmark      &   \cmark   & \cmark    \\
& LR Factor \big(Total LR = $\frac{\text{Base LR}}{\text{Encoder LR Factor}}\cdot \sqrt(\frac{\text{Batch Size}}{64})$\big) & \xmark     &   0.5     & \xmark        &    0.5  &    0.5 \\
& Layerwise LR Decay Factor ($\eta$)\newline \big($\text{LR}_{\text{layer l}} = \eta \cdot \text{LR}_{\text{layer l+1}}$\big) & \xmark     &   0.85     &  \xmark      &   0.85   &   0.85  \\
&  Weight Decay & \xmark          &  0.01       & \xmark    & 0.01    & 0.01  \\
\midrule
\multirow{5}{*}{Encoder}                   & Type & ViT-B/16 / ViT-S/14      & ViT-S/14      & ViT-S/14        & ViT-S/14  & ViT-S/14          \\
& Pre-training                                                   & DINO/DINOv2          & --          & DINOv2           & DINOv2    & DINOv2         \\
& Patch Size                                                   & 16/14          & 14          & 14           & 14    & 14         \\
& Feature Dim. $D_\text{feat}$             & 768/384       & 384        & 384         & 384  & 384        \\
& Gradient Norm Clipping                                                                                                       & 1.0        & 0.1       & 0.1          & 0.1    & 0.1          \\ 
\midrule
\multirow{2}{*}{Target Encoder}                   & Type & ViT-B/16 / ViT-S/14      & ViT-S/14      & ViT-S/14        & ViT-S/14  & ViT-S/14          \\
& Pre-training                                                   & DINO/DINOv2          & DINOv2          & DINOv2           & DINOv2    & DINOv2         \\
\midrule
\multirow{4}{*}{Slot Attention}      & Slots                     & 7         & 7         & 7            & 7  & 7     \\
& Iterations                & 3          & 3          & 3           & 3         & 3        \\
& Slot Dim.~$D_\text{slots}$          & 256        & 256        & 256          & 256   & 256  \\
& MLP Hidden Dim.          & 1024        & 1024        & 1024          & 1024    & 1024   \\
\midrule
\multirow{4}{*}{Decoder}            & Type                      & MLP          & MLP      & MLP  & MLP    & MLP   \\
& Layers                    & 4          & 4          & 4            & 4       & 4 \\

& MLP Hidden Dim.               & 2048       & 2048        & 2048         & 2048     & 2048         \\    
& Top-k               & \xmark       & \xmark        & \xmark         & \xmark     & 3         \\    
\bottomrule
\end{tabular}}
\end{table}

\section{Methods \& Hyperparameters}
\label{app:sec:method-details}

\paragraph{\DINOSAUR \citep{seitzer2023bridging}} 
\DINOSAUR introduced the idea of applying slot attention on a pre-trained encoder and training the model by reconstructing the features of this pre-trained encoder.
This also forms the base of our proposed approach.
In \DINOSAUR, the encoder is kept fixed while the slot attention and the decoder modules are trainable. 
While the original paper considers two kinds of decoders --- (1) Transformer Decoder and (2) MLP decoder --- in this work we mainly compare against \DINOSAUR with the MLP decoder. 
We consider two variants of \DINOSAUR, using \DINO{} \citep{Caron2021DINO} and \DINO{}v2 \citep{oquab2023dinov2} pre-trained backbones respectively. 
We list the hyperparameters used for \DINOSAUR in Table \ref{app:tab:hps} (first column). 

\paragraph{\method} 
Our method is implemented upon \DINOSAUR and thus shares low-level implementation details.
In Table \ref{app:tab:hps}, we list the hyperparameters for the following models mentioned in \cref{sec:finetuning} and listed in Table \ref{tab:method-ablation-coco}: (1) \DINOSAUR + Training from Random Init., (2) \DINOSAUR + FT w/G-HP's, (3) \DINOSAUR + FT w/G-HP's \& E-HP's, (4) \DINOSAUR + FT, + Top-k, + High-Res. Finetuning. 
While the models listed in Table \cref{app:tab:hps} all use \DINO{}v2 with the ViT-S/14 backbone, the same hyperparamters are applicable for models using ViT-B/16 and ViT-B/14 backbones as well. 
For training our model, we use a single A100 GPU per run. 
Each training run of the proposed finetuning approach requires 2--3 days of training. 

\paragraph{SlotDiffusion \citep{wu2023slotdiffusion}} 
SlotDiffusion utilizes a latent diffusion model as the decoder.
The specific variant of the SlotDiffusion model which we consider here is the one which uses a pretrained \DINO{} encoder (ViT-B/16) to encode the images similar to \DINOSAUR. 
We use the pre-trained checkpoint released by the authors\footnote{\url{https://github.com/Wuziyi616/SlotDiffusion}} for all the comparisons in this work.

\paragraph{\SPOT \citep{kakogeorgiou2023spot}} 
SPOT uses a two-stage training procedure. 
In the first stage, a \DINOSAUR model is trained similar to \citet{seitzer2023bridging}. 
In the second stage, a student-teacher setup is employed, where the model trained model from the first stage acts as a teacher and the student is a new model. 
During this stage, the model is trained with two objectives: 
(1) a feature reconstruction loss, where the targets come from the teacher, and (2) a attention distillation loss, where the teachers attention masks from slot attention are distilled into the student. 
Moreover, \SPOT uses a Transformer encoder as opposed to an MLP decoder. 
Similar to SlotDiffusion, for \SPOT, also we use a pre-trained checkpoint released by the authors\footnote{\url{https://github.com/gkakogeorgiou/spot}} for the evaluations in this work. 
The pre-trained checkpoint uses a ViT-B/16 encoder initialized with \DINO{} weights.

\paragraph{Segment Anything \citep{kirillov2023segment}} 
The Segment Anything model (\SAM) is a large foundation model for object detection and segmentation trained supervised. 
It has three stages of training: (1) a manual stage, where the model is trained using 120k images annotated with 4.3M masks obtained from human labelers; (2) a semi-automatic stage, where the model is trained on 180k annotated with 5.9M masks partly annotated by human labelers and partly annotated by itself; and (3) a fully automatic stage, where the model is trained on 11M images with 1.1B masks annotated by the model itself. 
We consider 2 variants of \SAM{}: \textit{comp.} (Comparable) and \textit{best}. 
Note that \SAM includes an IoU prediction MLP which outputs an estimated IoU for each predicted mask. 
For the \textit{comp.} variant, we use the ViT-Base model considering the top $K$ masks by predicted IoU, where the value of $K$ is based on the optimal number of objects for each dataset as listed in \cref{app:tab:dataset_properties}. For the \textit{best} variant, we use the ViT-Huge model keeping all masks above a IoU threshold $\tau$.
We evaluated values for $\tau \in \{0.9, 0.95, 0.99\}$ and found that $\tau = 0.9$ works best across all datasets.

For inference, we use a single A100 GPU for each of the baselines and the proposed approach. 

\section{Datasets}
\label{app:sec:datasets}

\begin{table}
    \centering
    \caption{Number of images per dataset and the used number of slots for training and evaluating on each dataset.}
    \begin{tabular}{@{}lcc@{}}
        \toprule
        Dataset & Num.~Images & Num.~Slots \\
        \midrule
        \COCO 2017 train & \np{118287} & 7 \\
        \COCO 2017 validation & \np{5000} & 7 \\
        \EntitySeg train & \np{31789} & 7 \\
        \EntitySeg validation & \np{1498} & 7 \\
        \PASCALVOC 2012 train & \np{10582} & 7 \\
        \PASCALVOC 2012 validation & \np{1449} & 7 \\
        \MOVi-C train & \np{87633} & 11 \\
        \MOVi-C validation & \np{4200} & 11 \\
        \MOVi-E train & \np{87633} & 11 \\
        \MOVi-E validation & \np{4176} & 11 \\
        \ScanNet train & \np{10000} & 6 \\
        \ScanNet validation & \np{2000} & 6 \\
        \YCB train & \np{10000} & 6 \\
        \YCB validation & \np{2000} & 6 \\
        \ClevrTex train & \np{40000} & 11 \\
        \ClevrTex validation & \np{5000} & 11 \\        
        \bottomrule
    \end{tabular}
    \label{app:tab:dataset_properties}
\end{table}

This section gives a detailed description of the datasets that comprise the introduced zero-shot benchmark for object-centric representation learning (\cref{subsec:zero-shot-benchmark}). 
The benchmark consists of 8 different datasets, ranging from synthetic to real-world scenes.
See also \cref{app:tab:dataset_properties} for an overview over the number of images per dataset.

\paragraph{\COCO~\citep{Lin2014COCO}}
This dataset contains complex images containing real-world objects in their natural context. For training, we use the \COCO 2017 dataset which consists of \np{118287} images. 
For evaluation, we use \np{5000} images from the validation sets. 
Similar to \citet{seitzer2023bridging}, we use instance masks to evaluate object discovery. 
Additionally, we also add the task of panoptic segmentation to our evaluation suite for the \COCO dataset, using the panoptic labeling provided by \citet{kirillov2019panoptic}. 
Panoptic segmentation combines the task of instance segmentation, which requires the model to segment each object/foreground/thing instance, and semantic segmentation, which requires the model to segement each background/stuff class.
The metrics we use for measuring panoptic segmentation are panoptic ARI and panoptic quality (see \cref{app:sec:metrics}).
Following \citet{seitzer2023bridging}, we evaluate square center crops, where the input images are resized to $224 \times 224$ pixels, and the targets masks are resized to $320 \times 320$ pixels.

\paragraph{\EntitySeg~\citep{qilu2023entityseg}}
This dataset consists of complex real world images spanning a diverse range of entities. 
In contrast to \COCO, \EntitySeg is an open-world dataset and does not have a pre-defined set of object classes.
It consists of a large number of high-resolution images (71.25\% and 86.23\% of the images are of high resolution with at least \np{2000}px for the width and \np{1000}px for the height). 
Each image is annotated with high-quality fine-grained mask annotations.
The version of the dataset utilized in this work consists of \np{31789} images for training and \np{1498} images for evaluation. 
We evaluate the instance segmentation masks for object discovery.
As in \COCO, we evaluate square center crops, where the input images are resized to $224 \times 224$ pixels, and the targets masks are resized to $320 \times 320$ pixels.

\paragraph{\PASCALVOC~\citep{Everingham2012PASCALVOC}}
Similar to \citet{seitzer2023bridging}, we use the ``trainaug'' variant of the \PASCALVOC dataset for training. 
It consists of a total of \np{10582} images for training, where \np{1464} are from the segmentation train set and \np{9118} are from the SBD dataset \citep{hariharan2011semantic}. 
For evaluating object discovery, we use the official instance segmentation validation split with \np{1449} images.
Following \citet{seitzer2023bridging}, we evaluate square center crops, where the input images are resized to $224 \times 224$ pixels, and the targets masks are resized to $320 \times 320$ pixels.

\paragraph{\MOVi-C and \MOVi-E~\citep{Greff2021Kubric}}
The \MOVi datasets are synthetically generated video datasets consisting of multiple objects per video. 
Each video is generated by placing 3D scanned objects on real-world backgrounds. 
\MOVi-C contains up to 11 objects per video and MOVI-E contains up to 23 objects per video. 
Additionally, \MOVi-E also features the camera moving in random directions. 
For our case, we treat these datasets as image datasets. 
We sample 9 frames per video which yields a total of \np{87633} training images for \MOVi-C and \np{87741} images on \MOVi-E. 
For evaluation, we use \np{4200} frames for \MOVi-C and \np{4176} frames for \MOVi-E from the validation sets in each case. 
We use a resolution of $128 \times 128$ for both input images and target masks.

\paragraph{\ScanNet and \YCB~\citep{yang2022promising}}
These datasets consist of real-world objects on black backgrounds and were originally introduced to test limitations of object-centric learning methods~\citep{yang2022promising}.
\ScanNet (originally from \citet{dai2017scannet}) consists of objects that can be typically be found in indoor scenes (\eg furniture) and \YCB (originally from \citet{Calli2015YCB}) consists of 21 different classes of everyday objects (\eg food items, kitchen items, tools, etc.). 
Each of these dataset consist of \np{10000} training images and \np{2000} evaluation images. 
Both datasets consist of 2--6 objects per scene.
We use a resolution of $128 \times 128$ for both input images and target masks.

\paragraph{\ClevrTex~\citep{Karazija2021clevrtex}}
This is a synthetically constructed dataset where each scene consists of 3--10 simple geometric 3D shapes arranged in a background sampled from a catalogue of 60 different materials. 
The materials of the objects are also sampled from the same catalogue. 
This dataset contains \np{40000} images for training and \np{10000} for validation and test each. 
We use the \np{5000} images from the validation set for our evaluation. 
\ClevrTex also offers various OOD splits which utilize materials not seen during training. 
We do not use these splits; for our zero-shot generalization evaluation, we can directly use the main split since it usually is not a part of the training set we use to train the object-centric model. 
We use a resolution of $240 \times 240$ for both input images and target masks.

\section{Metrics}
\label{app:sec:metrics}

\paragraph{FG-ARI} 
The \textit{adjusted rand index} (ARI) measures the similarity between two clusterings \citep{hubert1985comparing}. We use the instance/object masks as the targets. 
We only compute this metric for pixels in the foreground (hence, FG-ARI). 
Unlabeled pixels are treated as background. 

\paragraph{mBO} 
To compute the mBO~\citep{Arbelaez2014MCG}, each predicted mask is assigned to the ground truth mask with highest overlap in terms of IoU. 
The mBO is computed as the average IoU of these mask pairs. 

\paragraph{Panoptic ARI} 
Panoptic ARI is computed as ARI, but uses panoptic mask annotations as ground truth targets. 
Panoptic masks \citep{kirillov2019panoptic} provide more detailed mask annotations for an image by assigning a different mask for separate instances of the same object (``things'') and also segmenting background regions (``stuff''). 
We only compute the Panoptic ARI for those images which have at least two masks. 

\paragraph{Panoptic Quality} 
The Panoptic Quality (PQ)~\citep{kirillov2019panoptic} is computed by first assigning each predicted mask to the ground truth mask with the highest overlap in terms of IoU, removing all matches that do not have an IoU overlap of at least $0.5$; this results in a unique matching~\citep{kirillov2019panoptic}.
These mask pairs form the set of true positives (TP). 
Ground truth masks that were not assigned a predicted mask form the set of false negatives (FN).
Similarly, predicted masks that were not assigned to a ground truth mask form the set of false positives (FP).
Predicted masks that have an IoU overlap of more than $0.5$ with pixels labeled as ``void'' or ``crowd'' are removed from the set of false positives.
The panoptic quality is then computed as:
\begin{equation}
    PQ = \frac{\sum_{(p, g) \in \mathrm{TP}}\mathrm{IoU}(p, g)}{|\mathrm{TP}| + 0.5 |\mathrm{FP}| + 0.5 | \mathrm{FN}| }
\end{equation}

\section{Examples}
\label{app:sec:examples}

In this section, we show example predictions for \DINOSAUR, \SPOT, Slot Diffusion, \method, and \SAM, where all methods besides \SAM were trained on the \COCO dataset.
\method uses a ViT-B/14 encoder with top-k and hi-res adaptation, \ie the model evaluated in \cref{tab:eval-comparison-prior-work-coco,fig:eval-comparison-ood-average}.

\begin{figure}
  \newcommand{\myig}[1]{\includegraphics[width=0.15\textwidth,valign=c]{images/#1}}
  \newcommand{\mycaption}[1]{{\footnotesize#1}}
  \renewcommand{\arraystretch}{4.7}
  \setlength{\tabcolsep}{2pt}
  \centering
  \vspace{-0.1\textwidth}
  \begin{tabular}{@{}cccccc@{}}
      \mycaption{Image} & \mycaption{\DINOSAUR{}} & \mycaption{\SPOT} & \mycaption{SlotDiffusion} & \mycaption{\methodshort{}} & \mycaption{\SAM} \\[-1.5em]

      \myig{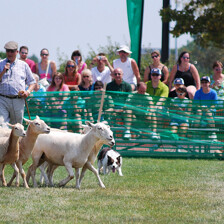} &
      \myig{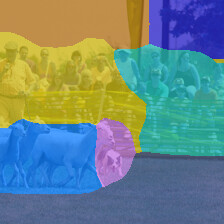} &
      \myig{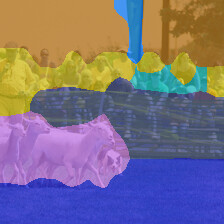} & 
      \myig{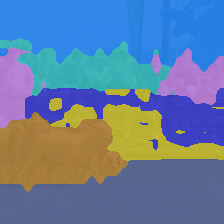} &
      \myig{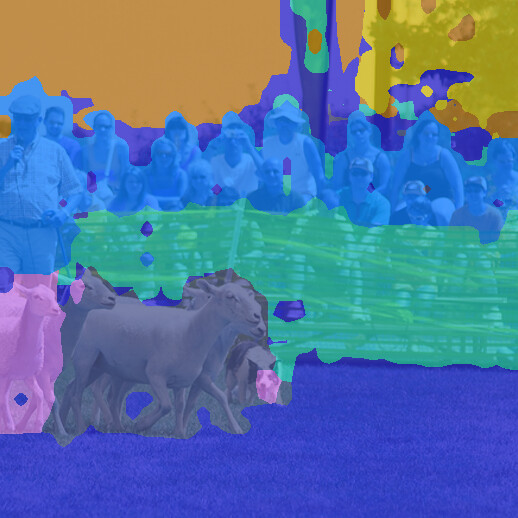} &
      \myig{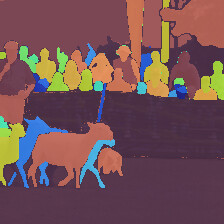} \\

      \myig{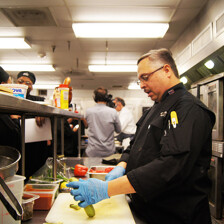} &
      \myig{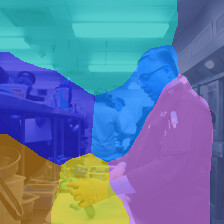} &
      \myig{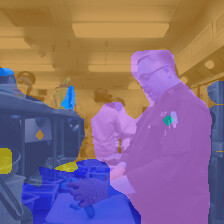} & 
      \myig{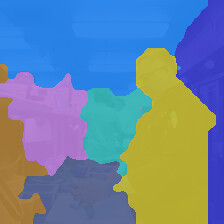} &
      \myig{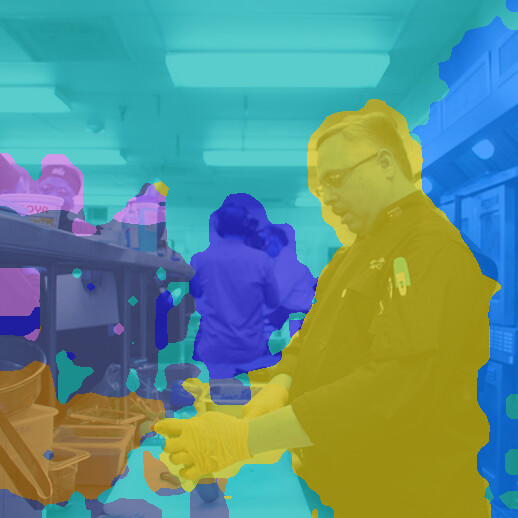} &
      \myig{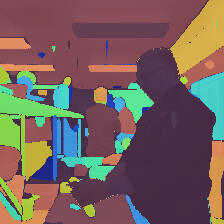} \\

      \myig{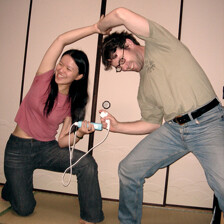} &
      \myig{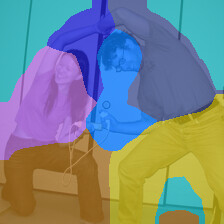} &
      \myig{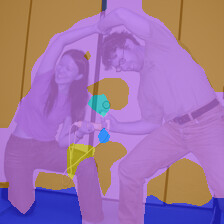} & 
      \myig{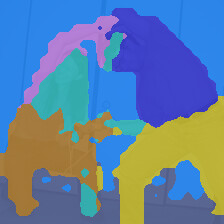} &
      \myig{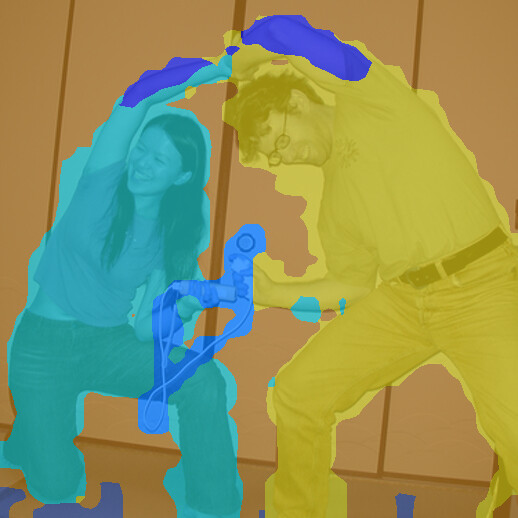} &
      \myig{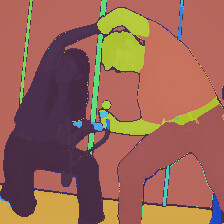} \\

      \myig{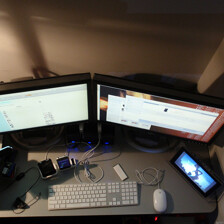} &
      \myig{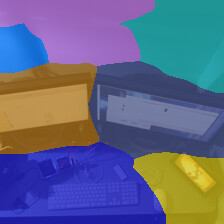} &
      \myig{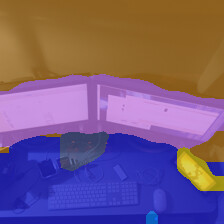} & 
      \myig{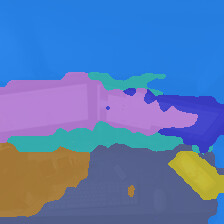} &
      \myig{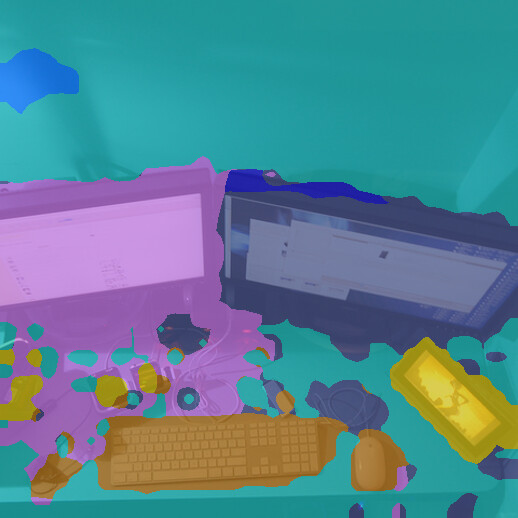} &
      \myig{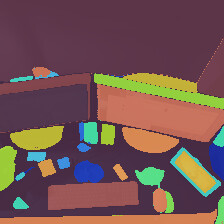} \\

      \myig{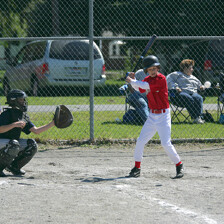} &
      \myig{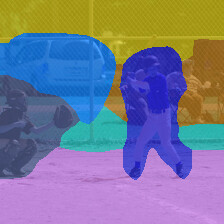} &
      \myig{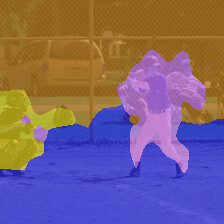} & 
      \myig{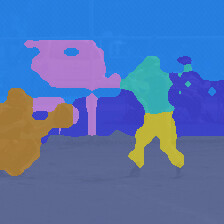} &
      \myig{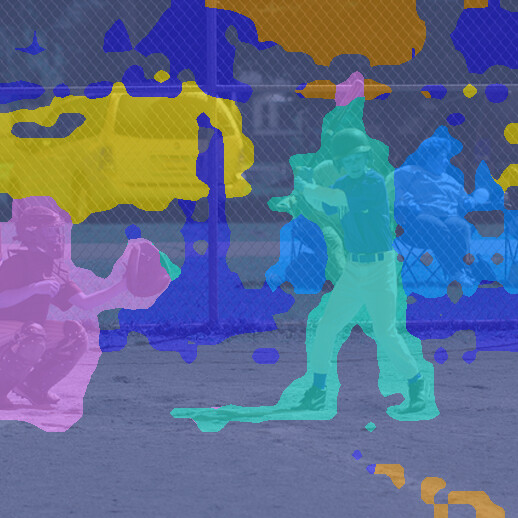} &
      \myig{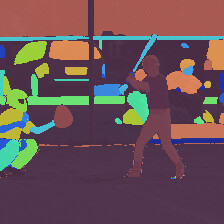} \\

      \myig{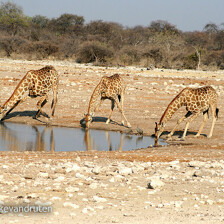} &
      \myig{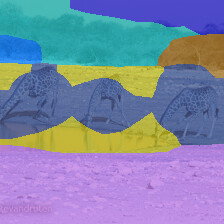} &
      \myig{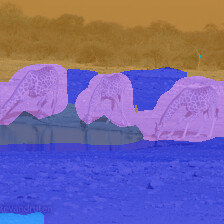} & 
      \myig{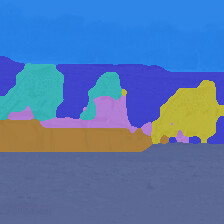} &
      \myig{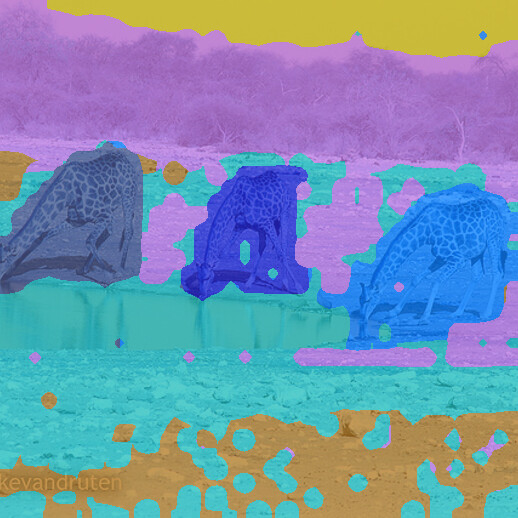} &
      \myig{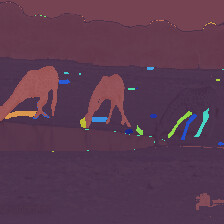} \\

      \myig{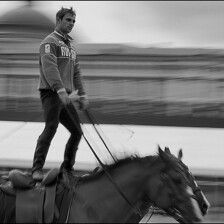} &
      \myig{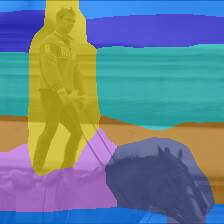} &
      \myig{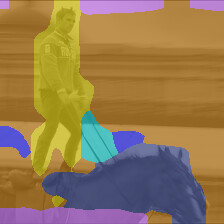} & 
      \myig{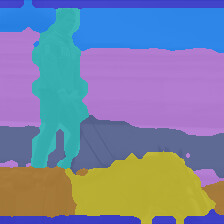} &
      \myig{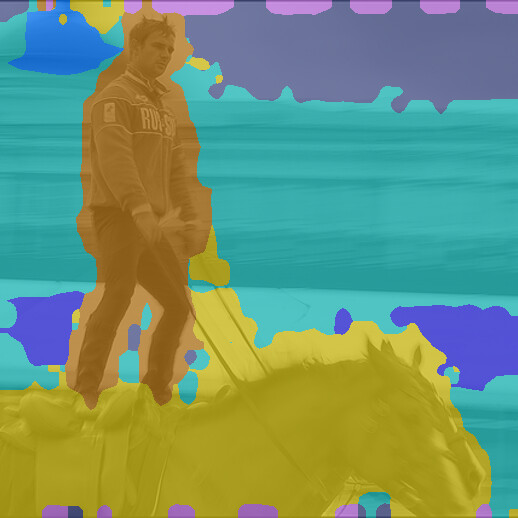} &
      \myig{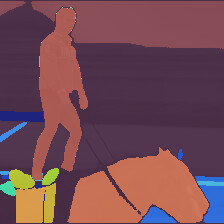} \\

      \myig{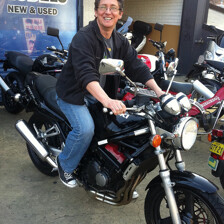} &
      \myig{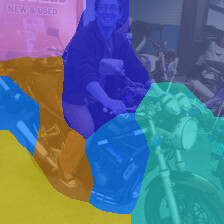} &
      \myig{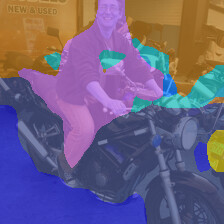} & 
      \myig{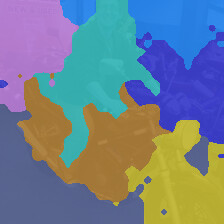} &
      \myig{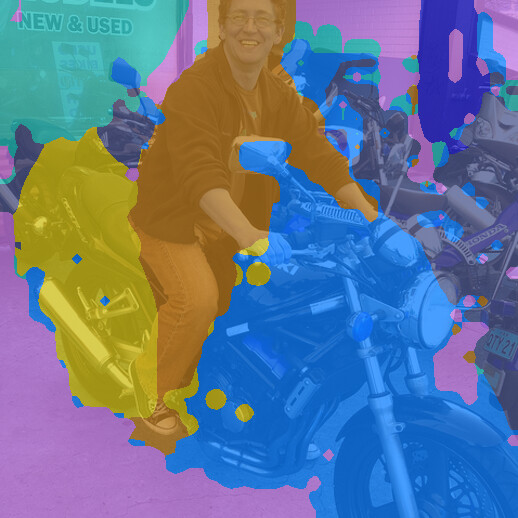} &
      \myig{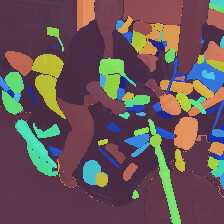} \\

      \myig{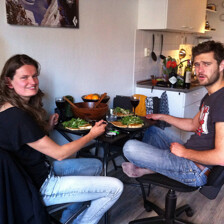} &
      \myig{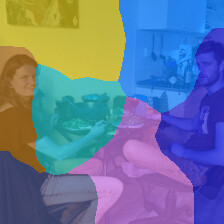} &
      \myig{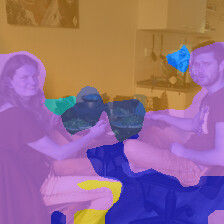} & 
      \myig{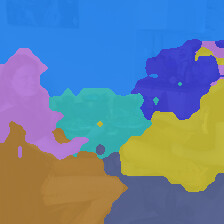} &
      \myig{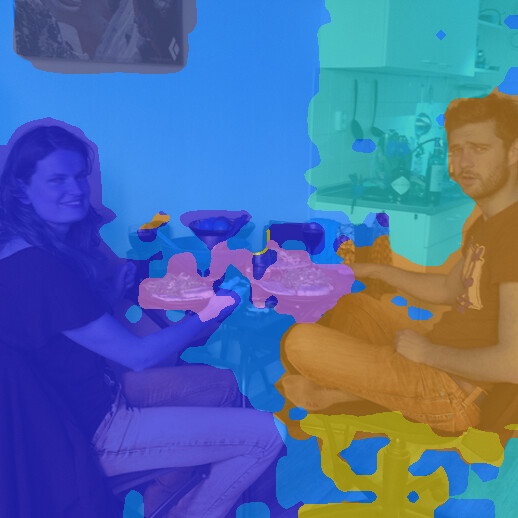} &
      \myig{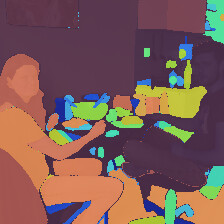} \\

      \myig{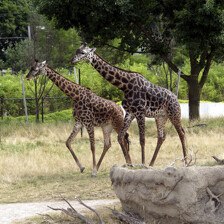} &
      \myig{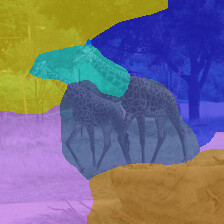} &
      \myig{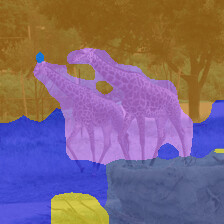} & 
      \myig{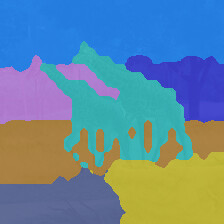} &
      \myig{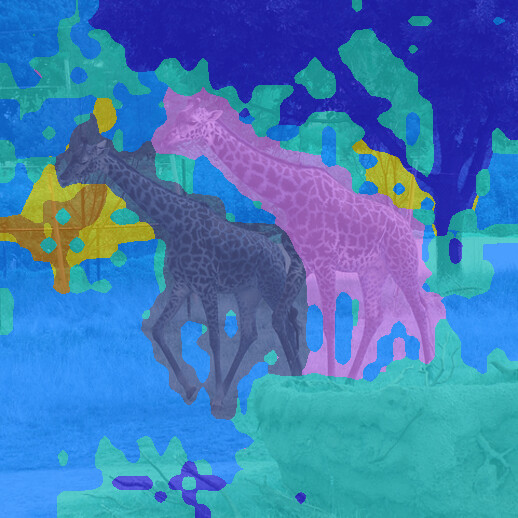} &
      \myig{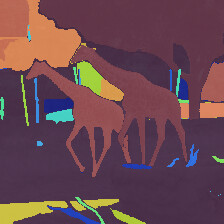} \\
  \end{tabular}
  \captionof{figure}{In-distribution examples on \COCO.}
  \label{app:fig:examples-coco}
\end{figure}

\begin{figure}
    \newcommand{\myig}[1]{\includegraphics[width=0.15\textwidth,valign=c]{images/#1}}
    \newcommand{\mycaption}[1]{{#1}}
    \renewcommand{\arraystretch}{4.7}
    \setlength{\tabcolsep}{2pt}
    \centering
    \vspace{-0.1\textwidth}
    \begin{tabular}{@{}cccccc@{}}
        \mycaption{Image} & \mycaption{\DINOSAUR{}} & \mycaption{\SPOT} & \mycaption{SlotDiffusion} & \mycaption{\methodshort{}} & \mycaption{\SAM} \\[-1.5em]

        \myig{images/entityseg/001} &
        \myig{dinosaur/entityseg/001} &
        \myig{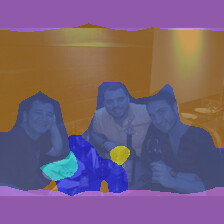} & 
        \myig{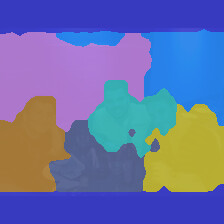} &
        \myig{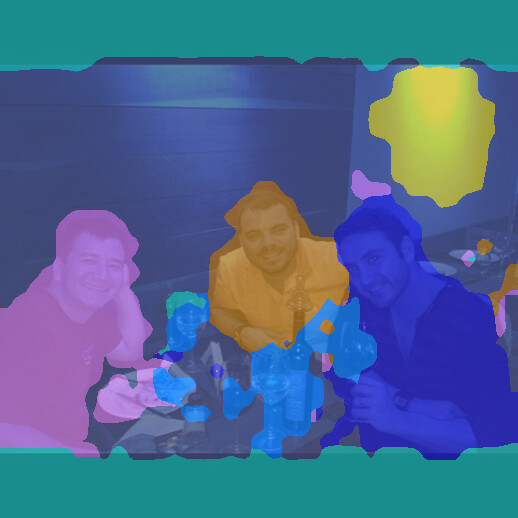} &
        \myig{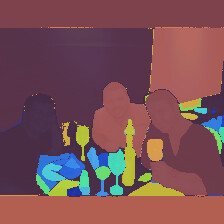} \\

        \myig{images/entityseg/003} &
        \myig{dinosaur/entityseg/003} &
        \myig{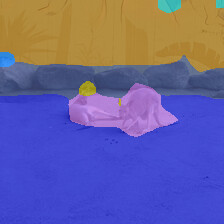} & 
        \myig{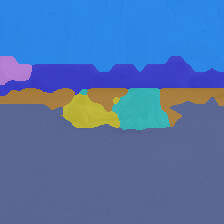} &
        \myig{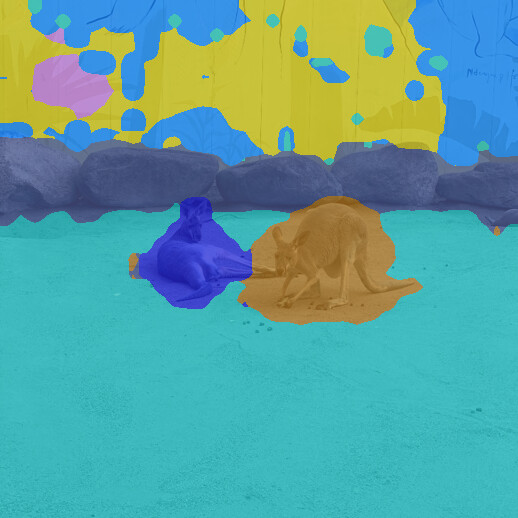} &
        \myig{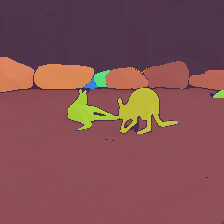} \\

        \myig{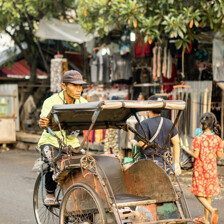} &
        \myig{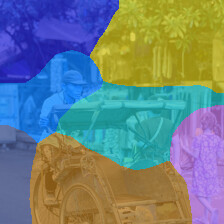} &
        \myig{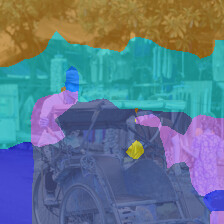} & 
        \myig{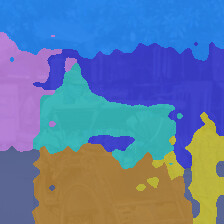} &
        \myig{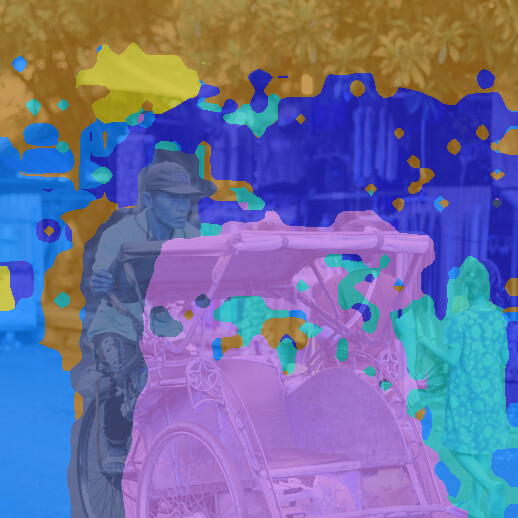} &
        \myig{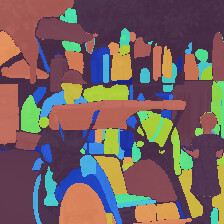} \\

        \myig{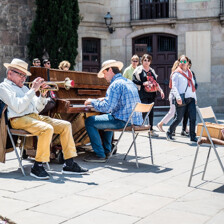} &
        \myig{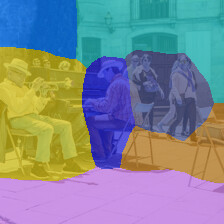} &
        \myig{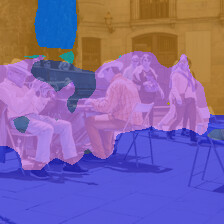} & 
        \myig{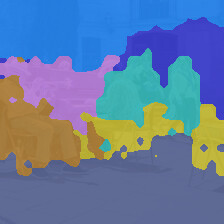} &
        \myig{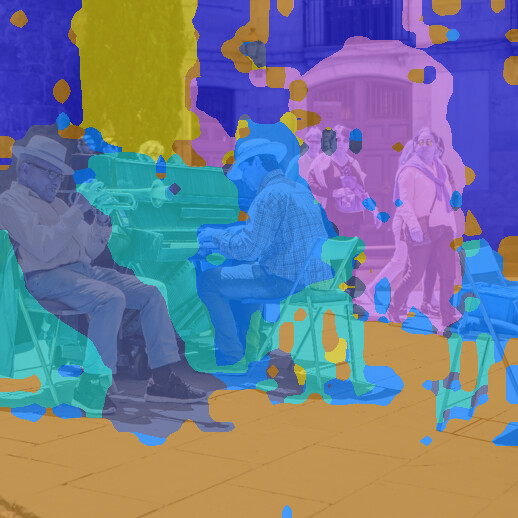} &
        \myig{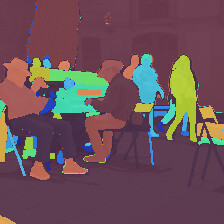} \\

        \myig{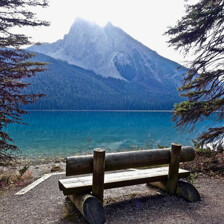} &
        \myig{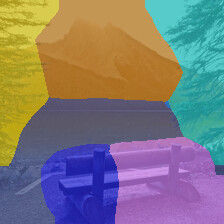} &
        \myig{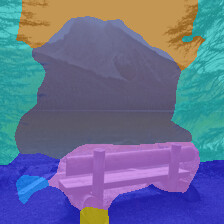} & 
        \myig{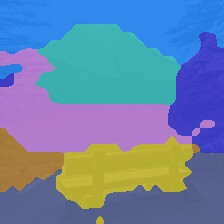} &
        \myig{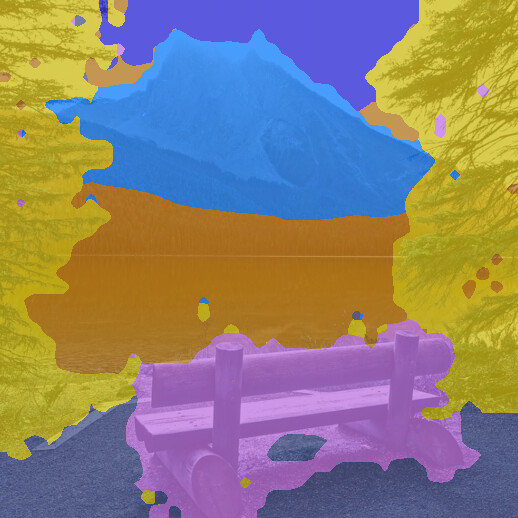} &
        \myig{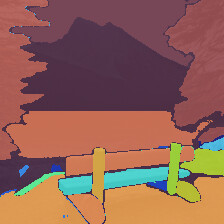} \\

        \myig{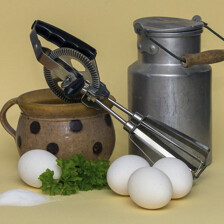} &
        \myig{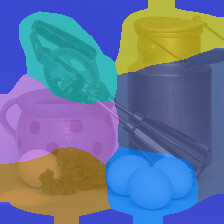} &
        \myig{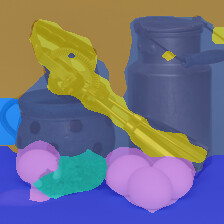} & 
        \myig{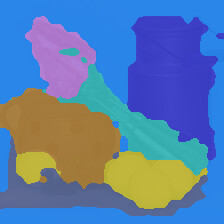} &
        \myig{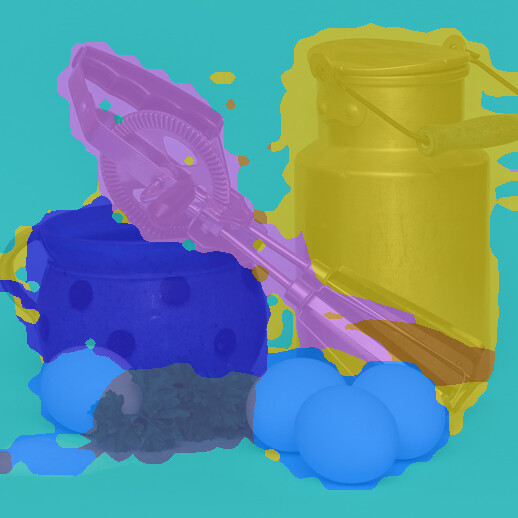} &
        \myig{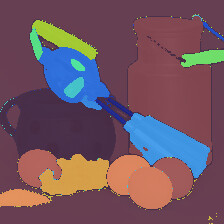} \\

        \myig{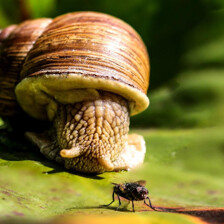} &
        \myig{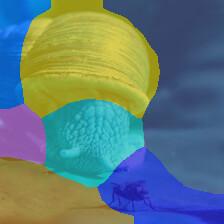} &
        \myig{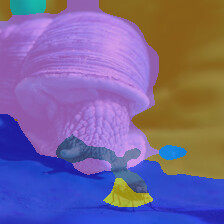} & 
        \myig{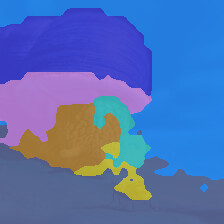} &
        \myig{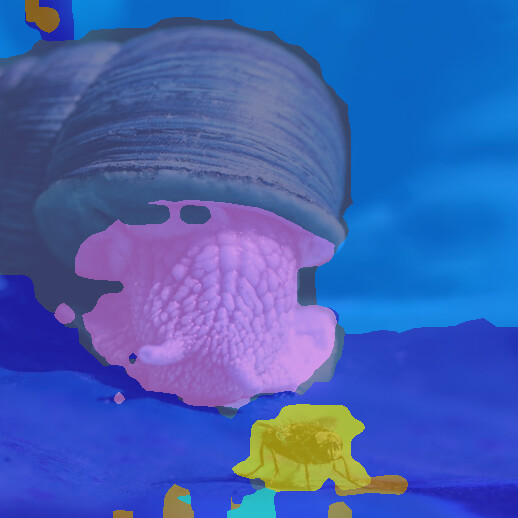} &
        \myig{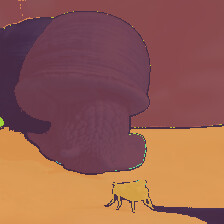} \\

        \myig{images/entityseg/048} &
        \myig{dinosaur/entityseg/048} &
        \myig{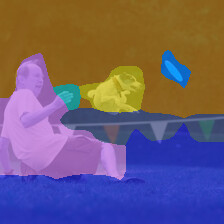} & 
        \myig{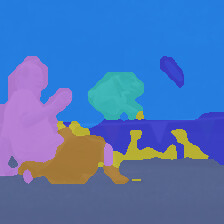} &
        \myig{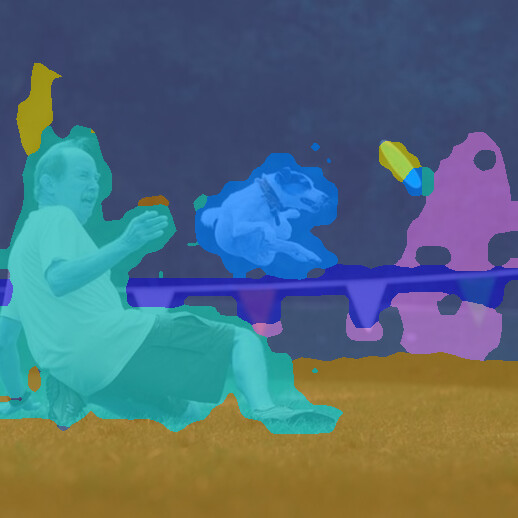} &
        \myig{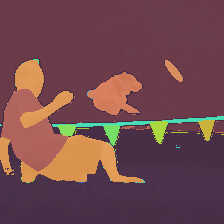} \\

        \myig{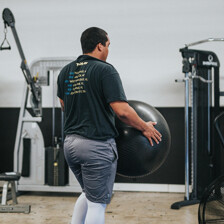} &
        \myig{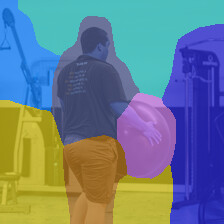} &
        \myig{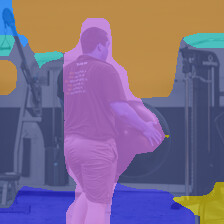} & 
        \myig{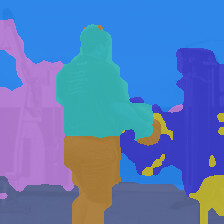} &
        \myig{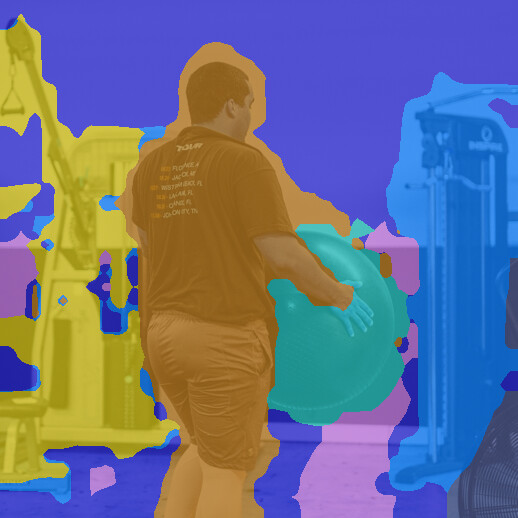} &
        \myig{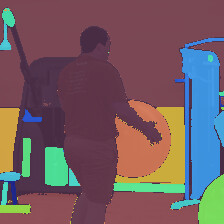} \\

        \myig{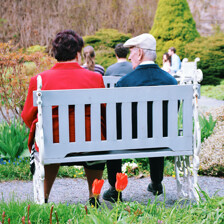} &
        \myig{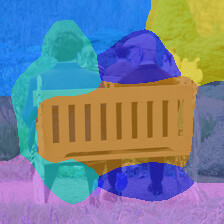} &
        \myig{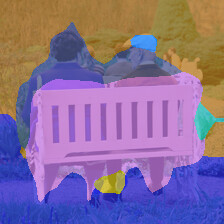} & 
        \myig{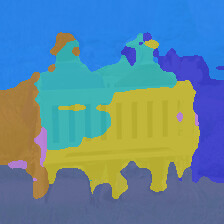} &
        \myig{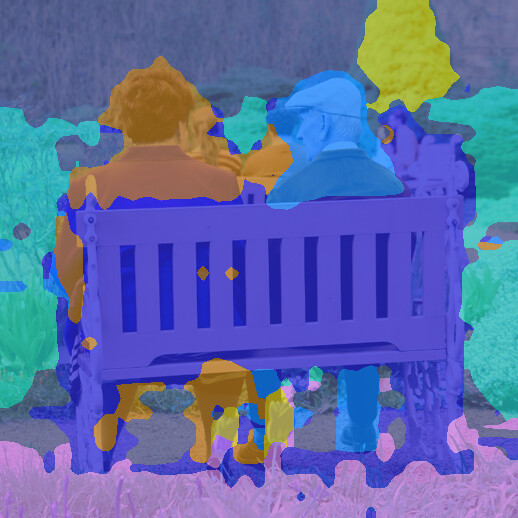} &
        \myig{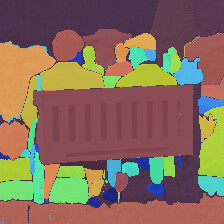} \\
    \end{tabular}
    \captionof{figure}{Zero-shot examples on \EntitySeg.}
    \label{app:fig:examples-entityseg}
\end{figure}

\begin{figure}
  \newcommand{\myig}[1]{\includegraphics[width=0.15\textwidth,valign=c]{images/#1}}
  \newcommand{\mycaption}[1]{{\footnotesize#1}}
  \renewcommand{\arraystretch}{4.7}
  \setlength{\tabcolsep}{2pt}
  \centering
  \vspace{-0.1\textwidth}
  \begin{tabular}{@{}cccccc@{}}
      \mycaption{Image} & \mycaption{\DINOSAUR{}} & \mycaption{\SPOT} & \mycaption{SlotDiffusion} & \mycaption{\methodshort{}} & \mycaption{\SAM} \\[-1.5em]

      \myig{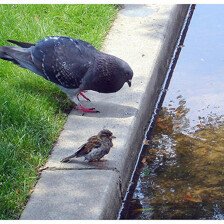} &
      \myig{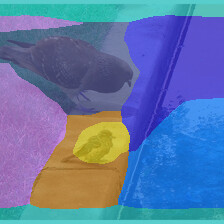} &
      \myig{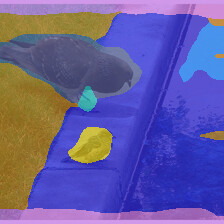} & 
      \myig{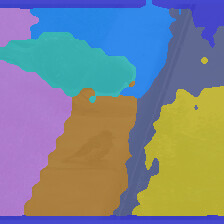} &
      \myig{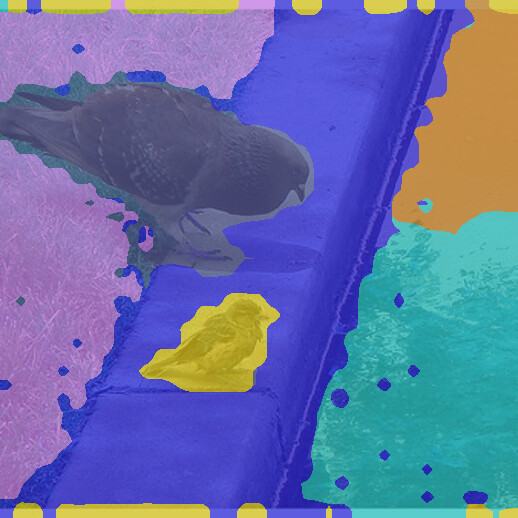} &
      \myig{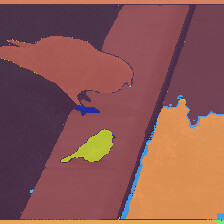} \\

      \myig{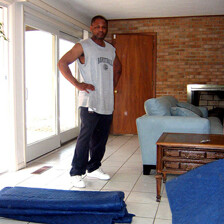} &
      \myig{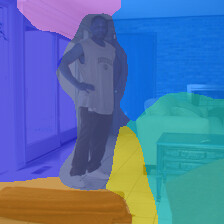} &
      \myig{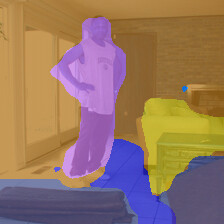} & 
      \myig{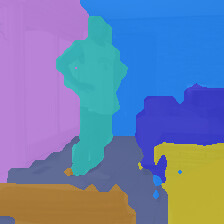} &
      \myig{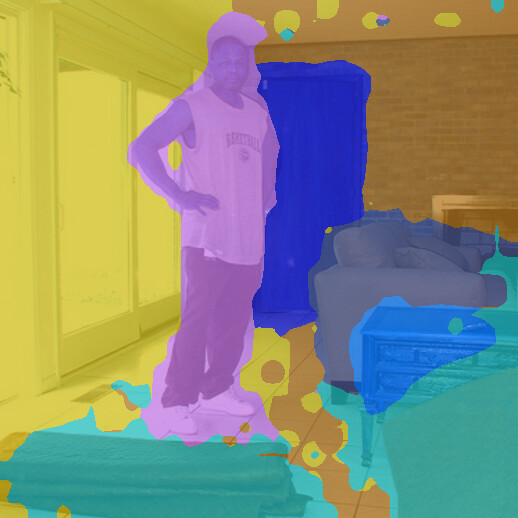} &
      \myig{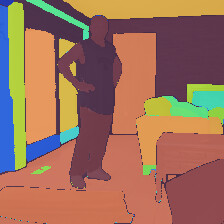} \\

      \myig{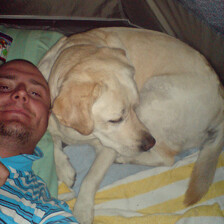} &
      \myig{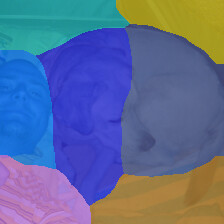} &
      \myig{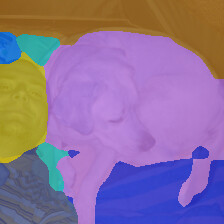} & 
      \myig{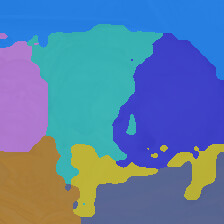} &
      \myig{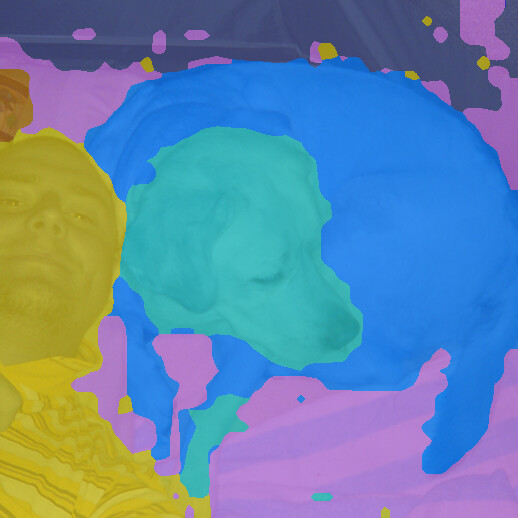} &
      \myig{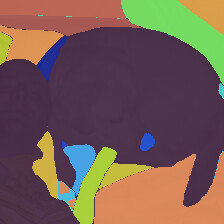} \\

      \myig{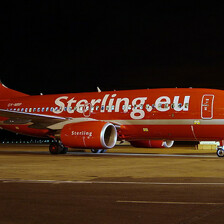} &
      \myig{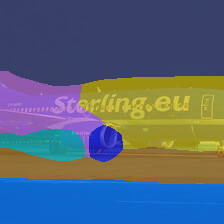} &
      \myig{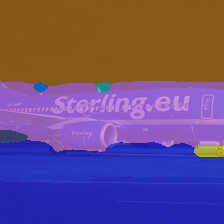} & 
      \myig{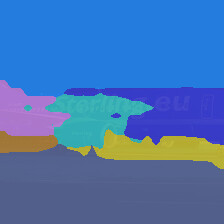} &
      \myig{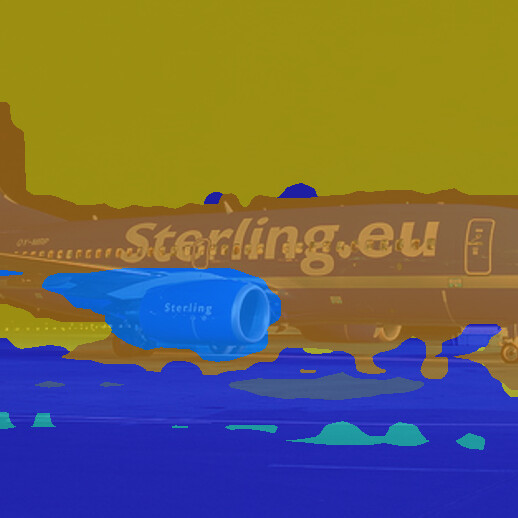} &
      \myig{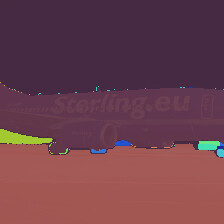} \\

      \myig{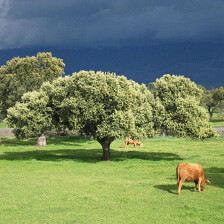} &
      \myig{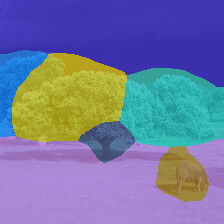} &
      \myig{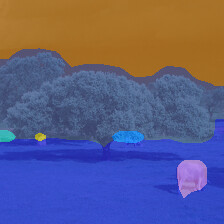} & 
      \myig{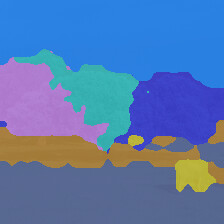} &
      \myig{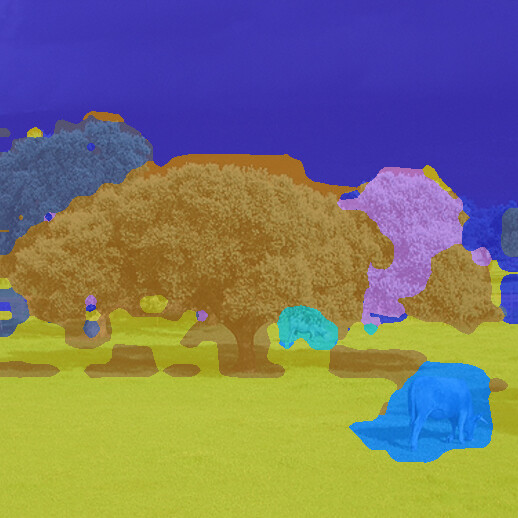} &
      \myig{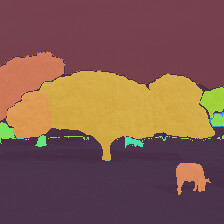} \\

      \myig{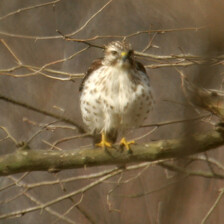} &
      \myig{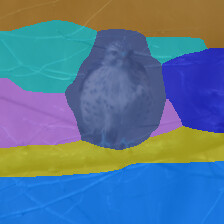} &
      \myig{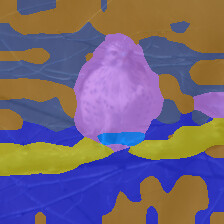} & 
      \myig{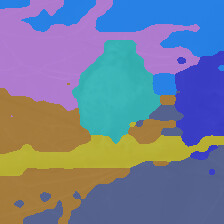} &
      \myig{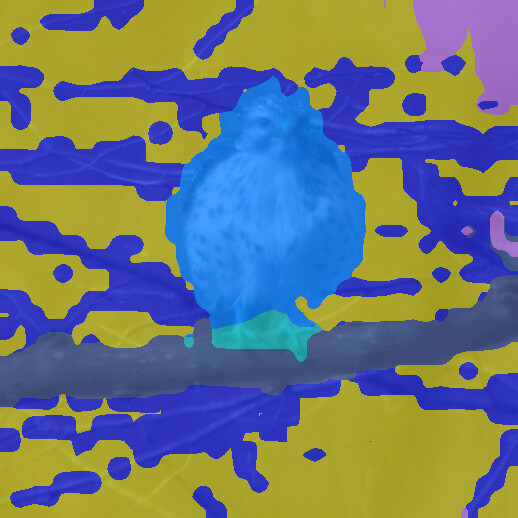} &
      \myig{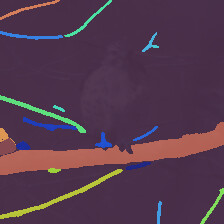} \\

      \myig{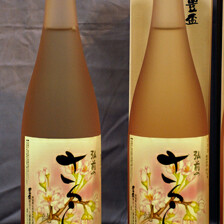} &
      \myig{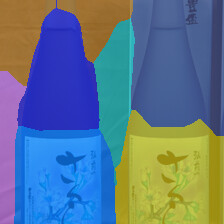} &
      \myig{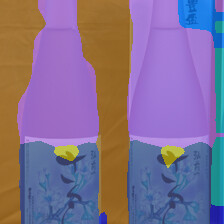} & 
      \myig{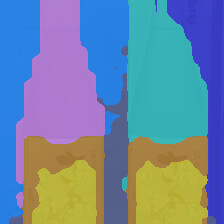} &
      \myig{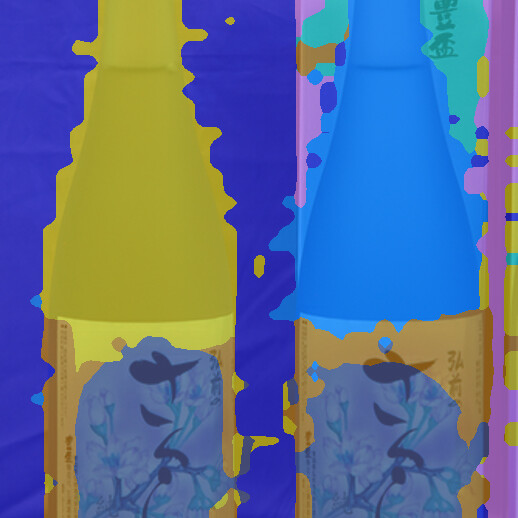} &
      \myig{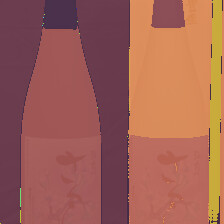} \\

      \myig{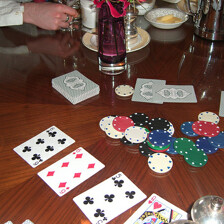} &
      \myig{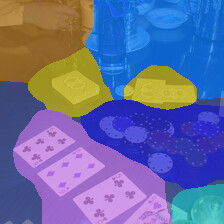} &
      \myig{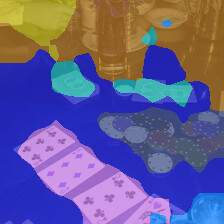} & 
      \myig{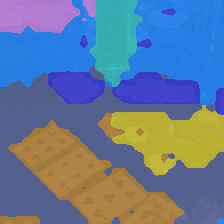} &
      \myig{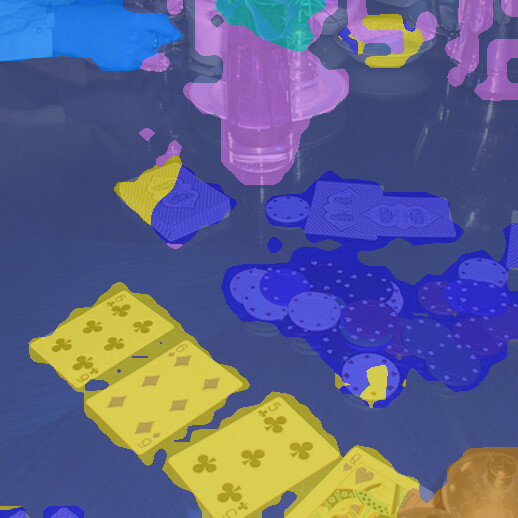} &
      \myig{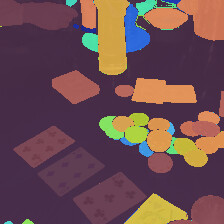} \\

      \myig{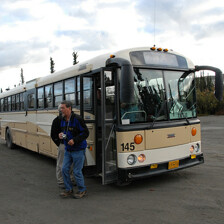} &
      \myig{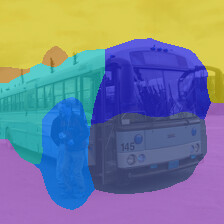} &
      \myig{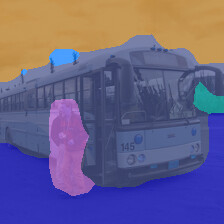} & 
      \myig{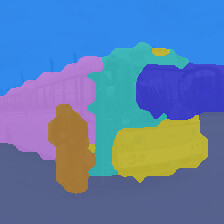} &
      \myig{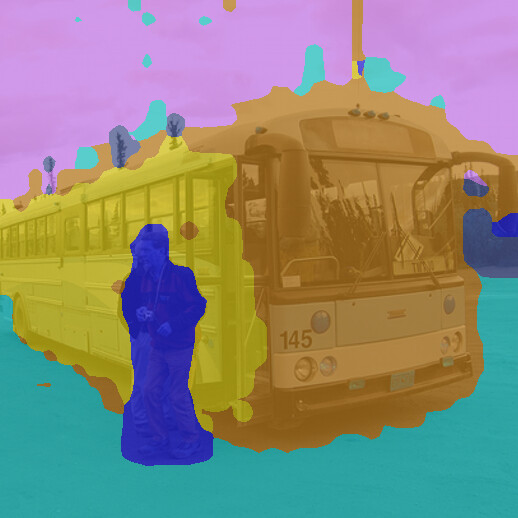} &
      \myig{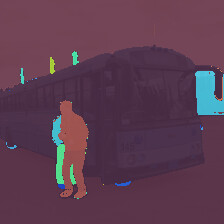} \\

      \myig{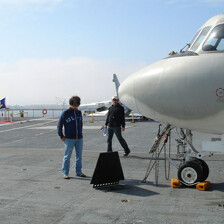} &
      \myig{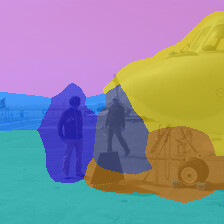} &
      \myig{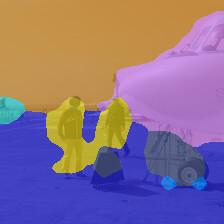} & 
      \myig{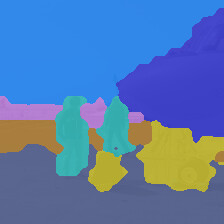} &
      \myig{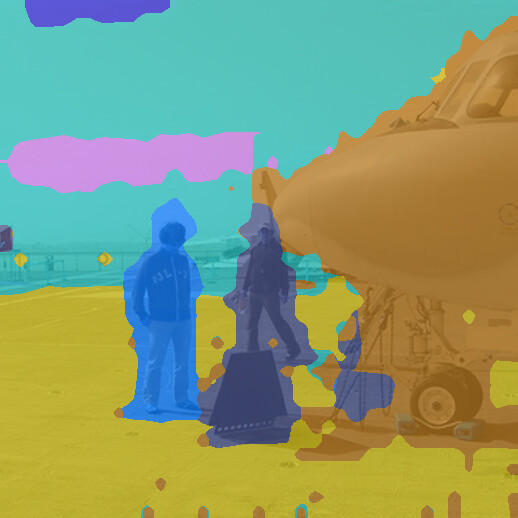} &
      \myig{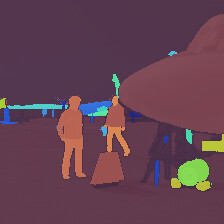} \\
  \end{tabular}
  \captionof{figure}{Zero-shot examples on PASCAL VOC.}
  \label{app:fig:examples-pascal}
\end{figure}

\begin{figure}
    \newcommand{\myig}[1]{\includegraphics[width=0.15\textwidth,valign=c]{images/#1}}
    \newcommand{\mycaption}[1]{{#1}}
    \renewcommand{\arraystretch}{4.7}
    \setlength{\tabcolsep}{2pt}
    \centering
    \vspace{-0.1\textwidth}
    \begin{tabular}{@{}cccccc@{}}
        \mycaption{Image} & \mycaption{\DINOSAUR{}} & \mycaption{\SPOT} & \mycaption{SlotDiffusion} & \mycaption{\methodshort{}} & \mycaption{\SAM} \\[-1.5em]

        \myig{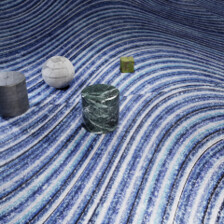} &
        \myig{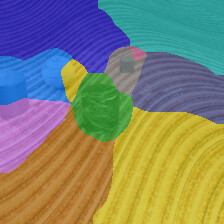} &
        \myig{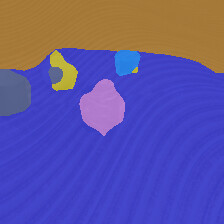} & 
        \myig{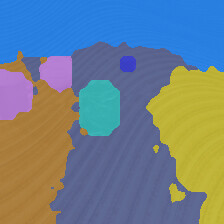} &
        \myig{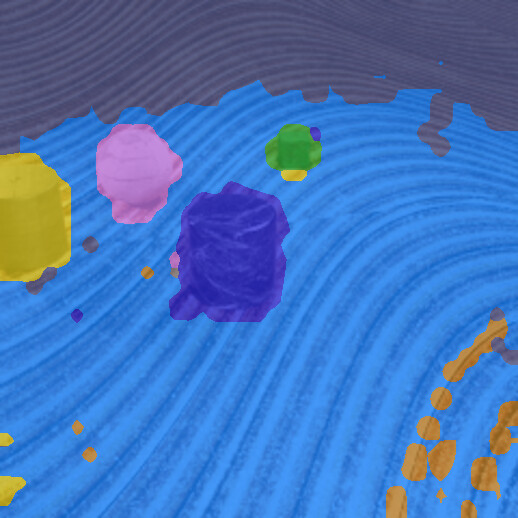} &
        \myig{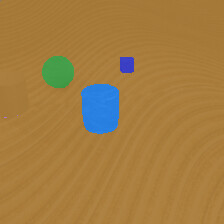} \\

        \myig{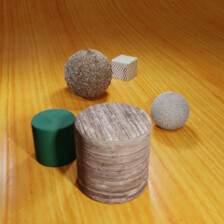} &
        \myig{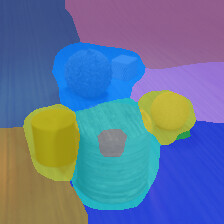} &
        \myig{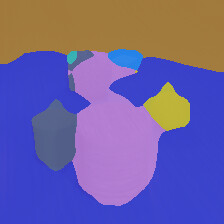} & 
        \myig{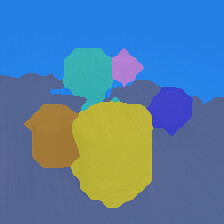} &
        \myig{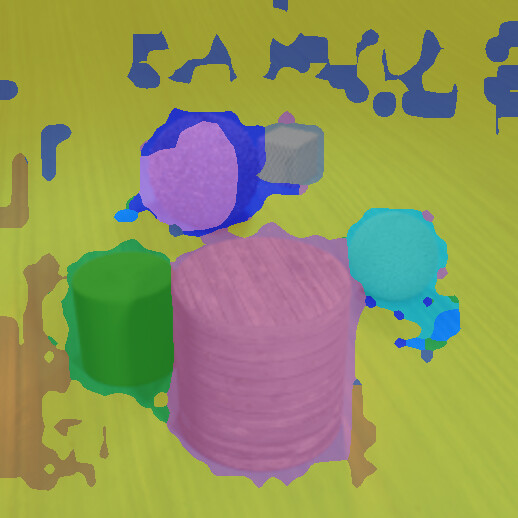} &
        \myig{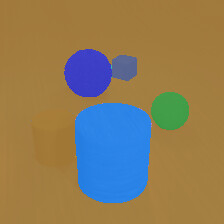} \\

        \myig{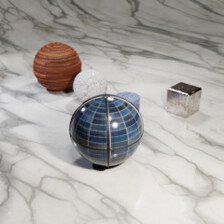} &
        \myig{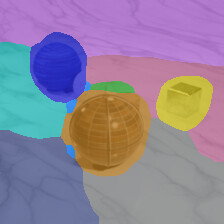} &
        \myig{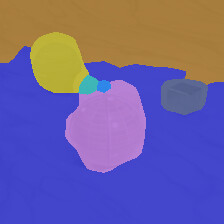} & 
        \myig{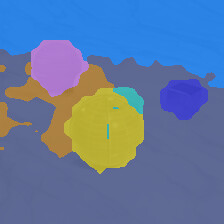} &
        \myig{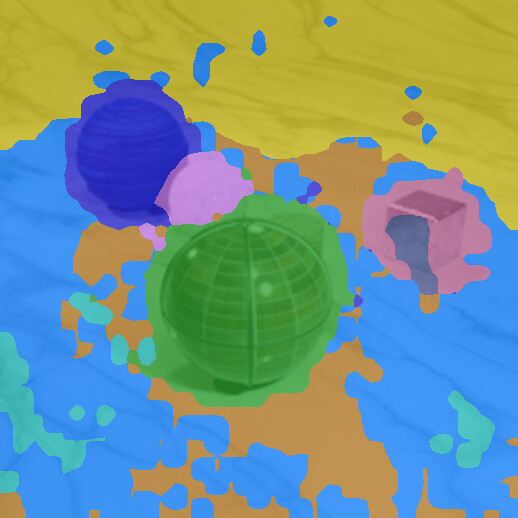} &
        \myig{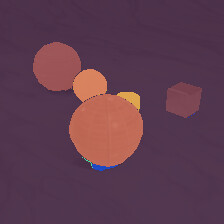} \\

        \myig{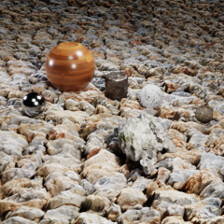} &
        \myig{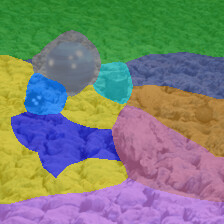} &
        \myig{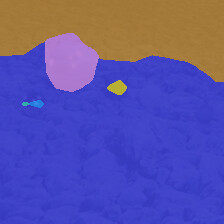} & 
        \myig{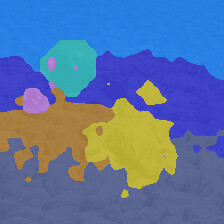} &
        \myig{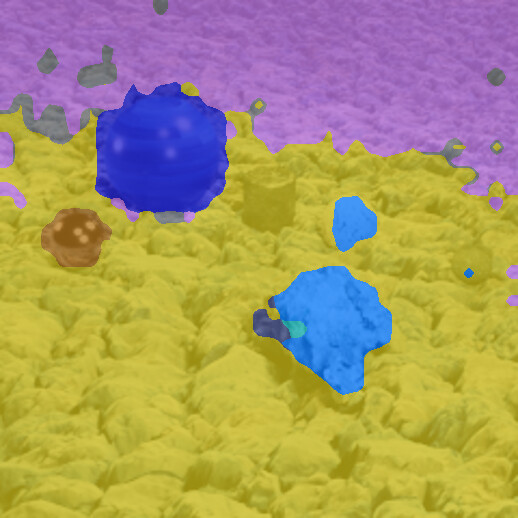} &
        \myig{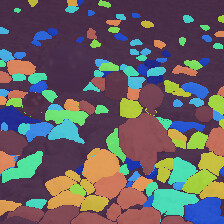} \\

        \myig{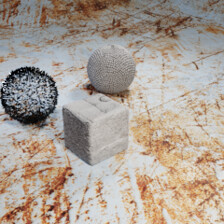} &
        \myig{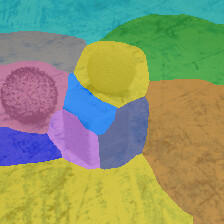} &
        \myig{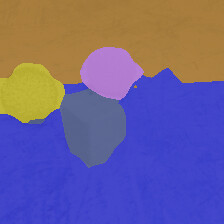} & 
        \myig{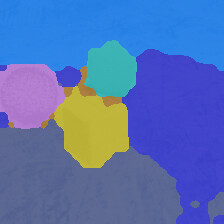} &
        \myig{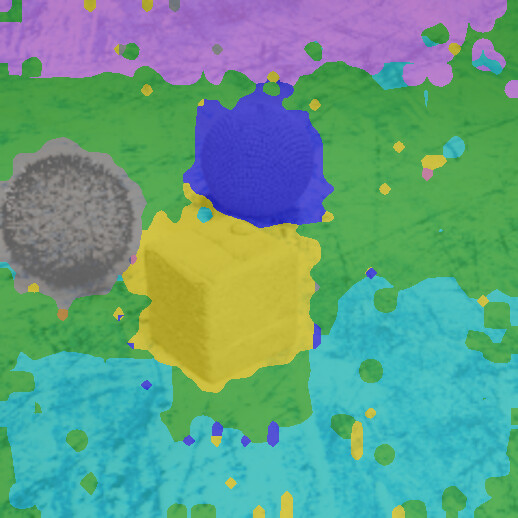} &
        \myig{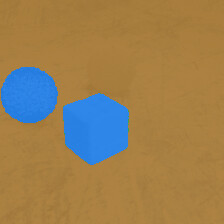} \\

        \myig{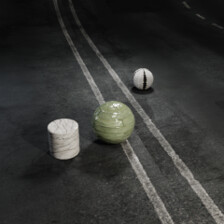} &
        \myig{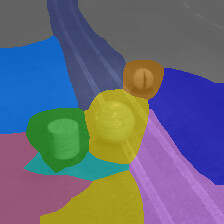} &
        \myig{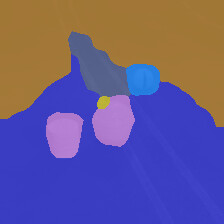} & 
        \myig{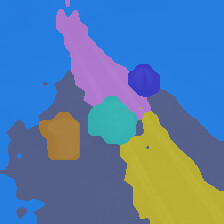} &
        \myig{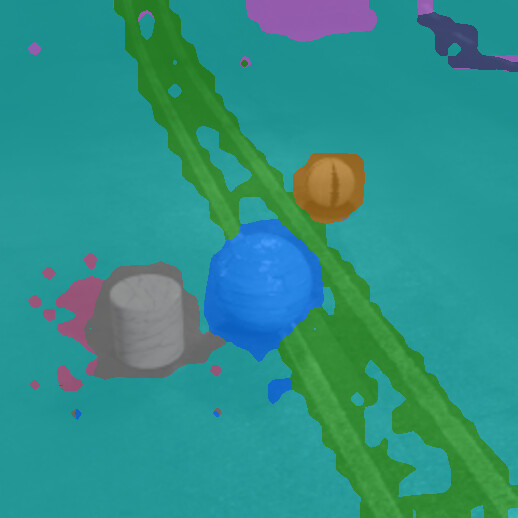} &
        \myig{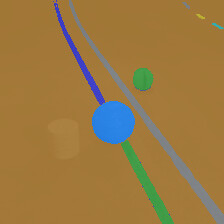} \\

        \myig{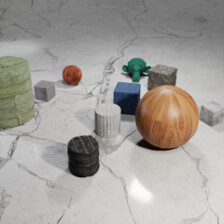} &
        \myig{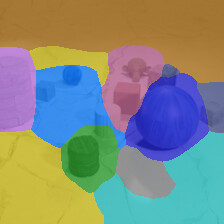} &
        \myig{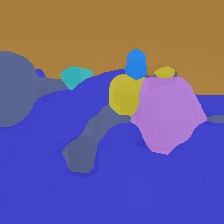} & 
        \myig{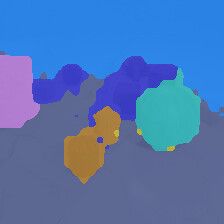} &
        \myig{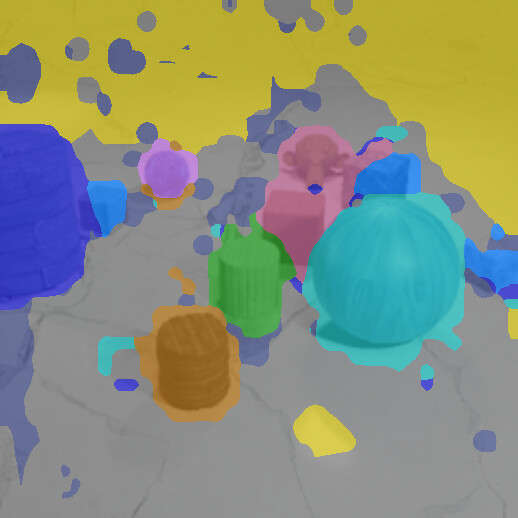} &
        \myig{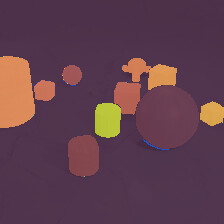} \\

        \myig{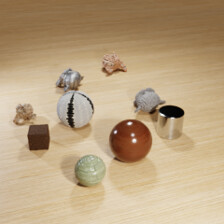} &
        \myig{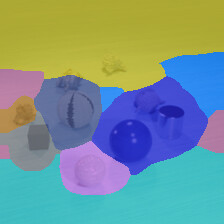} &
        \myig{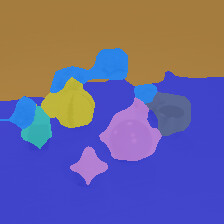} & 
        \myig{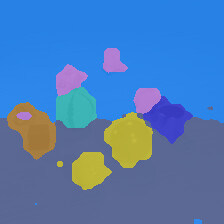} &
        \myig{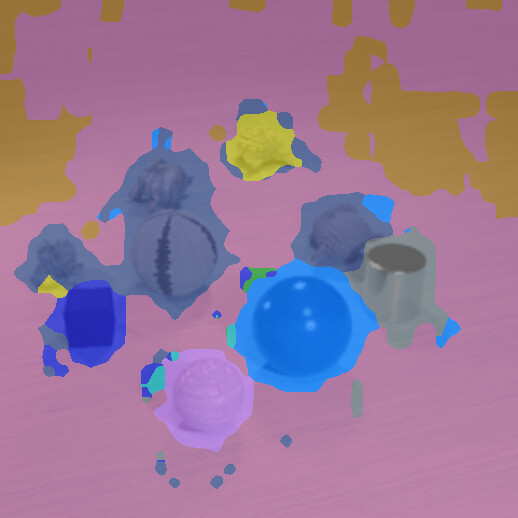} &
        \myig{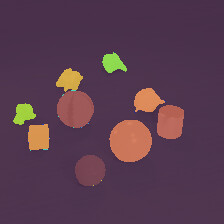} \\

        \myig{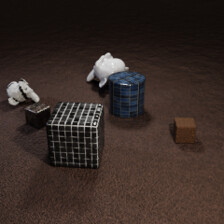} &
        \myig{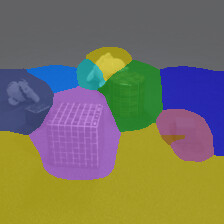} &
        \myig{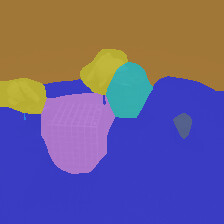} & 
        \myig{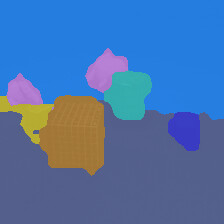} &
        \myig{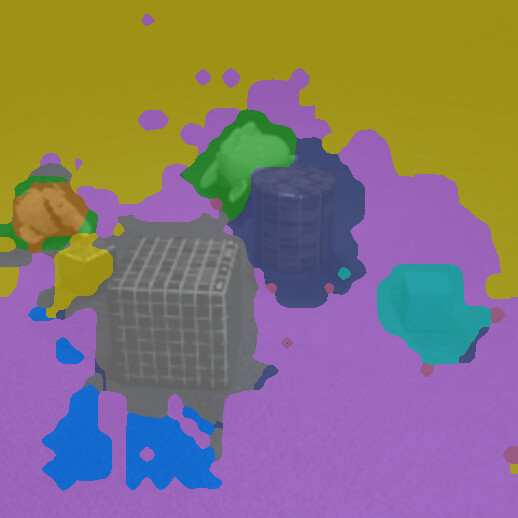} &
        \myig{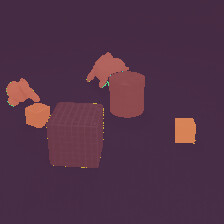} \\

        \myig{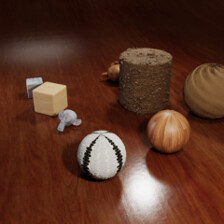} &
        \myig{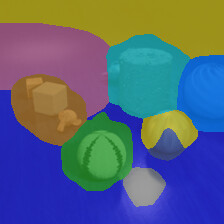} &
        \myig{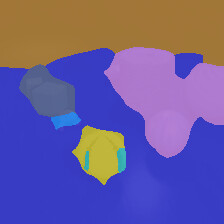} & 
        \myig{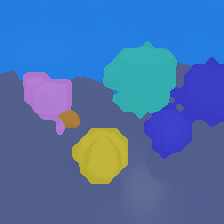} &
        \myig{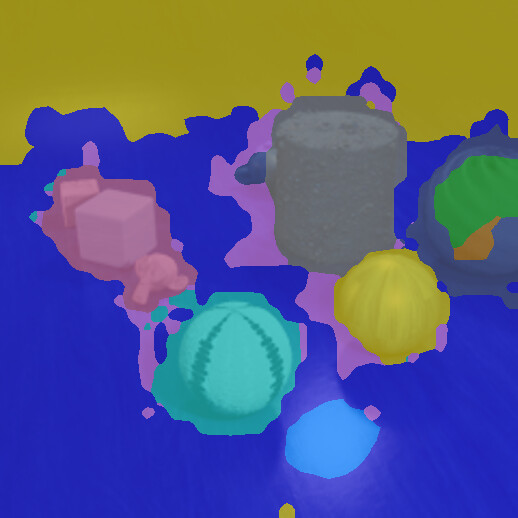} &
        \myig{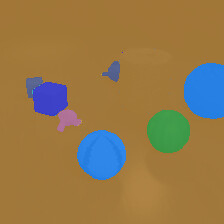} \\
    \end{tabular}
    \captionof{figure}{Zero-shot examples on \ClevrTex.}
    \label{app:fig:examples-clevrtex}
\end{figure}

\begin{figure}
  \newcommand{\myig}[1]{\includegraphics[width=0.15\textwidth,valign=c]{images/#1}}
  \newcommand{\mycaption}[1]{{\footnotesize#1}}
  \renewcommand{\arraystretch}{4.7}
  \setlength{\tabcolsep}{2pt}
  \centering
  \vspace{-0.1\textwidth}
  \begin{tabular}{@{}cccccc@{}}
      \mycaption{Image} & \mycaption{\DINOSAUR{}} & \mycaption{\SPOT} & \mycaption{SlotDiffusion} & \mycaption{\methodshort{}} & \mycaption{\SAM} \\[-1.5em]

      \myig{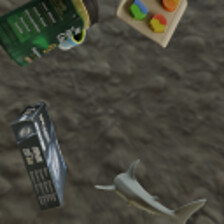} &
      \myig{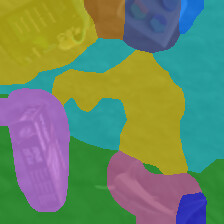} &
      \myig{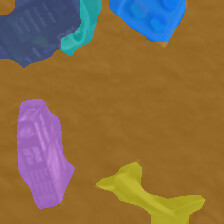} & 
      \myig{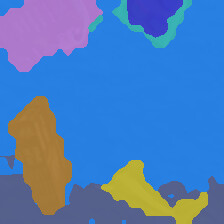} &
      \myig{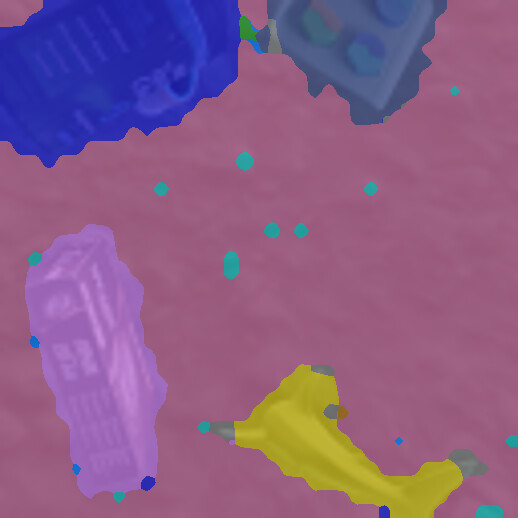} &
      \myig{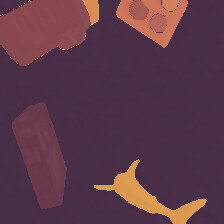} \\

      \myig{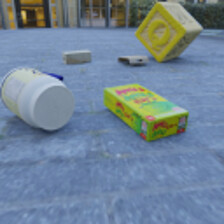} &
      \myig{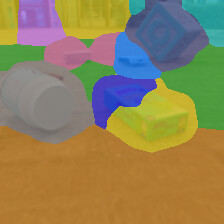} &
      \myig{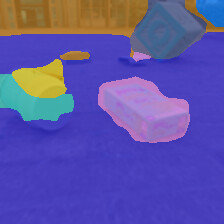} & 
      \myig{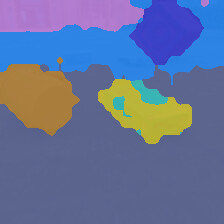} &
      \myig{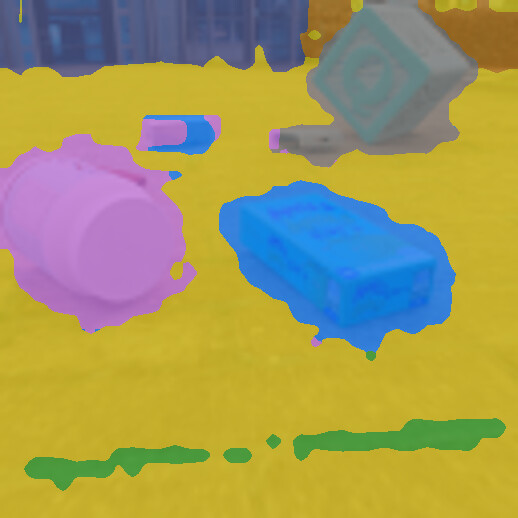} &
      \myig{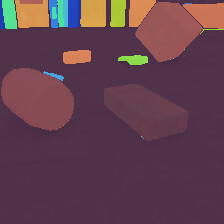} \\

      \myig{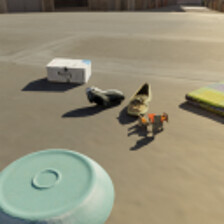} &
      \myig{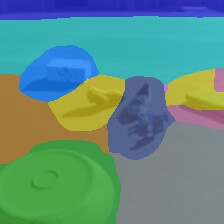} &
      \myig{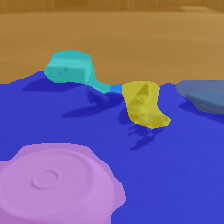} & 
      \myig{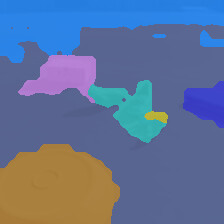} &
      \myig{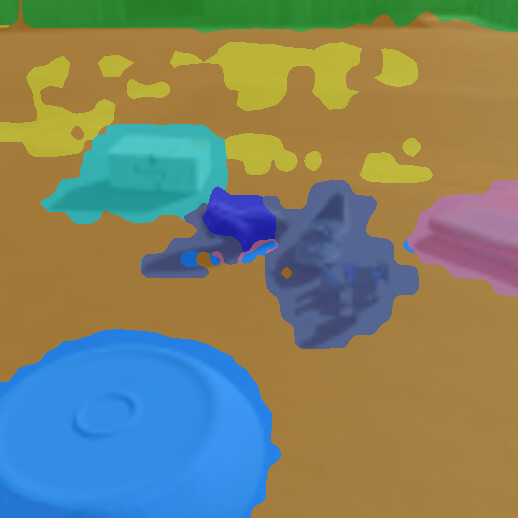} &
      \myig{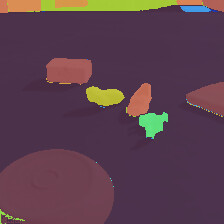} \\

      \myig{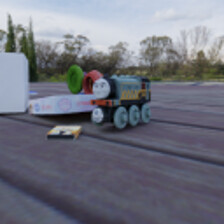} &
      \myig{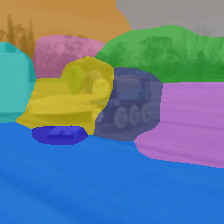} &
      \myig{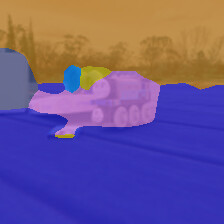} & 
      \myig{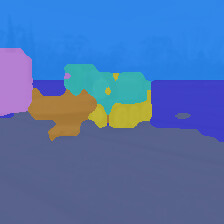} &
      \myig{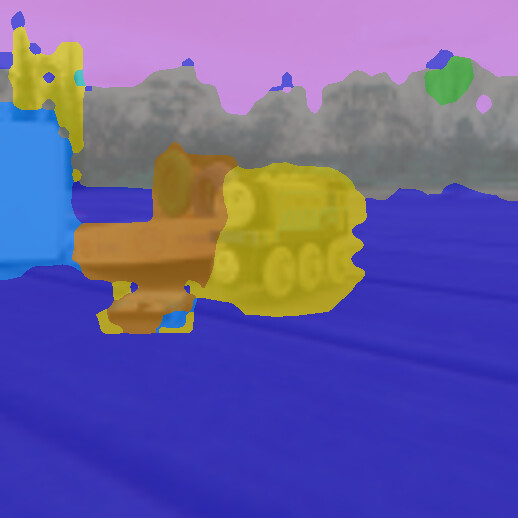} &
      \myig{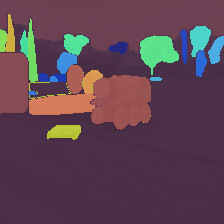} \\

  \end{tabular}
  \captionof{figure}{Zero-shot examples on \MOVi-C.}
  \label{app:fig:examples-movic}
\end{figure}

\begin{figure}
  \newcommand{\myig}[1]{\includegraphics[width=0.15\textwidth,valign=c]{images/#1}}
  \newcommand{\mycaption}[1]{{\footnotesize#1}}
  \renewcommand{\arraystretch}{4.7}
  \setlength{\tabcolsep}{2pt}
  \centering
  \vspace{-0.1\textwidth}
  \begin{tabular}{@{}cccccc@{}}
      \mycaption{Image} & \mycaption{\DINOSAUR{}} & \mycaption{\SPOT} & \mycaption{SlotDiffusion} & \mycaption{\methodshort{}} & \mycaption{\SAM} \\[-1.5em]

      \myig{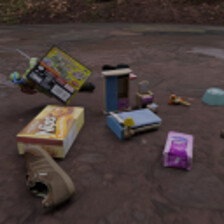} &
      \myig{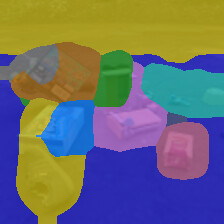} &
      \myig{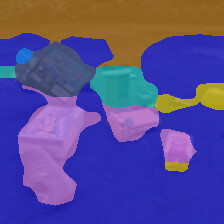} & 
      \myig{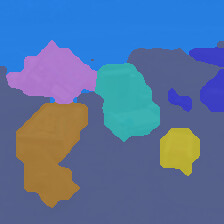} &
      \myig{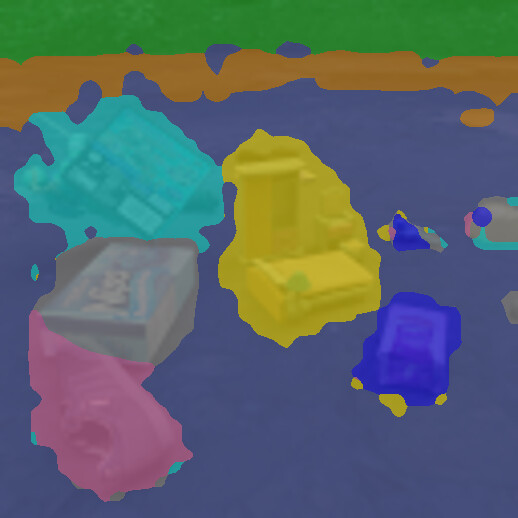} &
      \myig{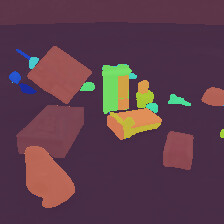} \\

      \myig{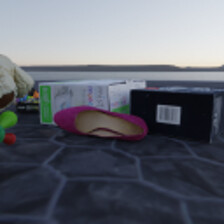} &
      \myig{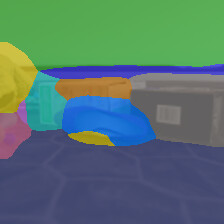} &
      \myig{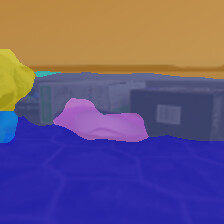} & 
      \myig{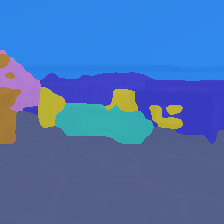} &
      \myig{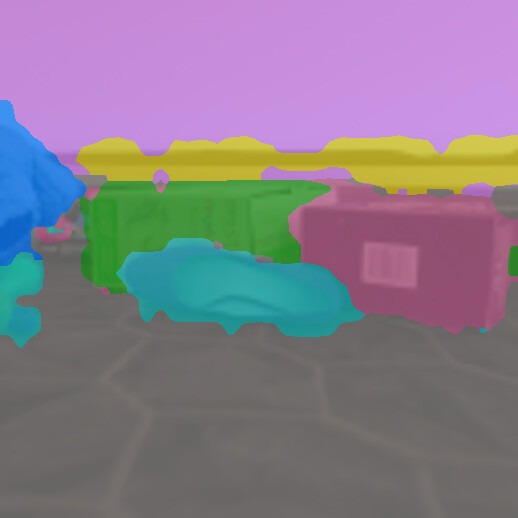} &
      \myig{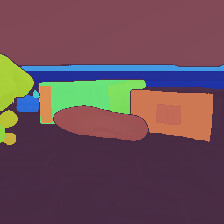} \\

      \myig{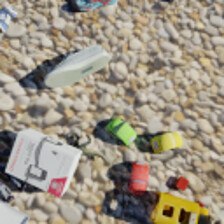} &
      \myig{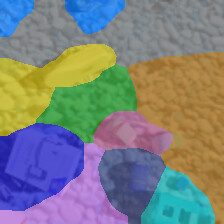} &
      \myig{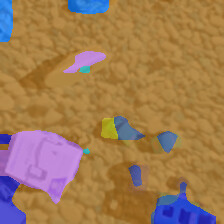} & 
      \myig{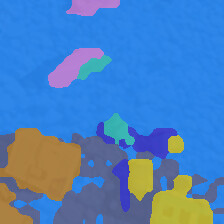} &
      \myig{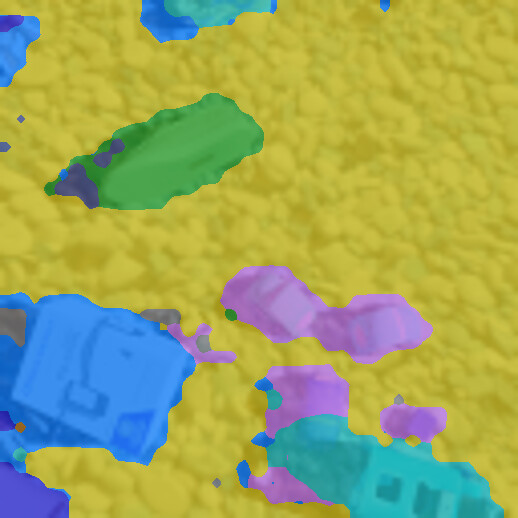} &
      \myig{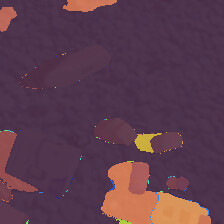} \\

      \myig{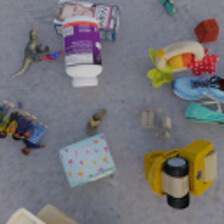} &
      \myig{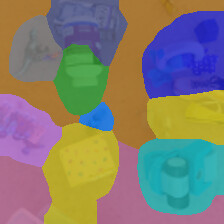} &
      \myig{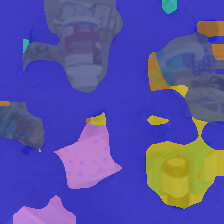} & 
      \myig{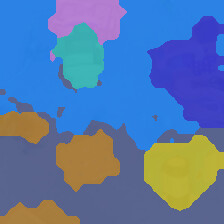} &
      \myig{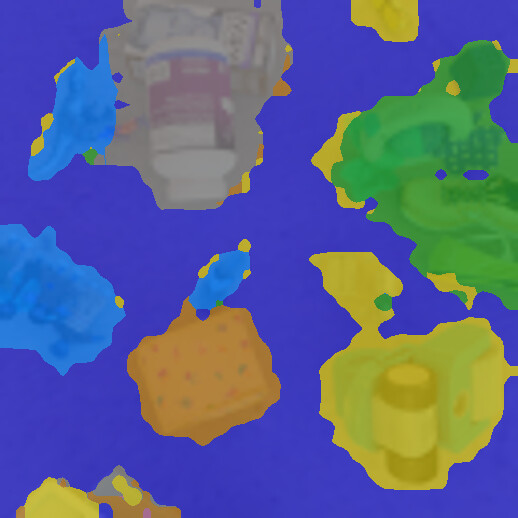} &
      \myig{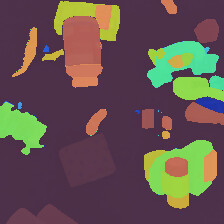} \\

  \end{tabular}
  \captionof{figure}{Zero-shot examples on \MOVi-E.}
  \label{app:fig:examples-movie}
\end{figure}

\begin{figure}
  \newcommand{\myig}[1]{\includegraphics[width=0.15\textwidth,valign=c]{images/#1}}
  \newcommand{\mycaption}[1]{{\footnotesize#1}}
  \renewcommand{\arraystretch}{4.7}
  \setlength{\tabcolsep}{2pt}
  \centering
  \vspace{-0.1\textwidth}
  \begin{tabular}{@{}cccccc@{}}
      \mycaption{Image} & \mycaption{\DINOSAUR{}} & \mycaption{\SPOT} & \mycaption{SlotDiffusion} & \mycaption{\methodshort{}} & \mycaption{\SAM} \\[-1.5em]

      \myig{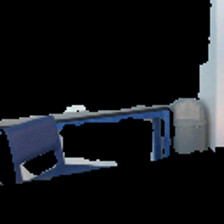} &
      \myig{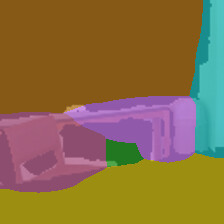} &
      \myig{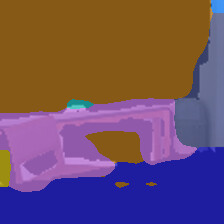} & 
      \myig{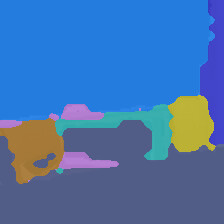} &
      \myig{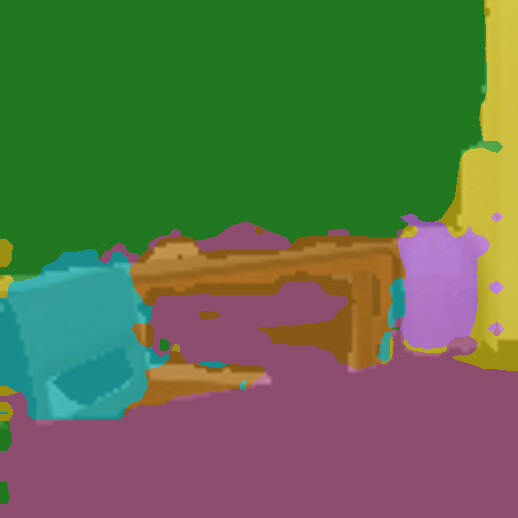} &
      \myig{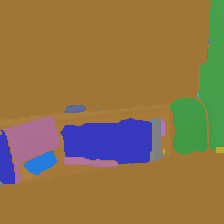} \\

      \myig{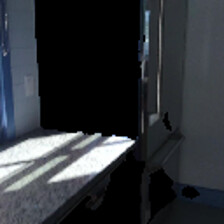} &
      \myig{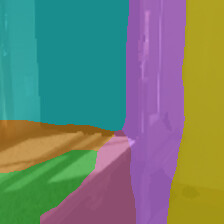} &
      \myig{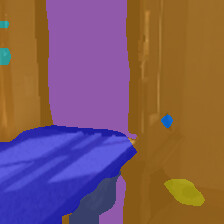} & 
      \myig{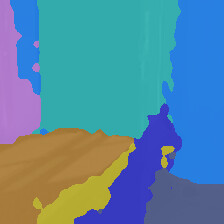} &
      \myig{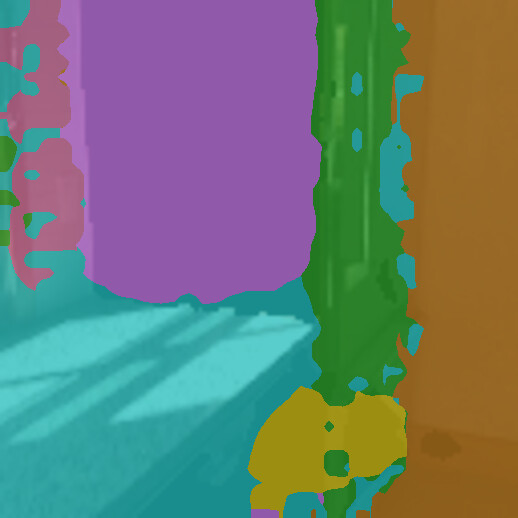} &
      \myig{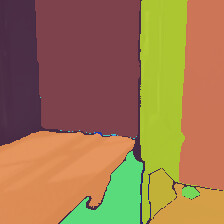} \\

      \myig{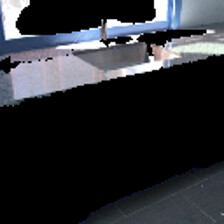} &
      \myig{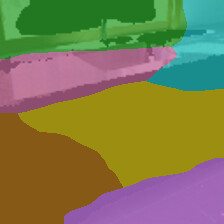} &
      \myig{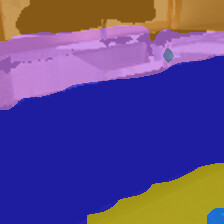} & 
      \myig{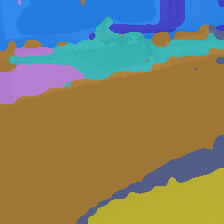} &
      \myig{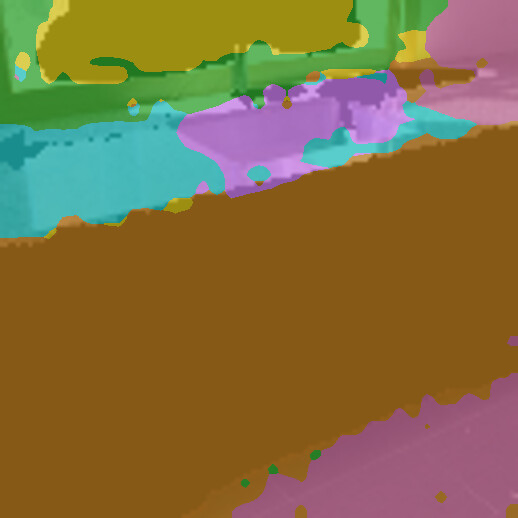} &
      \myig{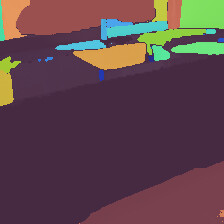} \\

      \myig{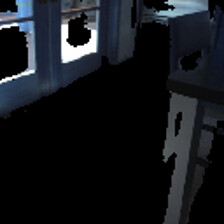} &
      \myig{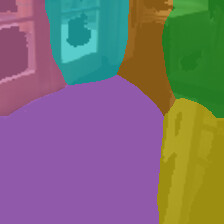} &
      \myig{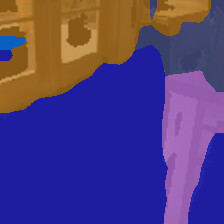} & 
      \myig{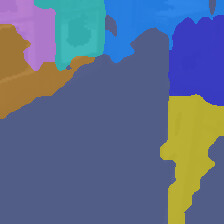} &
      \myig{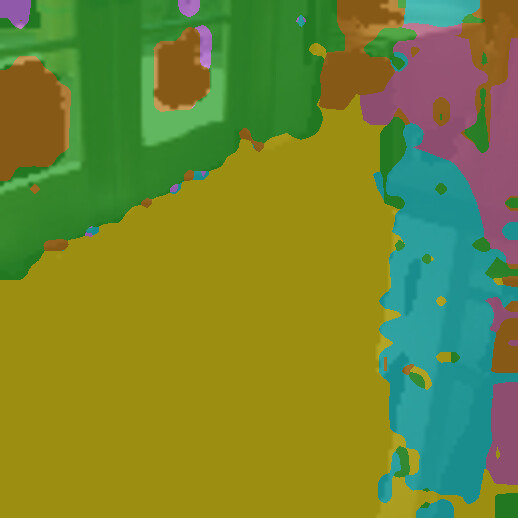} &
      \myig{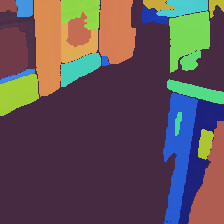} \\
  \end{tabular}
  \captionof{figure}{Zero-shot examples on \ScanNet.}
  \label{app:fig:examples-scannet}
\end{figure}

\begin{itemize}
    \item \cref{app:fig:examples-coco}: in-distribution predictions on \COCO.
    \item \cref{app:fig:examples-entityseg}: zero-shot predictions on \EntitySeg.
    \item \cref{app:fig:examples-pascal}: zero-shot predictions on \PASCALVOC.
    \item \cref{app:fig:examples-clevrtex}: zero-shot predictions on \ClevrTex.
    \item \cref{app:fig:examples-movic}: zero-shot predictions on \MOVi-C.
    \item \cref{app:fig:examples-movie}: zero-shot predictions on \MOVi-E.
    \item \cref{app:fig:examples-scannet}: zero-shot predictions on \ScanNet.
\end{itemize}

\end{document}